\documentclass[]{ceurart}

\usepackage{subfig}
\usepackage{amsmath}
\usepackage{amssymb}
\usepackage{amsthm}
\usepackage[all]{foreign}
\usepackage{paralist}

\usepackage{tikz}

\usetikzlibrary{positioning,arrows}

\newtheorem{definition}{Definition}
\newtheorem{example}{Example}
\newtheorem{proposition}{Proposition}
\newtheorem{lemma}{Lemma}

\usepackage{listings}

\definecolor{backcolour}{HTML}{F6F6F6}
\definecolor{codegreen}{HTML}{008000}

\lstdefinelanguage{EPDDL} {
    sensitive=true,           
    morecomment=[l]{;},       
    alsoletter={:,-,.,!,=,/},   
    backgroundcolor=\color{backcolour},
    commentstyle=\color{codegreen},
    basicstyle=\small\normalfont\ttfamily,
    basewidth=0.6em,
    keywordstyle=\color{blue},
    morekeywords={
        define,domain,problem,action-type-library,basic,
        not,and,or,imply,iff,when,forall,exists,exists!,=,/=,in,either,
        :and,:forall,
        if,else-if,else,
        :domain,:requirements,:types,object,agent,world,event,:constants,
        :objects,:agents,:agent-groups,
        :predicates,:fact,:action-type-libraries,
        :event,:action,:action-type,:conditions,
        :propositional-precondition,:propositional-postconditions,:propositional-event,
        :trivial-precondition,:trivial-postconditions,:trivial-event,
        :non-trivial-precondition,:non-trivial-postconditions,:non-trivial-event,
        :parameters,:precondition,:effects,
        :events,:relations,:designated,
        :observability-types,:observability-conditions,
        :init,:facts,:facts-init,:worlds,:labels,
        :goal,
        default,
        C.,Kw.
    }
}

\usepackage{todonotes}
\usepackage{etoolbox}
\usepackage{environ}

\usepackage{xparse}

\newcommand{\defaultStyle}[1]{\ensuremath{#1}}
\newcommand{\agentStyle}[1]{\defaultStyle{#1}}
\newcommand{\fluentStyle}[1]{\defaultStyle{#1}}
\newcommand{\objectStyle}[1]{\defaultStyle{#1}}
\newcommand{\actionStyle}[1]{\defaultStyle{#1}}

\newcommand{\mymathit}[1]{\text{\textit{#1}}}

\newcommand{\LProp}{\mathcal{L}_{Prop}}
\newcommand{\LangSingle}{\mathcal{L}_{\atomSet}}
\newcommand{\Lang}[1][]{\mathcal{L}_{\atomSet,\agentSet}^{#1}}

\newcommand{\B}[1]{\Box_{\agentStyle{#1}}}
\newcommand{\D}[1]{\Diamond_{\agentStyle{#1}}}
\newcommand{\CK}[1]{C_{\agentStyle{#1}}}
\newcommand{\CKd}[1]{\hat{C}_{\agentStyle{#1}}}
\newcommand{\Kw}[1]{\mymathit{Kw}_{\agentStyle{#1}}}
\newcommand{\Kwd}[1]{\hat{\mymathit{Kw}}_{\agentStyle{#1}}}

\newcommand{\atomSet}{\mymathit{P}}
\newcommand{\agentSet}{\mymathit{Ag}}
\newcommand{\actionSet}{\mymathit{Act}}
\newcommand{\obsTypeSet}{\mymathit{ObsTypes}}

\newcommand{\obsType}[2][]{\mymathit{type}_{#1}(\agentStyle{#2})}
\newcommand{\inducedAct}[2]{\actionStyle{#1}_{\downarrow #2}}
\newcommand{\abstraction}[1]{#1_{\uparrow}}

\newcommand{\objectSet}{\mymathit{Obj}}
\newcommand{\groupSet}{\mymathit{Groups}}
\newcommand{\typeSet}{\mymathit{Types}}
\newcommand{\primitiveTypeSet}{\mymathit{PrimitiveTypes}}
\newcommand{\predSet}{\mymathit{Pred}}
\newcommand{\constSet}{\mymathit{Const}}
\newcommand{\entitySet}{\mymathit{Entities}}
\newcommand{\entityType}{\mymathit{entityType}}
\newcommand{\factSet}{\mymathit{\mymathit{Facts}}}
\newcommand{\positiveFacts}{s_\mymathit{{facts}}}

\newcommand{\compTypes}[2]{\mathtt{#1} \preceq \mathtt{#2}}
\newcommand{\typedEntitySet}[1]{\mymathit{Ent}_{\mathtt{#1}}}

\newcommand{\Pre}[1]{\mathsf{Pre}(\actionStyle{#1})}
\NewDocumentCommand{\Eff}{ O{} O{} m }{\mathsf{Eff}_{#1}^{#2}(\actionStyle{#3})}

\makeatletter

\newtoks\AllEBNFToks
\AllEBNFToks={}

\newcommand\AppendToAllEBNFTokenwise[1]{%
  \begingroup
    \toks0=\expandafter{\the\AllEBNFToks}%
    \toks2=\expandafter{\unexpanded{#1}}%
    \global\AllEBNFToks=\expandafter{\the\toks0\the\toks2}%
  \endgroup
}

\NewEnviron{ebnf}[1][lrl]{%
  \vspace{6pt}\noindent\footnotesize
  \begin{tabular}{#1}
    \BODY
  \end{tabular}
  \vspace{6pt}%
}

\newcommand{\ebnfToken}[1]{{\texttt{#1}}}

\newcommand{\ebnfDecl}[2][]{{\texttt{$\langle$#2$\rangle^{\text{#1}}$}}}

\newcommand{\ebnfList}[1]{{\texttt{#1}$^\text{*}$}}
\newcommand{\ebnfNonEmptyList}[1]{{\texttt{#1}$^\text{+}$}}

\newcommand{\ebnfDeclList}[1]{{$\langle$\texttt{#1}$\rangle^\text{*}$}}
\newcommand{\ebnfDeclNonEmptyList}[1]{{$\langle$\texttt{#1}$\rangle^\text{+}$}}
\newcommand{\ebnfDeclOptional}[1]{{$\langle$\texttt{#1}$\rangle^\text{?}$}}

\newcommand{\ebnfProdArrow}[1][]{{\texttt{$^{\text{#1}}{::=}$}}}
\newcommand{\ebnfProdOr}[1][]{{\texttt{$^{\text{#1}}{\mid}$}}}

\newcommand{\ebnfLeftPar}{{\texttt{(}}}
\newcommand{\ebnfRightPar}{{\texttt{)}}}

\newcommand{\ebnfLeftCurl}{{\texttt{\{}}}
\newcommand{\ebnfRightCurl}{{\texttt{\}}}}

\tikzstyle{world} =[circle,    thick, draw=black,         fill=black, minimum size=5pt, inner sep=0pt]
\tikzstyle{dworld}=[circle,    thick, draw=black, double, fill=black, minimum size=5pt, inner sep=0pt]
\tikzstyle{event} =[rectangle, thick, draw=black,         fill=black, minimum size=5pt, inner sep=0pt, outer sep=2pt]
\tikzstyle{devent}=[rectangle, thick, draw=black, double, fill=black, minimum size=5pt, inner sep=0pt, outer sep=2pt]

\tikzstyle{graphicalworld} =[rectangle,         rounded corners, draw=black, thick, inner sep=6pt, outer sep=2pt]
\tikzstyle{graphicaldworld}=[rectangle, double, rounded corners, draw=black, thick, inner sep=6pt, outer sep=2pt]

\tikzstyle{block}=[rectangle, draw=black, thick, minimum size=20pt, outer sep=1pt]
\tikzstyle{wblock}=[rectangle, draw=white, thick, minimum size=20pt, outer sep=1pt]
\tikzstyle{stack}=[draw=white, minimum width=20pt]

\begin{document}
    \copyrightyear{2026}
    \copyrightclause{Copyright for this paper by its authors.
    Use permitted under Creative Commons License Attribution 4.0
    International (CC BY 4.0).}

    \conference{Official Guideline Reference for the Epistemic Planning Track at IPC-26}

    \title{The Epistemic Planning Domain Definition Language: Official Guideline}

    \author[1]{Alessandro Burigana}[%
        orcid=0000-0002-9977-6735,
        email=alessandro.burigana@unibz.it
    ]

    \author[2]{Francesco Fabiano}[%
        orcid=0000-0002-1161-0336,
        email=francesco.fabiano@cs.ox.ac.uk,
    ]

    \address[1]{Free University of Bozen-Bolzano, Bolzano, Italy}
    \address[2]{University of Oxford, Oxford, UK}


    \maketitle

    \begin{abstract}
    Epistemic planning extends (multi-agent) automated planning by making agents' knowledge and beliefs first-class aspects of the planning formalism.
    One of the most well-known frameworks for epistemic planning is \emph{Dynamic Epistemic Logic} (DEL), which offers an rich and natural semantics for modelling problems in this setting.
    The high expressive power provided by DEL make DEL-based epistemic planning a challenging problem to tackle both theoretically, and in practical implementations.
    As a result, existing epistemic planners often target different DEL fragments, and typically rely on ad hoc languages to represent benchmarks, and sometimes no language at all.
    This fragmentation hampers comparison, reuse, and systematic benchmark development.
    We address these issues by introducing the \emph{Epistemic Planning Domain Definition Language} (\textsc{epddl}).
    \textsc{epddl} provides a unique \textsc{pddl}-like representation that captures the entire DEL semantics, enabling uniform specification of epistemic planning tasks.
    Our main contributions are:
    \begin{inparaenum}
        \item A formal development of \emph{abstract event models}, a novel representation for epistemic actions used to define the semantics of our language;
        \item A formal specification of \textsc{epddl}'s syntax and semantics grounded in DEL with abstract event models.
    \end{inparaenum}
    Through examples of representative benchmarks, we illustrate how \textsc{epddl} facilitates interoperability, reproducible evaluation, and future advances in epistemic planning.
\end{abstract}

    \section{Introduction}
    This guideline describes \textsc{epddl}, a standardized language for expressing epistemic planning problems.
    This document is the official guideline for \textsc{epddl} to be used by participants of the epistemic planning track at IPC 2026. It should be used as the reference for encoding problems, submitting benchmarks, and interpreting results in that competition.
    
    The main goals of EPDDL are to:
    \begin{inparaenum}
        \item Provide a clear and concise syntax for representing epistemic planning benchmarks;
        \item Give a formal semantic foundation to the language based on Dynamic Epistemic Logic (DEL); and
        \item Provide a standard way of defining DEL fragments, to enable meaningful comparison of different epistemic planners.
    \end{inparaenum}

    The guide is written for two audiences.
    For experts in epistemic planning it provides a precise specification and reference semantics.
    For readers outside the subfield it includes an accessible introduction to the underlying DEL framework.
    The preliminaries on DEL are presented with minimal technical overhead so that practitioners familiar with classical planning can follow the modelling and semantics sections without prior DEL expertise.

    The remainder of the guideline is structured as follows.
    In Section~\ref{sec:preliminaries} we provide an accessible introduction to DEL-based epistemic planning.
    We start from the basics of automated planning and then we increment the planning formalisms piece by piece, by adding knowledge/beliefs, multiple agents, local perspectives, non determinism and common knowledge.
    To illustrate all new concepts, we rely on a variation of the well-known \emph{Blocks World} problems called \emph{Epistemic Blocks World}.
    In Section~\ref{sec:abstract-epistemic-action}, we introduce \emph{abstract epistemic actions}, a slight modification of standard actions in DEL-based epistemic planning that are going to be the foundation of actions' semantics in \textsc{epddl}.
    Sections~\ref{sec:epddl-syntax} and~\ref{sec:epddl-semantics} illustrate, we introduce the syntax and semantics of the \textsc{epddl} language, respectively, providing helpful examples to illustrate each component.
    Finally, in Section~\ref{sec:json-syntax} we provide a simplified JSON syntax for ground epistemic planning tasks that can be directly used as input of epistemic planners.
    A JSON specification can be automatically generated from an \textsc{epddl} specification by using the \emph{\textsf{plank} toolkit for DEL-based epistemic planning}.
    The tool is publicly available online in its \href{https://github.com/a-burigana/plank}{GitHub repository}.
    Documentation for installation and usage of \textsf{plank} are available in the repository.

    For all informations regarding organizational aspects of the Epistemic Planning Track at IPC 2026, please consult the \href{https://sites.google.com/view/epistemic-competition/}{official website}, or contact the organizers.
    
    \section{DEL-based Epistemic Planning}\label{sec:preliminaries}
    In this section, we give the foundations of \emph{Dynamic Epistemic Logic} (\emph{DEL}) and DEL-based epistemic planning.
    To provide an accessible introduction also to a less familiar reader, we begin from the framework of classical planning and we gradually move towards DEL-based epistemic planning by incrementally enriching the formalism.
    In this transition from classical to epistemic planning, we discuss in detail each of the many features that the DEL framework offers in its rich semantics.
    During the exposition, we use the well-known \emph{Blocks World} domain as a running example, progressively enriching it to demonstrate increasingly complex epistemic planning scenarios and their natural representation in DEL.
    Readers already familiar with DEL and DEL-based epistemic planning may skip this section, as it is primarily intended for those new to epistemic planning.

    \subsection{Classical Planning}\label{sec:classical-planning}
    We recall the basic notions of classical planning.
    For a more complete overview consult \cite{books/elsevier/Gallab2004}.
    Let $\atomSet$ be a fixed finite set of atomic propositions (\emph{atoms}).
    The language $\LProp$ of propositional logic on $\atomSet$ is given by the BNF: $ \phi ::= \fluentStyle{p} \mid \neg \phi \mid \phi \land \phi$.
    The standard propositional symbols $\lor$, $\rightarrow$, $\leftrightarrow$, $\top$ are defined in the usual way (\ie $\phi\lor\psi\equiv\neg(\neg\phi\land\neg\psi)$, $\phi\to\psi\equiv\neg\phi\lor\psi$, $\phi\leftrightarrow\psi\equiv(\phi\to\psi)\land(\psi\to\phi)$, $\top\equiv p\lor\neg p$, $\bot\equiv\neg\top$).
    A \emph{literal} $\ell$ of $\atomSet$ is either an atom $\fluentStyle{p}\in \atomSet$ or its negation.
    If $L$ is a set of literals, we let $L^+$ denote the set of \emph{positive literals} of $L$ (the atoms in $L$), and $L^- = L \setminus L^+$ the set of \emph{negative literals} of $L$.
    
    A \emph{classical (planning) state} of $\LProp$ is a subset $S \subseteq 2^\atomSet$ of atoms, where $\fluentStyle{p} \in S$ means that $\fluentStyle{p}$ is true in the state of affairs represented by $S$.
    As customary, we denote the fact that a formula $\phi \in \LProp$ holds in a classical state $S$ (according to standard propositional semantics) with $S \models \phi$.
    We say that $\phi$ is \emph{satisfiable} if there exists a state $S$ such that $S \models \phi$, and that $\phi$ is \emph{valid}, denoted $\models \phi$, if $\phi$ holds in all classical states of $\LProp$.
    A \emph{classical (planning) action} of $\LProp$ is a pair $\actionStyle{a} = \langle\Pre{a}, \Eff{a}\rangle$, where $\Pre{a} \in \LProp$ is the \emph{precondition} of $\actionStyle{a}$, and $\Eff{a}$ is its set of \emph{conditional effects} of the form $c \triangleright e$, where $c \in \LProp$ is a formula and $e$ is a set of literals of $\atomSet$.
    We say that $\actionStyle{a}$ is \emph{applicable} in a classical state $S$ iff $S \models \Pre{a}$, and, if so, the \emph{update} of $\actionStyle{a}$ in $S$ is the classical state $\displaystyle S \circ \actionStyle{a} = (S \setminus \Eff[S][-]{a}) \cup \Eff[S][+]{a}$, where $\Eff[S]{a} = \{c \triangleright e \in \Eff{a} \mid S \models c\}$ is the set of conditional effects that are triggered in $S$, $\Eff[S][+]{a} = \{e^+ \mid c \triangleright e \in \Eff[S]{a}\}$ are the positive effects and $\Eff[S][-]{a} = \{e^- \mid c \triangleright e \in \Eff[S]{a}\}$ the negative effects.
    We require that conditional effects are consistent, \ie $\Eff[S][+]{a} \cap \Eff[S][-]{a} = \varnothing$.


    \begin{definition}[Classical Planning Task]%
    \label{def:calssical-planning-task}
        A \emph{classical planning task} is a triple $T = (I, A, G)$, where $I$ is a classical state (the \emph{initial state}), $A$ is a finite set of classical actions, and $G \in \LProp$ is the \emph{goal formula}.
        A \emph{solution} (or \emph{plan}) to $T$ is a finite sequence $\pi = \actionStyle{a}_1, \dots, \actionStyle{a}_l$ of classical actions of $A$ such that:
        \begin{enumerate}
            \item For all $1 \leq k \leq l$, $\actionStyle{a}_k$ is applicable in $I \circ \actionStyle{a}_1 \circ \dots \circ \actionStyle{a}_{k-1}$;
            \item $I \circ \actionStyle{a}_1 \circ \dots \circ \actionStyle{a}_l \models G$.
        \end{enumerate}
    \end{definition}

    We now introduce our basic version of the famous Blocks World problem~\cite{books/aw/RN2020}.
    Here we present a slight variation of the traditional problem where blocks can only be piled on a non-empty, finite set of columns.

\begin{figure}
        \centering
        \begin{tikzpicture}
    [node distance=1pt and 16pt,
    block/.style={rectangle, rounded corners, draw=black, thick, minimum size=20pt},
    stack/.style={draw=white, minimum width=20pt}]

    \node[block] (b1)                   {$\objectStyle{b}_1$};
    \node[block] (b2) [above=of b1]     {$\objectStyle{b}_2$};
    \node[block] (b3) [right=of b1]     {$\objectStyle{b}_3$};
    \node[block] (b4) [right=of b3]     {$\objectStyle{b}_4$};

    \draw[gray, very thick] ([xshift=-30pt,yshift=-2pt]b1.south) -- ([xshift=104pt,yshift=-2pt]b1.south);

    \node[stack] (s1) [below=4pt of b1] {$\objectStyle{c}_1$};
    \node[stack] (s2) [right=of s1]     {$\objectStyle{c}_2$};
    \node[stack] (s3) [right=of s2]     {$\objectStyle{c}_3$};
\end{tikzpicture}
        \caption{Initial state $I$ of Example \ref{ex:classical-bw}. Each block is represented by a rounded square and is labeled by its name. Column names are placed below the gray line.}
        \label{fig:classical-bw}
    \end{figure}
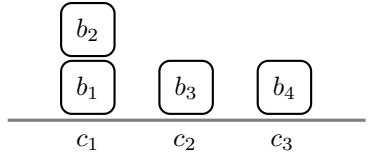

    \begin{example}[Blocks World]\label{ex:classical-bw}
        Let $B = \{\objectStyle{b}_1, \dots, \objectStyle{b}_k\}$ be a set of cube-shaped blocks and let $C = \{\objectStyle{c}_1, \dots, \objectStyle{c}_h\}$ be a set of columns.
        A mechanical arm can move one (clear) block at a time from its current position to either the top of another block or on an empty column.
        Blocks are stacked in such a way that at most one block can fit on top of another.
        Given an initial configuration of the blocks, the goal is to move the blocks in order to obtain some desired configuration.
        Let $\fluentStyle{\mymathit{On}}(\objectStyle{b}, \objectStyle{x})$ denote that block $\objectStyle{b}$ is on top $\objectStyle{x}$, and $\mymathit{Clear}(\objectStyle{x})$ that no block is on top of $\objectStyle{x}$, where $\objectStyle{x}$ is either another block or a column, so $\atomSet = \{ \fluentStyle{\mymathit{On}}(\objectStyle{b}, \objectStyle{x}), \mymathit{Clear}(\objectStyle{x}) \mid \objectStyle{b} \in B, \objectStyle{x} \in B \cup C \text{ and } \objectStyle{b} \neq \objectStyle{x} \}$.
        Let $I = \{ \fluentStyle{\mymathit{On}}(\objectStyle{b}_1, \objectStyle{c}_1), \fluentStyle{\mymathit{On}}(\objectStyle{b}_2, \objectStyle{b}_1), \fluentStyle{\mymathit{On}}(\objectStyle{b}_3, \objectStyle{c}_2), \fluentStyle{\mymathit{On}}(\objectStyle{b}_4, \objectStyle{c}_3), \mymathit{Clear}(b_2), \mymathit{Clear}(b_3), \mymathit{Clear}(b_4) \}$ be the initial state, graphically represented in Figure~\ref{fig:classical-bw}.
        The move of block $\objectStyle{b}$ from position $\objectStyle{x}$ to $\objectStyle{y}$ is described by the classical action $\actionStyle{move}(\objectStyle{b}, \objectStyle{x}, \objectStyle{y})$, where $\Pre{\actionStyle{move}(\objectStyle{b}, \objectStyle{x}, \objectStyle{y})} = \fluentStyle{\mymathit{On}}(\objectStyle{b}, \objectStyle{x}) \land \mymathit{Clear}(\objectStyle{b}) \land \mymathit{Clear}(\objectStyle{y})$ and $\Eff{\actionStyle{move}(\objectStyle{b}, \objectStyle{x}, \objectStyle{y})} = \top \triangleright \{\fluentStyle{\mymathit{On}}(\objectStyle{b}, \objectStyle{y}),$ $\mymathit{Clear}(\objectStyle{x}), \neg \fluentStyle{\mymathit{On}}(\objectStyle{b}, \objectStyle{x}), \neg \mymathit{Clear}(\objectStyle{y})\}$.
        The action set is then $A = \{ \actionStyle{move}(\objectStyle{b}, \objectStyle{x}, \objectStyle{y}) \mid \objectStyle{b} \in B, \objectStyle{x},\objectStyle{y} \in B \cup C, \objectStyle{b} \neq \objectStyle{x},\objectStyle{y} \text{ and } \objectStyle{x} \neq \objectStyle{y} \}$.
        Let $G = \fluentStyle{\mymathit{On}}(\objectStyle{b}_4, \objectStyle{b}_1) \land \fluentStyle{\mymathit{On}}(\objectStyle{b}_3, \objectStyle{b}_2)$ be the goal formula.
        A plan for the classical planning task $(I, A, G)$ is the action sequence:
        $
            \pi = \actionStyle{\mymathit{move}}(\objectStyle{b}_2, \objectStyle{b}_1, \objectStyle{b}_3), \actionStyle{\mymathit{move}}(\objectStyle{b}_4, \objectStyle{c}_3, \objectStyle{b}_1), \actionStyle{\mymathit{move}}(\objectStyle{b}_2, \objectStyle{b}_3, \objectStyle{c}_3), \actionStyle{\mymathit{move}}(\objectStyle{b}_3, \objectStyle{c}_2, \objectStyle{b}_2).
        $
    \end{example}

    So far we have seen the basics of classical planning.
    We are now ready to take the first step towards epistemic planning.
    We begin from \emph{epistemic logic}, where we move from classical states to epistemic states.


    \subsection{Epistemic Logic and Semantics}
    Epistemic logic\footnote{In the narrow sense, the expression \emph{epistemic logic} refers to the logic of knowledge, which differs from \emph{doxastic logic}, that is the logic of belief. Here, we use ``epistemic'' in a generic sense that includes both notions.} \cite{journals/sl/Hintikka1962} is a propositional modal logic that deals with notions of \emph{knowledge} and \emph{belief}.
    In this logic we can express statements such as ``Anne knows that outside is sunny'' and ``Bob believes that Carl believes that he won't pass his exam''.
    To provide a more accessible introduction to epistemic logic, we begin with the single-agent case and we generalise to the multi-agent setting later (Section~\ref{sec:multi-agent}).
    We then further generalise our semantics by introducing \emph{multiple designated worlds} (Section~\ref{sec:multi-pointed}), which allow us to represent \emph{local} epistemic states, \ie epistemic states that are seen from the perspective of some agent, and non-deterministic states.
    We continue the section by discussing \emph{common knowledge} (Section \ref{sec:common-knowledge}), and we conclude by specifying the epistemic language and states that we are later going to employ in the semantics of \textsc{epddl} (Section~\ref{sec:language-states-epddl}).

    Given a finite set $\atomSet$ of atomic propositions, the language $\LangSingle$ of \emph{epistemic logic} is given by the BNF:
    \begin{equation*}
        \phi ::= \fluentStyle{p} \mid \neg \phi \mid \phi \land \phi \mid \B{}\phi
    \end{equation*}

    \noindent where $p \in \atomSet$.
    We read the formula $\B{}\phi$ as ``the agent knows/believes that $\phi$''.
    We also define the dual modal operator $\D{}$ as $\neg \B{}\neg$, and we read $\D{}\phi$ as ``the agent considers $\phi$ to be possible''.
    Note that from the duality of $\B{}$ and $\D{}$ we have that $\B{}\phi$ means that it is not the case that the agent considers $\phi$ not to be possible (since $\B{} \phi \equiv \neg \D{}\neg \phi$), \ie that $\phi$ is true in all scenarios that the agent considers to be possible.
    To make this more precise, we now introduce the semantics of epistemic logic, which is defined in terms of \emph{epistemic states}.
    These can be seen as a generalization of classical planning states, which not only allow to represent the physical aspects of a situation of interest, but also the knowledge/beliefs of the agent.
    In epistemic logic (and also in DEL), epistemic states are defined as pointed Kripke models.


    \begin{definition}[Single-Agent Epistemic Model and State]%
    \label{def:epistemic-state-single-agent}
        An \emph{(epistemic) model of $\LangSingle$} is a triple $M = (W, R, L)$, where:
        \begin{itemize}
            \item $W \neq \varnothing$ is a finite set of \emph{possible worlds};
            \item $R \subseteq W \times W$ is an \emph{accessibility relation}; and
            \item $L: W \rightarrow 2^\atomSet$ assigns to each world a \emph{label}, being a finite set of atoms.
        \end{itemize}
        An \emph{(epistemic) state of $\LangSingle$} is a pair $s = (M, w_d)$, where $M$ is an epistemic model of $\LangSingle$ and $w_d \in W$ is the \emph{designated world}.
    \end{definition}
    
    \noindent We often write $wRv$ for $(w, v) \in R$.
    The pair $(W, R)$ is often called the \emph{frame} of the model.
    A possible world $w$ can be thought of as a classical state, where the label $L(w)$ describes what atoms are true in $w$.
    Each world represents a possible configuration of a state of affairs, and the designated world represents the \emph{actual} configuration, namely the real state of affairs.
    The accessibility relation $R$ tells us what configurations are considered possible by the agent in each world: $w R v$ means that in the state of affairs described by $w$ the agent considers $v$ to be possible.
    Note that accessibility relations represent the uncertainty of the agent: the more edges in a state, the more configurations the agent considers to be possible, and thus the more uncertainty the agent has.

    \begin{definition}[Truth in a Single-Agent Epistemic Model]%
    \label{def:truth-single-agent}
        Let $M = (W, R, L)$ be an epistemic model of $\LangSingle$ and let $w \in W$.
        We define truth by induction on the structure of formulas of $\LangSingle$:
        \begin{equation*}
            \begin{array}{lll}
                (M, w) \models \fluentStyle{p} & \textnormal{iff} & \fluentStyle{p} \in L(w) \\
                (M, w) \models \neg \phi       & \textnormal{iff} & (M, w) \not\models \phi \\
                (M, w) \models \phi \land \psi & \textnormal{iff} & (M, w) \models \phi \textnormal{ and } (M, w) \models \psi \\
                (M, w) \models \B{}\phi        & \textnormal{iff} & \textnormal{for all } v \in W \textnormal{ if } wRv, \textnormal{ then } (M, v) \models \phi
            \end{array}
        \end{equation*}
    \end{definition}
    
    A propositional atom is true in a world if it is contained in the label of that world.
    The cases of propositional connectives are standard.
    A formula $\B{}\phi$ holds in $w$ iff $\phi$ holds in all worlds $v$ that are accessible from $w$.
    In other words, in the state of affairs described by $w$ the agent knows/believes that $\phi$ holds iff $\phi$ is true in all worlds that the agent considers to be possible.
    Note that it easily follows from Definition~\ref{def:truth-single-agent} that $(M, w) \models \D{}\phi$ iff there exists a world $v$ such that $w R v$ and $(M, v) \models \phi$, \ie if the agent considers possible a world $v$ where $\phi$ holds.
    We say that a formula $\phi \in \LangSingle$ is \emph{satisfiable} if there exists an epistemic state $s$ such that $s \models \phi$.
    We say that $\phi \in \LangSingle$ is \emph{valid}, denoted $\models \phi$, if it holds in all states of $\LangSingle$.

    A formula is called \emph{propositional} if no modal operators appear in it.
    Otherwise, the formula is called \emph{epistemic}, if we interpret $\B{}$ as a knowledge modality, \emph{doxastic}, if we interpret it as a belief modality, or \emph{modal}, if we do not make any assumption on the kind of modality.
    Note that, by the first three items of Definition~\ref{def:truth-single-agent}, to check whether a propositional formula $\phi$ holds in $(M, w)$ we only need to perform membership checks on the set $L(w)$ of the propositional atoms occurring in $\phi$.
    On the other hand, by the last item of Definition~\ref{def:truth-single-agent}, to check the truth of modal formulas we also need to account for the possible worlds that are accessible (or reachable) from $w$.
    In general, checking whether nested modal formulas hold in $(M, w)$ requires us to visit different parts of the model starting from world $w$.
    For instance, to see whether $(M, w) \models \B{}(p \land \D{}(\neg q \lor \B{}r))$, we have to check that all worlds accessible from $w$ satisfy $p$, and that, in all such worlds, we can reach in one step a world, call it $x$, in which $q$ does not hold, or $r$ holds in all worlds accessible from $x$.

    

    \begin{figure}
        \centering
        \begin{tikzpicture}[node distance=32pt and 64pt, >=stealth']
    \node[graphicaldworld, scale=0.6, label=below:$w_1$] (w1)                                          {\input{img/examples/single-agent-bw/single-agent-bw-1.tex}};
    \node[graphicalworld,  scale=0.6, label=below:$w_2$] (w2) [right=of w1]                            {\input{img/examples/single-agent-bw/single-agent-bw-2.tex}};
    \node[graphicalworld,  scale=0.6, label=below:$w_3$] (w3) [above right=64pt and 32pt of w1.center] {\input{img/examples/single-agent-bw/single-agent-bw-3.tex}};

    \path
        (w1) edge[<->, thick] (w2)
        (w2) edge[<->, thick] (w3)
        (w3) edge[<->, thick] (w1)
        (w1) edge[->,  thick, loop, out=170, in=190, looseness=4] (w1)
        (w2) edge[->,  thick, loop, out= 10, in=-10, looseness=4] (w2)
        (w3) edge[->,  thick, loop, out=105, in= 75, looseness=4] (w3)
    ;
\end{tikzpicture}
        \caption{Epistemic state of Example \ref{ex:single-agent-bw}. Rectangles represent worlds, labelled by their name (the actual world has a double line). Edges denote the accessibility relation. We represent labels graphically, similarly to Example~\ref{ex:classical-bw}, \eg $L(w_1) = \{ \mymathit{On}(\objectStyle{b}_1, \objectStyle{c}_1), \mymathit{On}(\objectStyle{b}_2, \objectStyle{b}_1), \mymathit{On}(\objectStyle{b}_3, \objectStyle{c}_2), \mymathit{On}(\objectStyle{b}_4, \objectStyle{c}_3) \}$.}
        \label{fig:single-agent-bw}
    \end{figure}


    \begin{example}[Single agent Epistemic Blocks World]\label{ex:single-agent-bw}
        In the classical Blocks World problem (Example~\ref{ex:classical-bw}) we implicitly assumed that our agent (the mechanical arm) has no uncertainty about the world, \ie that it knows at each time the correct position of all blocks.
        We now lift this assumption and consider a slight variation of our problem where the agent observes the situation via a camera that is placed directly above the table.
        From this perspective, the agent can only see the blocks that are on the top of the columns ($\objectStyle{b}_2$, $\objectStyle{b}_3$, and $\objectStyle{b}_4$). We also assume that the agent knows that there are four blocks in total.
        Figure \ref{fig:single-agent-bw} shows the epistemic state $s$ of this scenario.
        Since the agent realises that it sees only three blocks out of four, it considers three situations to be possible: either the missing block $\objectStyle{b}_1$ is under block $\objectStyle{b}_2$ (world $w_1$), block $\objectStyle{b}_3$ (world $w_2$), or block $\objectStyle{b}_4$ (world $w_3$).
        We assume that block $b_1$ is actually under $b_2$, so $w_1$ is the designated world.
        Moreover, in each world the agent considers all worlds to be possible, as it has no information about the actual position of the missing block.
        Note that state $s$ correctly captures the fact that the agent does not know the position of block $\objectStyle{b}_1$, \ie $s \models \neg \B{} \mymathit{On}(\objectStyle{b}_2, \objectStyle{b}_1) \land \neg \B{} \mymathit{On}(\objectStyle{b}_3, \objectStyle{b}_1) \land \neg \B{} \mymathit{On}(\objectStyle{b}_4, \objectStyle{b}_1)$.

    \end{example}

    Depending on the desired setting, an epistemic state can be used to represent either the \emph{knowledge} or the \emph{beliefs} of the agent.
    But what does it mean to know that something is the case, and how is this different from merely believing that it is the case?
    These questions, and many others connected to them, are as old as philosophy itself, so we are not going to attempt to provide an exhaustive account.
    Instead, we briefly describe how these two concepts are typically formalised in epistemic logic, which is our setting of interest.
    The modern account of the notions of knowledge and belief stems from the seminal work of Jaakko Hintikka, who characterised the two notions by the formal properties they should satisfy \cite{journals/sl/Hintikka1962}.
    Today, such properties of knowledge and belief are typically expressed via \emph{axiom schemata}, which are formulas that an epistemic state is required to satisfy in order for it to be considered a ``knowledge state'', or a ``belief state''.
    The axioms are the following:
    \begin{equation*}
        \begin{array}{llr}
            K & \B{} (\phi \rightarrow \psi) \rightarrow
                    (\B{} \phi \rightarrow \B{} \psi)          & \textnormal{(Distribution)}           \\
            T & \B{} \phi \rightarrow \phi                     & \textnormal{(Truth)}                  \\
            D & \B{} \phi \rightarrow \D{}\phi                 & \textnormal{(Consistency)}            \\
            4 & \B{} \phi \rightarrow \B{} \B{} \phi           & \textnormal{(Positive introspection)} \\
            5 & \neg \B{} \phi \rightarrow \B{} \neg \B{} \phi & \textnormal{(Negative introspection)}
        \end{array}
    \end{equation*}

    Axiom $K$ states that if the agent knows/believes an implication and its antecedent, then it also knows/believes the consequent.
    Axiom $T$ states that if the agent knows that $\phi$, then $\phi$ must be true.
    This principle describes a property that is only of knowledge: it is possible to have a \emph{false belief}, that is to believe to be true something that is not the case, but it is not possible to have ``false knowledge''.
    Axiom $D$ states that if the agent knows that $\phi$, then $\phi$ must be possible for the agent.
    This defines a principle of consistency in the knowledge/beliefs of the agent.
    Finally, the introspection axioms $4$ and $5$ state that the agent knows/believes what it knows/believes, and what it does not know/believe, respectively.

    Axioms $K$, $T$, $4$, and $5$ are generally considered to describe the properties of knowledge, and together they constitute the axioms of the \emph{logic S5}.\footnote{More precisely, axioms $K$, $T$, and $5$ are sufficient to characterise the S5 logic, as axiom $4$ can be derived from them. We explicitly consider it to simplify the exposition.}
    Similarly, axioms $K$, $D$, $4$, and $5$ are generally considered to describe the properties of belief, and together they constitute the axioms of the \emph{logic KD45}.
    Note that the only difference between the two logics is axioms $T$ and $D$: for something to be known, we require that it must be true, while for something to be believed, we only require that it is consistent with what the agent considers to be possible.
    For a logic $L$, we say that an epistemic model/state is an $L$-model/state, if it satisfies all axioms of that logic.

    Axioms $T$, $D$, $4$ and $5$ notoriously correspond to the following \emph{frame properties}, meaning that if an axiom holds in a state, we are guaranteed that its frame will satisfy the corresponding property:
    \begin{equation*}
        \begin{array}{llr}
            T & \forall w (wRw)                                 & \textnormal{(Reflexivity)}   \\
            D & \forall w \exists x (wRx)                       & \textnormal{(Seriality)}     \\
            4 & \forall w, x, y (wRx \land xRy \rightarrow wRy) & \textnormal{(Transitivity)}  \\
            5 & \forall w, x, y (wRx \land wRy \rightarrow xRy) & \textnormal{(Euclideanness)}
        \end{array}
    \end{equation*}

    Since reflexivity and Euclideanness together imply symmetry, the accessibility relation of an S5-state is an \emph{equivalence relation}.
    Conversely, in KD45-states symmetry does not hold in general, as seriality is a weaker condition than reflexivity, \ie a reflexive accessibility relation is always serial, but a serial one is not always reflexive.
    As a consequence, all S5-models are also KD45-models, but not all KD45-models are S5-models.

    The correspondence between modal axioms and frame properties is quite useful to check whether an epistemic state is an S5-state, or a KD45-state (or neither).
    For instance, the epistemic state in Figure \ref{fig:single-agent-bw} is an S5-state, as the accessibility relation of the agent is an equivalence relation, and thus it is also a KD45-state.
    Therefore, the state can be used to represent both the knowledge and the beliefs of the agent.

    In the next section, we are going to move from a single-agent setting to a multi-agent one.
    In this new setting, the axioms of knowledge and belief and their corresponding frame properties are defined in the same way for each agent, and the multi-agent logics of knowledge and belief are denoted as S5$_n$ and KD45$_n$, respectively.

    \subsubsection{Introducing Multiple Agents}\label{sec:multi-agent}
        In \textsc{epddl}, we wish to talk about the knowledge/beliefs of multiple agents.
        We thus need to generalise both the logical language and epistemic models and states to a multi-agent setting.
        Let $\agentSet = \{1, \dots, \agentStyle{n}\}$ be a finite set of agents. We use $\agentStyle{i}, \agentStyle{j}, \agentStyle{k}, \dots$ to denote elements of $\agentSet$. The language $\Lang$ of \emph{multi-agent epistemic logic} is given by the BNF:
        \begin{equation*}
            \phi ::= \objectStyle{p} \mid \neg \phi \mid \phi \land \phi \mid \B{i}\phi
        \end{equation*}

        \noindent where $\objectStyle{p} \in \atomSet$, $\agentStyle{i} \in \agentSet$ and the formula $\B{i}\phi$ is read as ``agent $\agentStyle{i}$ knows/believes that $\phi$''.
        As for the single agent case, we define the dual modal operators $\D{i}\phi = \neg \B{i}\neg\phi$ for each agent $\agentStyle{i}$.
        We say that \emph{agent $\agentStyle{i}$ knows whether $\phi$}, denoted $\Kw{i} \phi$, iff $\B{i} \phi \lor \B{i} \neg \phi$.
        Namely, agent $i$ knows whether $\phi$ holds iff the agent knows that the formula holds, or if it knows that it does not hold.
        The dual modality is $\Kwd{i} \phi = \neg \Kw{i} \neg \phi$.
        Note that we have
        $\Kwd{i} \phi
            = \neg \Kw{i} \neg \phi
            = \neg (\B{i} \neg \phi \lor \B{i} \neg \neg \phi)
            = \neg (\B{i} \neg \phi \lor \B{i} \phi)
            = \neg \Kw{i} \phi$.
        As a consequence, we can read $\Kwd{i} \phi$ as ``agent $\agentStyle{i}$ does not know whether $\phi$''.
        
        Multi-agent epistemic models generalise single-agent ones (Definition~\ref{def:epistemic-state-single-agent}) by assigning an accessibility relation to each agent.

        \begin{definition}[Multi-Agent Epistemic Model and State]%
        \label{def:epistemic-state}
            An \emph{(epistemic) model of $\Lang$} is a triple $M = (W, R, L)$ such that $W$ and $L$ are as in Definition~\ref{def:epistemic-state-single-agent} and $R: \agentSet \rightarrow 2^{W \times W}$ assigns to each agent $\agentStyle{i}$ an \emph{accessibility relation} $R_\agentStyle{i}$.
            An \emph{(epistemic) state of $\Lang$} is a pair $s = (M, w_d)$, where $M$ is an epistemic model of $\Lang$ and $w_d \in W$ is the \emph{designated world}.
        \end{definition}

        \noindent We often write $w R_\agentStyle{i} v$ for $(w, v) \in R_\agentStyle{i}$.
        For a world $w$ we let $R_\agentStyle{i}(w) = \{ v \in W \mid w R_\agentStyle{i} v \}$ be the set of possible worlds accessible by $\agentStyle{i}$ from $w$.
        Multi-agent epistemic states generalise single-agent ones (Definition~\ref{def:epistemic-state-single-agent}) in a natural way, where each agent is given an accessibility relation that describes the worlds that the agent considers to be possible.
        Truth in epistemic states of $\Lang$ is obtained by adapting the case of modal formulas in Definition~\ref{def:truth-single-agent}:

        \begin{definition}[Truth in a Multi-Agent Epistemic Model]%
        \label{def:truth-multi-agent}
            Let $M = (W, R, L)$ be an epistemic model of $\Lang$ and let $w \in W$.
            The cases of atoms and propositional connectives are as in Definition~\ref{def:epistemic-state-single-agent}, while the case of modal formulas is as follows:
            \begin{equation*}
                \begin{array}{lll}
                    (M, w) \models \B{i}\phi       & \textnormal{iff} & \textnormal{for all } v \in W, \textnormal{ if } w R_\agentStyle{i} v \textnormal{ then } (M, v) \models \phi
                \end{array}
            \end{equation*}
        \end{definition}

        \begin{figure}
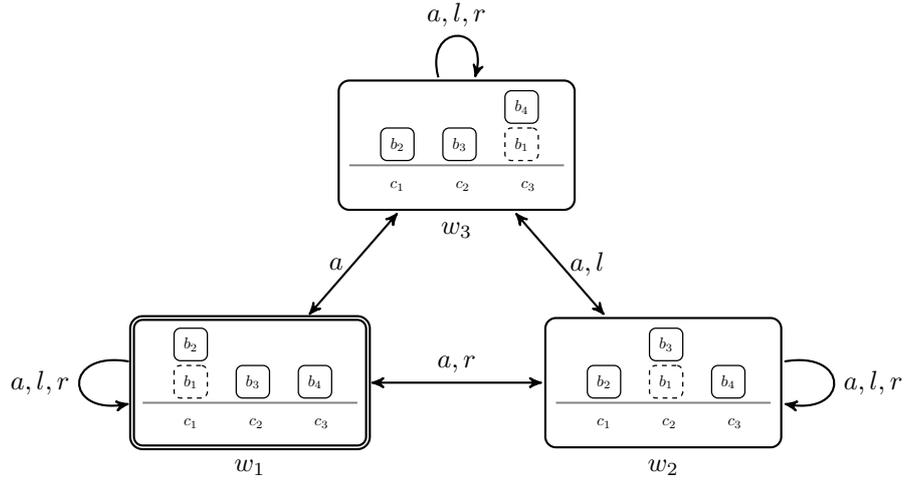

            \centering
            \begin{tikzpicture}[node distance=32pt and 64pt, >=stealth']
    \node[graphicaldworld, scale=0.6, label=below:$w_1$] (w1)                                          {\input{img/examples/single-agent-bw/single-agent-bw-1.tex}};
    \node[graphicalworld,  scale=0.6, label=below:$w_2$] (w2) [right=of w1]                            {\input{img/examples/single-agent-bw/single-agent-bw-2.tex}};
    \node[graphicalworld,  scale=0.6, label=below:$w_3$] (w3) [above right=64pt and 32pt of w1.center] {\input{img/examples/single-agent-bw/single-agent-bw-3.tex}};

    \path
        (w1) edge[<->, thick] node[above] {$\agentStyle{a},\agentStyle{r}$} (w2)
        (w2) edge[<->, thick] node[right] {$\agentStyle{a},\agentStyle{l}$} (w3)
        (w3) edge[<->, thick] node[left]  {$\agentStyle{a}$}   (w1)
        (w1) edge[->,  thick, loop, out=170, in=190, looseness=4] node[left]  {$\agentStyle{a},\agentStyle{l},\agentStyle{r}$} (w1)
        (w2) edge[->,  thick, loop, out= 10, in=-10, looseness=4] node[right] {$\agentStyle{a},\agentStyle{l},\agentStyle{r}$} (w2)
        (w3) edge[->,  thick, loop, out=105, in= 75, looseness=4] node[above] {$\agentStyle{a},\agentStyle{l},\agentStyle{r}$} (w3)
    ;
\end{tikzpicture}
            \caption{Epistemic state of Example \ref{ex:multi-agent-bw}. Since we now have multiple agents, each edge is labeled by the corresponding agent(s).}
            \label{fig:multi-agent-bw}
        \end{figure}

        \begin{example}[Multi-agent Epistemic Blocks World]\label{ex:multi-agent-bw}
            Consider the situation of Example \ref{ex:single-agent-bw} and let $\agentStyle{a}$ be the agent that can only see from above as described there.
            We now introduce two new agents, $\agentStyle{l}$ and $\agentStyle{r}$, that observe the situation via cameras placed in a top left and from a top right position wrt.\ to the table, respectively.
            In this setting agent $\agentStyle{l}$ can only see the blocks in the left column, plus the top block of each column (that is, all blocks) and, similarly, $\agentStyle{r}$ sees the right column and the top blocks (\ie blocks $\objectStyle{b}_2$, $\objectStyle{b}_3$ and $\objectStyle{b}_4$).
            We assume that all agents know about everybody's position.
            This situation is shown in the epistemic state $s$ of Figure \ref{fig:multi-agent-bw}.
            The worlds, designated world, and labels of $s$ are the same as the previous example.
            The perspective of each agent is modeled as follows.
            The accessibility relations of agent $\agentStyle{a}$ are as in Example~\ref{ex:single-agent-bw}.
            Agent $\agentStyle{l}$ sees all blocks, so from the actual world $w_1$ the agent only considers $w_1$ to be possible, \ie $w_1 R_\agentStyle{l} w_1$.
            At first glance, it might seem that this is enough to represent the beliefs of agent $\agentStyle{l}$.
            However, we also need to model the point of view of $\agentStyle{l}$ from the perspectives of agents $\agentStyle{a}$ and $\agentStyle{r}$.
            For all they know, in fact, there are two possibilities: either $b_1$ is under $b_2$ and thus agent $\agentStyle{l}$ sees all blocks, or $b_1$ is somewhere else, implying that $l$ does not see all blocks.
            We already handled the former case, so we now look at the latter.
            If $\agentStyle{l}$ does not see all blocks, then from its perspective block $\objectStyle{b}_1$ can either be under $\objectStyle{b}_3$ (world $w_2$) or $\objectStyle{b}_4$ (world $w_3$).
            Since agent $\agentStyle{l}$ does not know which is the case (under this hypothesis), the agent should consider both situations to be possible, so we have $w_i R_\agentStyle{l} w_j$ for all $i,j \in \{2,3\}$.
            Finally, the perspective of agent $\agentStyle{r}$ is modeled symmetrically to agent $\agentStyle{l}$: either the agent sees all blocks (world $w_3$), or it does not (worlds $w_1$ and $w_2$).
            Note that state $s$ correctly captures the fact that agent $\agentStyle{l}$ knows that agents $\agentStyle{a}$ and $\agentStyle{r}$ do not know where the missing block $\objectStyle{b}_1$ is: \ie $s \models \B{l} \bigwedge_{\agentStyle{i} = \agentStyle{a},\agentStyle{r}} (\neg \B{i} \mymathit{On}(\objectStyle{b}_2, \objectStyle{b}_1) \land \neg \B{i} \mymathit{On}(\objectStyle{b}_3, \objectStyle{b}_1) \land \neg \B{i} \mymathit{On}(\objectStyle{b}_4, \objectStyle{b}_1))$.

            Finally, we note that state $s$ is an S5$_n$-state, as the accessibility relations of all agents are equivalence relations.
        \end{example}

    \subsubsection{Introducing Multiple Designated Worlds}\label{sec:multi-pointed}
        So far, we have considered single-pointed epistemic models, namely epistemic states with a single designated world.
        This allowed us to model situations where we knew which world was the actual one.
        In other words, single-pointed epistemic models represent scenarios from the perspective of an omniscient and external observer.
        We can, however, do much more, like representing the internal perspective of one or more agents, and modeling non-determinism.
        This requires multiple designated worlds, as the next definition shows.

        \begin{definition}[Multi-Pointed Epistemic Model]%
        \label{def:multi-pointed-epistemic-state}
            An \emph{(epistemic) state}, also called a \emph{multi-pointed (epistemic) model}, of $\Lang$ is a pair $(M, W_d)$, where $M$ is as in Definition~\ref{def:epistemic-state} and $W_d \subseteq W$ is a non-empty set of \emph{designated worlds}.
        \end{definition}

        A formula of $\Lang$ holds in a multi-pointed model iff it holds in all designated worlds:

        \begin{definition}[Truth in a Multi-Pointed Epistemic Model]%
        \label{def:truth}
            %
            Let $(M, W_d)$ be a multi-pointed epistemic model of $\Lang$.
            \begin{equation*}
                \begin{array}{lll}
                    (M, W_d) \models \phi & \textnormal{iff} & \textnormal{for all } w \in W_d, (M, w) \models \phi
                \end{array}
            \end{equation*}
        \end{definition}

        As mentioned above, with multi-pointed models we can represent more sophisticated kinds of situations.
        We first describe \emph{local epistemic states}~\cite{journals/jancl/Bolander2011}.

        \begin{definition}[Local Epistemic State]%
        \label{def:local-epistemic-state}
            An epistemic state $s_i = ((W, R, L), W_d)$ of $\Lang$ is called a \emph{local (epistemic) state} for an agent $\agentStyle{i} \in \agentSet$, if $W_d$ is closed under $R_\agentStyle{i}$, \ie if for all $w \in W_d$ we have $R_\agentStyle{i}(w) \subseteq W_d$.
            A \emph{local (epistemic) state} is any epistemic state that is local for some agent.
        \end{definition}
        
        \noindent
        A single-pointed epistemic model is also called a \emph{global state}~\cite{journals/jancl/Bolander2011}, and it represents a situation from the perspective of some external and omniscient observer.
        Instead, a local state $s_i$ for agent $\agentStyle{i}$ allows us to take the internal perspective of some agent: The designated worlds of $s_i$ are precisely those worlds that the agent considers to be possible.
        Hence, multi-pointed models allow us to widen the spectrum of points of view we can describe, and provide a more general representation than single-pointed models.

        \begin{example}[Local Epistemic State]\label{ex:multi-agent-bw-local}
            The state $s = (M, W_d)$ shown in Figure~\ref{fig:multi-agent-bw} is a global state, where we assumed that the real configuration of the blocks is the one in world $w_1$.
            Let us now instead take the perspective of an agent involved in the situation, say agent $\agentStyle{r}$.
            This can be represented by the local state $s_\agentStyle{r} = (M, \{ w_1, w_2 \})$ for $\agentStyle{r}$, which only differs from $s$ in the choice of designated worlds.
            In fact, since we assume that from its top-right position agent $\agentStyle{r}$ sees that block $b_1$ is not under $\objectStyle{b}_4$, then $\agentStyle{r}$ considers two possibilities: either $\objectStyle{b}_1$ is under $\objectStyle{b}_2$ (world $w_1$) or $\objectStyle{b}_3$ (world $w_2$).
            Hence both $w_1$ and $w_2$ are designated worlds of $s_\agentStyle{r}$.
            Note that the formula $\psi = \B{\agentStyle{l}} \mymathit{On}(\objectStyle{b}_2, \objectStyle{b}_1)$ is true in $s$, as in Example~\ref{ex:multi-agent-bw} we assumed that agent $\agentStyle{l}$ sees all blocks.
            However, since $(M, w_2) \not\models \psi$, by Definition~\ref{def:truth} we have $s_\agentStyle{r} \not\models \psi$: agent $\agentStyle{r}$ also considers possible that agent $\agentStyle{l}$ does not know the position of the missing block.
        \end{example}

        On top of the internal perspectives of agents, multi-pointed epistemic models can also be employed to represent non-determinism, as we now show.
        The \emph{disjoint union} of two states $s = ((W, R, L), W_d)$ and $s' = ((W', R', L'), W_d')$ is the epistemic state $s \sqcup s' = ((W \sqcup W', R \sqcup R', L \sqcup L'), W_d \sqcup W_d')$, where $\sqcup$ denotes the disjoint union of two sets.
        Non-deterministic states are then defined as follows.

        \begin{definition}[Non-Deterministic Epistemic State]%
        \label{def:non-det-state}
            An epistemic state $s$ of $\Lang$ is called a \emph{non-deterministic (epistemic) state} if $s$ is the disjoint union of two epistemic states $t$, $t'$ for which there exists a formula $\phi \in \Lang$ such that $t \models \phi$ iff $t' \not\models \phi$\footnote{For readers familiar with modal logic, it is not hard to see that this is equivalent to requiring that $t$ and $t'$ are not \emph{bisimilar}, as both states are assumed to be finite~\cite{book/cup/Blackburn2001}.}.
            Otherwise, $s$ is called \emph{deterministic}.
        \end{definition}

        \noindent
        A non-deterministic state can be thought of as a state that simultaneously represents multiple states of affairs at once.
        To ensure that the union of states is non-trivial, we require that each state model a distinct situation from the other.
        For instance, we do not consider $s' = s \sqcup s$ to be a non-deterministic state, as the two disjoint components of $s'$ represent the same scenario.

        \begin{figure}
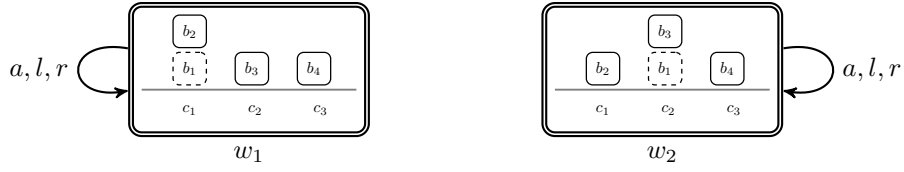

            \centering
            \begin{tikzpicture}[node distance=32pt and 64pt, >=stealth']
    \node[graphicaldworld, scale=0.6, label=below:$w_1$] (w1)                                          {\input{img/examples/single-agent-bw/single-agent-bw-1.tex}};
    \node[graphicaldworld, scale=0.6, label=below:$w_2$] (w2) [right=of w1]                            {\input{img/examples/single-agent-bw/single-agent-bw-2.tex}};

    \path
        (w1) edge[->,  thick, loop, out=170, in=190, looseness=4] node[left]  {$\agentStyle{a},\agentStyle{l},\agentStyle{r}$} (w1)
        (w2) edge[->,  thick, loop, out= 10, in=-10, looseness=4] node[right] {$\agentStyle{a},\agentStyle{l},\agentStyle{r}$} (w2)
    ;
\end{tikzpicture}
            \caption{Non-deterministic epistemic state of Example \ref{ex:non-deterministic-state}.}
            \label{fig:non-deterministic-state}
        \end{figure}

        \begin{example}[Non-Deterministic Epistemic State]%
        \label{ex:non-deterministic-state}
            Let $s$ be the epistemic state of Figure~\ref{fig:non-deterministic-state}.
            It is not hard to check that $s$ is non-deterministic, as it can be defined as the disjoint union of the sub-model rooted in $w_1$ with the one rooted in $w_2$.
            The state represents a situation where agents distinguish between two cases: one where block $b_1$ is under $b_2$ (world $w_1$), and one where it is under $b_3$ (world $w_2$).
            Note that $s$ does \emph{not} represent a situation of uncertainty.
            In fact, it is not hard to see that $s \models \bigwedge_{i \in \agentSet} \B{i} \mymathit{On}(b_2, b_1) \lor \bigwedge_{i \in \agentSet} \B{i} \mymathit{On}(b_3, b_1)$, \ie everybody knows that $b_1$ is under $b_2$, or everybody knows that $b_1$ is under $b_3$.

            Compare this with the epistemic state $s'$, obtained from $s$ by adding the world-pairs $(w_1, w_2),$ and $(w_2, w_1)$ to all accessibility relations.
            In this state, the agents can not distinguish between the two worlds (different from the case of state $s$), making them uncertain about the position of block $b_1$, \ie $s' \models \bigwedge_{i \in \agentSet} \D{i} \mymathit{On}(b_2, b_1) \land \bigwedge_{i \in \agentSet} \D{i} \mymathit{On}(b_3, b_1)$.
        \end{example}

    \subsubsection{Group Knowledge and Common Knowledge}\label{sec:common-knowledge}
        In many scenarios it is often of interest to talk about the knowledge/beliefs of a group of agents.
        For a non-empty set $G \subseteq \agentSet$ of agents, we say that \emph{everybody in $G$ knows/believes that $\phi$}, denoted $\B{G} \phi$,\footnote{In the literature, a more common notation for this kind of formulas is $E_G \phi$. We here choose $\B{G} \phi$ as it is closer to the \textsc{epddl} syntax of modal formulas (see Section~\ref{sec:syntax-common-formulas}).} iff $\bigwedge_{i \in G} \B{i}\phi$, \ie
        if all agents in the group independently know/believe that $\phi$ is true.
        We also define the dual modality $\D{G} \phi = \neg \B{G} \neg \phi$.
        Note that $\D{G} \phi = \neg \B{G} \neg \phi = \neg \bigwedge_{i \in G} \B{i} \neg\phi = \bigvee_{i \in G} \neg \B{i} \neg\phi = \bigvee_{i \in G} \D{i} \phi$.
        We define the group knowing-whether modalities, $\Kw{G}$ and $\Kwd{G}$, in an analogous way and read $\Kw{G} \phi$/$\Kwd{G} \phi$ as ``everybody in $G$ knows/doesn't know whether $\phi$''.

        \emph{Common knowledge/belief}~\cite{books/wb/Lewis1969,journals/as/Aumann1976,journlas/td/LismontM1994} requires a much stronger set of conditions.
        A formula $\phi$ is commonly known/believed by the agents in $G$ iff everybody in $G$ knows/believes that $\phi$, everybody in $G$ knows/believes that everybody in $G$ knows/believes that $\phi$, and so on \emph{ad infinitum}. More precisely:
        \begin{equation*}
            \CK{G} \phi = \bigwedge_{k > 0} \B{G}^k \phi,
        \end{equation*}

        \noindent where $\triangle_G^1 \phi = \triangle_G \phi$ and, for each $k \geq 1$, $\triangle_G^{k+1} \phi = \triangle_G \triangle_G^k \phi$ (with $\triangle \in \{\B{}, \D{} \}$).
        We read $\CK{G}\phi$ as ``the agents in $G$ have common knowledge/belief that $\phi$''.
        We also define the dual modality $\CKd{G} \phi = \neg \CK{G} \neg \phi = \bigvee_{k > 0} \D{G}^k \phi$.
        The language $\Lang[C]$ of \emph{multi-agent epistemic logic with common knowledge/belief} is obtained by augmenting $\Lang$ with the group modalities $\CK{G}\phi$ for all non-empty sets $G \subseteq \agentSet$ of agents.
        Epistemic models and states of $\Lang[C]$ are defined as in Definition~\ref{def:epistemic-state}.
        Truth in epistemic models of $\Lang[C]$ is defined as in Definition~\ref{def:truth}, plus the following case:
        %
        \begin{equation*}
            \begin{array}{lll}
                (M, w) \models \CK{G}\phi & \textnormal{iff} & \textnormal{for all } v \in W, \textnormal{ if } w R^+_G v \textnormal{ then } (M, v) \models \phi
            \end{array}
        \end{equation*}

        \noindent where $R_G = \bigcup_{i \in G} R_\agentStyle{i}$ and $R^+_G$ denotes the its \emph{transitive closure}.
        In world $w$ agents in $G$ have common knowledge/belief that $\phi$ iff $\phi$ holds in all worlds that are reachable from $w$ via the accessibility relations of agents in $G$.

        \begin{example}[Common Knowledge]\label{ex:ck}
            Let $s$ be the epistemic state of Example \ref{ex:multi-agent-bw}. The formula $\phi = \mymathit{Clear}(\objectStyle{b}_2) \land \mymathit{Clear}(\objectStyle{b}_3) \land \mymathit{Clear}(\objectStyle{b}_4)$ is commonly known by all agents, \ie $s \models \CK{\{\agentStyle{a}, \agentStyle{l}, \agentStyle{r}\}} \phi$, since in all worlds $\phi$ is true. However, the formula $\psi = \mymathit{On}(\objectStyle{b}_1, \objectStyle{c}_1) \land \mymathit{On}(\objectStyle{b}_2, \objectStyle{b}_1)$ is not common knowledge among all agents, \ie $s \not\models \CK{\{a, \agentStyle{l}, \agentStyle{r}\}} \psi$, since $(M, w_2) \not\models \psi$ and $(M, w_3) \not\models \psi$ ($\psi$ is only known by agent $\agentStyle{l}$).
        \end{example}

    \subsubsection{Epistemic Language and States of \textnormal{\bf\textsc{epddl}}}\label{sec:language-states-epddl}
        So far, we have introduced several kinds of modal operators and various logics and semantics based on different combinations of such operators.
        The most expressive logic we consider is the logic $\Lang[C]$, with which we can express the following:
        \begin{inparaenum}
            \item Knowledge/beliefs of individual agents;
            \item Knowledge/beliefs of groups of agents;
            \item Situations where an agent or a group of agents knows whether something is the case, or if it is not; and
            \item Common knowledge/belief of a group of agents.
        \end{inparaenum}
        In the literature, there exist other kinds of modalities, such as \emph{distributed knowledge}~\cite{journals/jacm/HalpernM90}, which we did not cover here, as we chose to introduce the modalities that are more commonly used by epistemic planners.
        We believe $\Lang[C]$ is a balanced logical language upon which to base the semantics of the first version of \textsc{epddl}.
        From now on, when we talk about an epistemic state, we implicitly assume that it is a (multi-pointed) epistemic state of $\Lang[C]$.
        In the future, we plan on considering other kinds of modalities to further increase the expressivity of the language.

    \subsection{Epistemic Actions and Product Update}\label{sec:epistemic-actions}
    Epistemic actions represent information change in dynamic epistemic logic, where by information we both mean ``physical'' properties of the worlds and the knowledge/beliefs of one or more agents.
In DEL, epistemic actions are formalised by \emph{event models}~\cite{baltagAL/1998/Logic,unpublished/cwi/vanEijck2004,benthemJJ/2006/Logics} (also called \emph{update models} or \emph{action models} in the literature).
Roughly speaking, an event model is to a classical action what an epistemic model is to a classical state.
Much like an epistemic model contains a set of worlds that represent different possible configurations of a state of affairs, an event model contains a set of \emph{events}, that instead represent different possible outcomes of an action, describing different perspectives of the agents on the action.
Intuitively, an event can be thought of as representing a classical action, as each event is associated with a precondition, being an epistemic formula, and a set of effects, here called \emph{postconditions}.
As in epistemic models, events are linked via accessibility relations that describe the perspective of the agents on the action.

Epistemic actions generalise classical actions in the same way epistemic states generalise classical states.
In this section, we thus directly provide the general definition of event model (Definition~\ref{def:event-model}), and we do not repeat the incremental construction from the classical to the epistemic setting.
We then introduce the \emph{product update} operator (Definition~\ref{def:product-update}), which describes the effects of applying an epistemic action to a state.
We conclude the section with an analysis of different types of epistemic actions, and by showcasing their many features with a series of examples from the Epistemic Blocks World problem (Section~\ref{sec:epistemic-action-types}).

\begin{definition}[Event Models and Epistemic Actions]%
\label{def:event-model}
    An \emph{event model} of $\Lang[C]$ is a quadruple $A = (E, Q, \mymathit{pre}, \mymathit{post})$, where:
    \begin{itemize}
        \item $E \neq \varnothing$ is a finite set of events;
        \item $Q: \agentSet \rightarrow 2^{E \times E}$ assigns to each agent $\agentStyle{i}$ an \emph{accessibility relation} $Q_\agentStyle{i}$;
        \item $\mymathit{pre}: E \rightarrow \Lang[C]$ assigns to each event a \emph{precondition}; 
        \item $\mymathit{post}: E \times \atomSet \rightarrow \Lang[C]$ assigns to each event-atom pair a \emph{postcondition}.
    \end{itemize}
    An \emph{(epistemic) action}, also called a \emph{multi-pointed event model}, of $\Lang[C]$ is a pair $\actionStyle{a} = (A, E_d)$, where $A$ is an event model of $\Lang[C]$ and $E_d \subseteq E$ is a non-empty set of \emph{designated events}.
\end{definition}

\noindent We often write $e Q_\agentStyle{i} f$ for $(e, f) \in Q_\agentStyle{i}$.
An event $e$ represents a possible outcome of the action, where the precondition $\mymathit{pre}(e)$ specifies when $e$ is applicable and the postcondition $\mymathit{post}(e, \fluentStyle{p})$ tells us how the event modifies the value of $\fluentStyle{p}$ (\ie the effects of $e$).
The accessibility relations specify the events that are considered to be possible by the agents, and thus their perspectives on the action.
We say that an event $e$ has a \emph{trivial precondition} if $\models \mymathit{pre}(e) \leftrightarrow \top$ (\ie if it is always applicable), and that it has \emph{trivial postconditions} if for all $\fluentStyle{p} \in \atomSet$ we have $\models \mymathit{post}(e, \fluentStyle{p}) \leftrightarrow p$ (\ie if it never changes the truth value of atoms).
An event is \emph{trivial} if both its pre- and postconditions are trivial.
Similarly to global and local states, we can both represent actions from the (\emph{global}) perspective of an external and omniscient observer, and from the (\emph{local}) perspective of some agent.

\begin{definition}[Global and Local Epistemic Actions~\cite{journals/jancl/Bolander2011}]%
\label{def:global-local-action}
    An epistemic action $\actionStyle{a} = ((E, Q, \mymathit{pre},$ $\mymathit{post}), E_d)$ of $\Lang[C]$ is called \emph{global} if $E_d$ is a singleton.
    It is called a \emph{local (epistemic) action} for an agent $i \in \agentSet$, if $E_d$ is closed under $Q_\agentStyle{i}$.
    A \emph{local (epistemic) action} is any epistemic action that is local for some agent.
\end{definition}

The application of an action to a state is formalised by the \emph{product update}.
We say that an event $e$ is \emph{applicable} in $w$ if $(M, w) \models \mymathit{pre}(e)$ and that an epistemic action $(A, E_d)$ is \emph{applicable} is an epistemic state $(M, W_d)$ if for each designated world $w \in W_d$ there exists an applicable designated event $e \in E_d$.
Namely, each designated world has to be updated by some designated event.
The following definition is adapted from~\cite{book/springer/vanDitmarsch2007}.

\begin{definition}[Product Update]%
\label{def:product-update}
    Let $\actionStyle{a} = ((E, Q, \mymathit{pre}, \mymathit{post}), E_d)$ be an epistemic action applicable in an epistemic state $s = ((W, R, L), W_d)$.
    The \emph{product update} of $s$ with $\actionStyle{a}$ is the epistemic state $s \otimes \actionStyle{a} = ((W', R', L'), W'_d)$, where:
    \begin{itemize}
        \item $W'         = \{(w, e) \in W {\times} E \mid (M, w) \models \mymathit{pre}(e)\}$;
        \item $R'_\agentStyle{i}       = \{((w, e), (v, f)) \in W' {\times} W' \mid w R_\agentStyle{i} v \text{ and } e Q_\agentStyle{i} f\}$;
        \item $L'(w, e) = \{\fluentStyle{p} \in \atomSet \mid (M, w) \models \mymathit{post}(e, \fluentStyle{p})\}$; and
        \item $W'_d       = \{(w, e) \in W' \mid w \in W_d \text{ and } e \in E_d\}$.
    \end{itemize}
\end{definition}

\noindent
The product update produces an updated state as follows.
The set of possible worlds of the updated epistemic state contains the world-event pairs $(w, e)$ such that $e$ is applicable in $w$.
Each such pair represents the result of applying event $e$ to world $w$.
The new knowledge/beliefs of agents are obtained from the combination of what they knew/believed before the action, and what they observed about the action:
If in world $w$ agent $\agentStyle{i}$ considers $\fluentStyle{v}$ to be possible, and in event $e$ the agent considers $\fluentStyle{f}$ to be possible, then in the new world $(w, e)$ agent $\agentStyle{i}$ will consider $(\fluentStyle{v}, \fluentStyle{f})$ to be possible.
The label of a world-event pair $(w, e)$ is calculated by looking at the postconditions of $e$: An atom $\fluentStyle{p}$ is in the label of $(w, e)$ iff $w$ satisfies the formula $\mymathit{post}(e, \fluentStyle{p})$.
A new world $(w, e)$ is designated iff both $w$ and $e$ were originally designated.

Up to this point, we have covered epistemic states and actions of $\Lang[C]$, and product update.
As mentioned in Section~\ref{sec:language-states-epddl}, the language $\Lang[C]$ and epistemic states of $\Lang[C]$ (Definition~\ref{def:multi-pointed-epistemic-state}) are later going to be employed in the semantics of \textsc{epddl} (see Sections~\ref{sec:semantics-formulas} and~\ref{sec:semantics-init-goal}).
The objects that are going to provide a semantics for actions in \textsc{epddl}, however, are \emph{not} the epistemic actions defined above, but a generalisation of such objects called \emph{abstract epistemic actions}.
We are going to motivate and introduce abstract epistemic actions in Section~\ref{sec:abstract-epistemic-action}.

Before we move to such objects, we explore some important types of epistemic actions and show how these are represented by multi-pointed event models.
We provide illustrative examples of several types of actions using our Epistemic Blocks World problem.

    \subsubsection{Types of Epistemic Actions}\label{sec:epistemic-action-types}
    Epistemic actions can be classified in different ways.
    For instance, we can distinguish actions based on the type of change they bring about, or on the way agents observe the actions.
    We now provide a brief, yet exhaustive taxonomy of epistemic actions, and illustrate with examples the effects of several types of actions.
    
    Classical actions bring about physical change, but epistemic actions need not in general.
    Epistemic actions that change some physical property of the world, like the position of a block, are called \emph{ontic} (or \emph{world-altering});
    otherwise, they are called \emph{purely epistemic}.
    Actions of the latter type only change the knowledge/beliefs of some agent, \eg when an agent shares some information with others (see Example~\ref{ex:public-announcement}).
    More precisely, an action is purely epistemic if for all events $e$ we have that whenever $e$ is applicable in some world $w$, $\fluentStyle{p}$ holds in $w$ iff the postcondition of $\fluentStyle{p}$ in $e$ does, \ie if
    $\models \mymathit{pre}(e) \rightarrow \bigwedge_{\fluentStyle{p} \in \atomSet} \left( \fluentStyle{p} \leftrightarrow \mymathit{post}(e, \fluentStyle{p}) \right)$.
    An epistemic action is called \emph{atomic} if it has only one event.

\begin{figure}
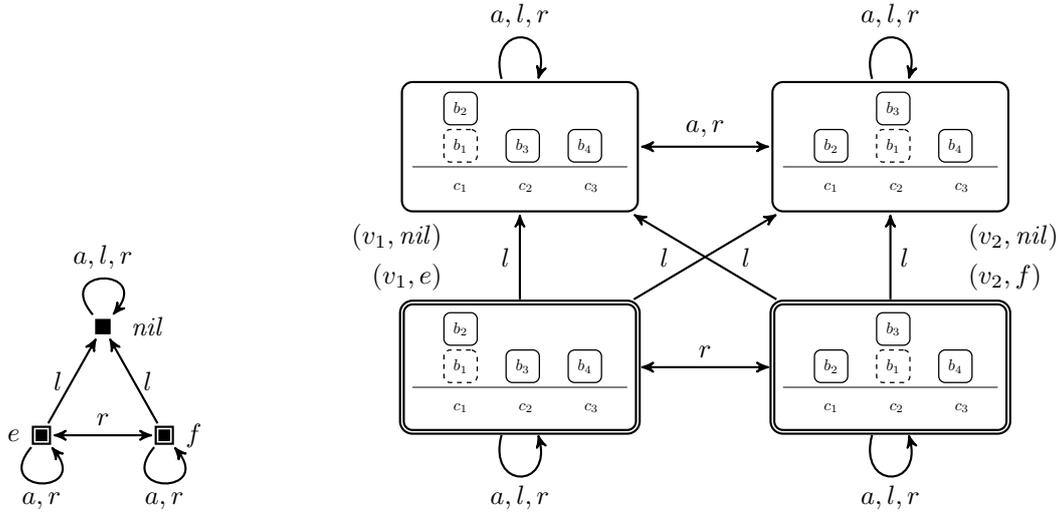

        \centering
        \subfloat{
            \begin{tikzpicture}[node distance=32pt and 14pt, >=stealth']
    \node[event,                      label=right:{$~\mathit{nil}$}] (nil) {};
    \node[devent, below left=of nil,  label=left :{$e$}] (e) {};
    \node[devent, below right=of nil, label=right:{$f$}] (f) {};

    \path
        (e) edge[->, thick, loop, out=-130, in=-50, looseness=8] node[below] {$a, r$} (e)
        (f) edge[->, thick, loop, out=-130, in=-50, looseness=8] node[below] {$a, r$} (f)
        (nil) edge[->, thick, loop, out=130, in=50, looseness=8] node[above] {$a, l, r$} (nil)
        (e) edge[<->, thick] node[above] {$r$} (f)
        (e) edge[->, thick] node[left] {$l$} (nil)
        (f) edge[->, thick] node[right] {$l$} (nil)
    ;
\end{tikzpicture}
        }
        \qquad
        \qquad
        \subfloat{
            \begin{tikzpicture}[node distance=32pt and 48pt, >=stealth']
    \node[graphicaldworld, scale=0.6, label=above left :{$(v_1, e)$}] (v1)                          {\input{img/examples/single-agent-bw/single-agent-bw-1.tex}};
    \node[graphicaldworld, scale=0.6, label=above right:{$(v_2, f)$}] (v2)            [right=of v1] {\input{img/examples/single-agent-bw/single-agent-bw-2.tex}};
    \node[graphicalworld,  scale=0.6, label=below left :{$(v_1, \mathit{nil})$}] (v3) [above=of v1] {\input{img/examples/single-agent-bw/single-agent-bw-1.tex}};
    \node[graphicalworld,  scale=0.6, label=below right:{$(v_2, \mathit{nil})$}] (v4) [right=of v3] {\input{img/examples/single-agent-bw/single-agent-bw-2.tex}};

    \path
        (v1) edge[<->, thick] node[above] {$\agentStyle{r}$} (v2)
        (v1) edge[->,  thick] node[left ] {$\agentStyle{l}$} (v3)
        (v1) edge[->,  thick] node[left ] {$\agentStyle{l~~~~}$} (v4)
        (v2) edge[->,  thick] node[right] {$\agentStyle{~~~~l}$} (v3)
        (v2) edge[->,  thick] node[right] {$\agentStyle{l}$} (v4)
        (v1) edge[->,  thick, loop, out=-105, in=-75, looseness=4] node[below] {$\agentStyle{a},\agentStyle{l},\agentStyle{r}$} (v1)
        (v2) edge[->,  thick, loop, out=-105, in=-75, looseness=4] node[below] {$\agentStyle{a},\agentStyle{l},\agentStyle{r}$} (v2)
        (v3) edge[<->, thick] node[above] {$\agentStyle{a},\agentStyle{r}$} (v4)
        (v3) edge[->,  thick, loop, out=105, in= 75, looseness=4] node[above] {$\agentStyle{a},\agentStyle{l},\agentStyle{r}$} (v3)
        (v4) edge[->,  thick, loop, out=105, in= 75, looseness=4] node[above] {$\agentStyle{a},\agentStyle{l},\agentStyle{r}$} (v4)
    ;
\end{tikzpicture}
        }
        \caption{Quasi-private sensing action $\mymathit{quasiPrivPeek}(a, r, b_2, b_1)$ and epistemic state $s''_r = s'_r \otimes \mymathit{quasiPrivPeek}(a, r, b_2, b_1)$. The pre- and postconditions of the events are described in Example~\ref{ex:quasi-private-sensing}.}
        \label{fig:quasi-private-sensing}
    \end{figure}

    For a group $G \subseteq \agentSet$ of agents, an action is said to be \emph{group observable by $G$}~\cite{journals/jancl/Bolander2011} if for each $\agentStyle{i} \in G$ the accessibility relation $Q_\agentStyle{i}$ is the identity (\ie no two distinct events are connected via $Q_\agentStyle{i}$).
    An action is called \emph{fully observable}, or \emph{public}, if is group observable by all agents;
    otherwise, it is called \emph{partially observable}.
    A partially observable action is said to be \emph{semi-private} if for each agent $\agentStyle{j} \in \agentSet \setminus G$ the accessibility relation $Q_\agentStyle{j}$ is the universal relation (\ie it connects all events):
    All agents know that the action is taking place, but only those in $G$ know the actual effects of the action.
    A partially observable action is said to be \emph{private} if for each agent $\agentStyle{j} \in \agentSet \setminus G$ we have $Q_\agentStyle{j} = \{ (e, \fluentStyle{f}) \mid \fluentStyle{f}$ is trivial$\}$:
    Only the agents in $G$ know that the action is taking place and its effects, while the remaining agents are oblivious to the action.
    Public, semi-private, and private epistemic actions capture different levels of observability of agents in event models.
    We can also combine private and semi-private as shown in Figure \ref{fig:quasi-private-sensing} to obtain a slightly more complex scenario where some agents observe the outcomes of the action (fully observant agents), some other only know about the execution of the action, but not its effects (partially observant agents), and the remaining agents are ignorant about the fact that the action is taking place (oblivious agents).
    We call these actions \emph{quasi-private actions}.
    Note that more complex and nuanced degrees of observability are also possible.
    For instance, we might have agents who believe that they are fully observant, while in fact they are deceived into believing so.

    The \emph{disjoint union} $\actionStyle{a} \sqcup \actionStyle{a}'$ of two actions $\actionStyle{a} = ((E, Q, \mymathit{pre}, \mymathit{post}), E_d)$ and $\actionStyle{a}' = ((E', Q', \mymathit{pre}',$ $\mymathit{post}'), E'_d)$ is the epistemic action $a \sqcup a' = ((E \sqcup E', Q \sqcup Q', \mit{pre} \sqcup \mit{pre}', \mit{post} \sqcup \mit{post}'), E_d \sqcup E_d')$ (we recall that $\sqcup$ denotes the disjoint union of two sets).
    Non-deterministic actions are then defined as follows.

    \begin{definition}[Non-Determinstic Epistemic Action]%
    \label{def:non-det-action}
        An epistemic action $\actionStyle{a}$ of $\Lang[C]$ is called a \emph{non-deterministic (epistemic) action} if $\actionStyle{a}$ is the disjoint union of two epistemic actions $\actionStyle{b}$, $\actionStyle{b}'$ for which there exists an epistemic state $s$ of $\Lang[C]$ and a formula $\phi \in \Lang[C]$ such that $s \otimes \actionStyle{b} \models \phi$ iff $s \otimes \actionStyle{b}' \not\models \phi$.
        Otherwise, $\actionStyle{a}$ is called \emph{deterministic}.
    \end{definition}

    \noindent
    A non-deterministic action can be thought of as multiple actions happening simultaneously.
    To ensure that the union of actions is non-trivial, we require that each action brings about different changes than the others.
    An action $((E, Q, \mymathit{pre}, \mymathit{post}), E_d)$ is called \emph{globally deterministic} if all preconditions are mutually inconsistent, \ie $\models \mymathit{pre}(e) \land \mymathit{pre}(f) \rightarrow \bot$ for all distinct events $e, f \in E$~\cite{journals/jancl/Bolander2011}.
    A \emph{sensing action} is an epistemic action such that:
    \begin{inparaenum}
        \item it is purely epistemic,
        \item it is globally deterministic, and
        \item its preconditions cover the logical space, \ie $\models \bigvee_{e \in E} \mymathit{pre}(e) \leftrightarrow \top$~\cite{journals/jancl/Bolander2011}.
    \end{inparaenum}

    We now look at some examples.
    Perhaps the simplest epistemic actions are \emph{public announcements}, which are purely epistemic, fully observable, deterministic, atomic actions.
    Public announcements are used by agents to synchronously broadcast some piece of information, as the next example shows.

    \begin{figure}
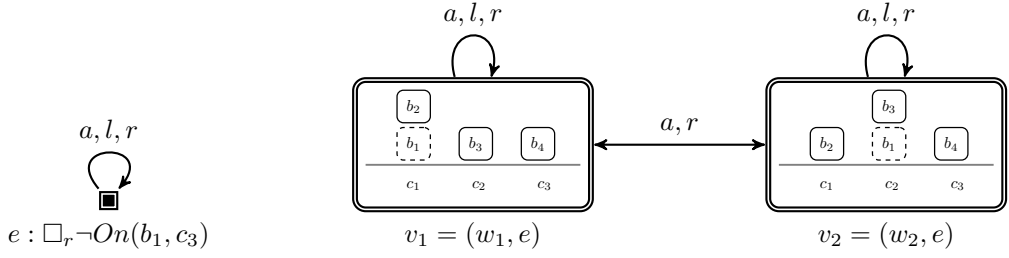

        \centering
        \subfloat{
            \begin{tikzpicture}[>=stealth']
    \node[devent, label=below:{$~e : \B{r} \neg On(\objectStyle{b}_1, \objectStyle{c}_3)~$}] (e) {};

    \path
        (e) edge[->, thick, loop, out=130, in=50, looseness=8] node[above] {$\agentStyle{a},\agentStyle{l},\agentStyle{r}$} (e)
    ;
\end{tikzpicture}
        }
        \qquad
        \qquad
        \subfloat{
            \begin{tikzpicture}[node distance=32pt and 64pt, >=stealth']
    \node[graphicaldworld, scale=0.6, label=below:{$v_1 = (w_1, e)$}] (v1)               {\input{img/examples/single-agent-bw/single-agent-bw-1.tex}};
    \node[graphicaldworld, scale=0.6, label=below:{$v_2 = (w_2, e)$}] (v2) [right=of v1] {\input{img/examples/single-agent-bw/single-agent-bw-2.tex}};

    \path
        (v1) edge[<->, thick] node[above] {$\agentStyle{a},\agentStyle{r}$} (v2)
        (v1) edge[->,  thick, loop, out=105, in= 75, looseness=4] node[above] {$\agentStyle{a},\agentStyle{l},\agentStyle{r}$} (v1)
        (v2) edge[->,  thick, loop, out=105, in= 75, looseness=4] node[above] {$\agentStyle{a},\agentStyle{l},\agentStyle{r}$} (v2)
    ;
\end{tikzpicture}
        }
        \caption{Public announcement $\actionStyle{a} = \actionStyle{ann}(\B{r} \neg \mymathit{On}(\objectStyle{b}_1, \objectStyle{c}_3))$ (left) and epistemic state $s'_r = s_\agentStyle{r} \otimes \actionStyle{a}$ (right) of Example \ref{ex:public-announcement}.
        Events are graphically represented by squares (designated events are boxed).
        Each event $e$ is labeled by a pre-/postconditions pair $\langle \mymathit{pre}(e), \mymathit{post}(e)\rangle$ (if $e$ has trivial postconditions, we simply label it with $\mymathit{pre}(e)$).}
        \label{fig:public-announcement-update}
    \end{figure}

    \begin{example}[Public Announcement]\label{ex:public-announcement}
        A public announcement of a formula $\phi$ is any epistemic action $\actionStyle{ann}(\phi) = ((\{e\}, Q, \mymathit{pre}, \mymathit{post}), \{e\})$, where $Q_\agentStyle{i} = \{(e, e)\}$ for all $\agentStyle{i} \in \agentSet$, $\mymathit{pre}(e) = \phi$ and such that $e$ has trivial postconditions~\cite{baltagAL/1998/Logic}.
        Let $s_r$ be the epistemic state from Example \ref{ex:multi-agent-bw-local}.
        As agent $\agentStyle{r}$ sees that block $\objectStyle{b}_1$ is not in the right column, $\agentStyle{r}$ could publicly announce this information to all agents, \ie $r$ could publicly announce that $\B{r} \neg \mymathit{On}(\objectStyle{b}_1, \objectStyle{c}_3)$.
        The public announcement $\actionStyle{ann}(\B{r} \neg \mymathit{On}(\objectStyle{b}_1, \objectStyle{c}_3))$ and the epistemic state $s_\agentStyle{r} \otimes \actionStyle{ann}(\B{r} \neg \mymathit{On}(\objectStyle{b}_1, \objectStyle{c}_3))$  are shown in Figure \ref{fig:public-announcement-update}.
        Following Definition~\ref{def:product-update}, the only event $e$ of $\actionStyle{ann}(\B{r} \neg \mymathit{On}(\objectStyle{b}_1, \objectStyle{c}_3))$ is applied to all worlds of $s_\agentStyle{r}$ that satisfy its precondition, \ie $w_1$ and $w_2$.
        As $w_1$, $w_2$ and $e$ are all designated, so are $(w_1, e)$ and $(w_2, e)$.
        Since $e$ has trivial postconditions, applying $e$ does not change the labels of worlds.
        Finally, since $Q_\agentStyle{i} = \{(e, e)\}$ for all $i \in \agentSet$, the update preserves all edges between the new worlds.
        Note that after the update we have $s_\agentStyle{r} \otimes \actionStyle{ann}(\B{r} \neg \mymathit{On}(\objectStyle{b}_1, \objectStyle{c}_3)) \models \CK{\agentSet} \B{r} \neg \mymathit{On}(\objectStyle{b}_1, \objectStyle{c}_3)$: Everybody knows that $\B{r} \neg \mymathit{On}(\objectStyle{b}_1, \objectStyle{c}_3)$ and, since this is a \emph{public} announcement, everybody knows that everybody knows that $\B{r} \neg \mymathit{On}(\objectStyle{b}_1, \objectStyle{c}_3)$, and so on.
    \end{example}

    \begin{figure}
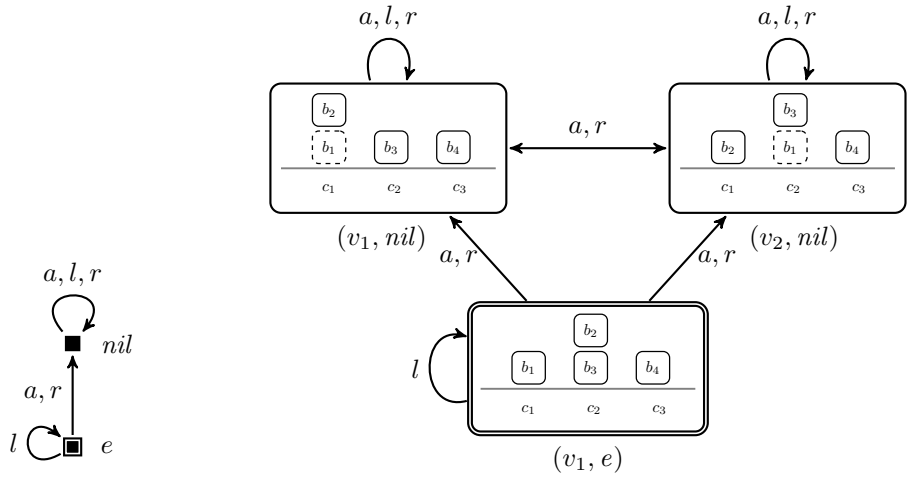

        \centering
        \subfloat{
            \begin{tikzpicture}[>=stealth']
    \node[devent,                 label=right:{$~e$}] (e) {};
    \node[event,  above=3em of e, label=right:{$~\mathit{nil}$}] (nil) {};

    \path
        (e) edge[->, thick, loop, out=210, in=140, looseness=8] node[left] {$\agentStyle{l}$} (e)
        (nil) edge[->, thick, loop, out=130, in=50, looseness=8] node[above] {$\agentStyle{a},\agentStyle{l},\agentStyle{r}$} (nil)
        (e) edge[->, thick] node[left] {$\agentStyle{a},\agentStyle{r}$} (nil)
    ;
\end{tikzpicture}
        }
        \qquad
        \qquad
        \subfloat{
            \begin{tikzpicture}[node distance=32pt and -16pt, >=stealth']
    \node[graphicaldworld, scale=0.6, label=below:{$(v_1, \mathit{e})$}] (v1)                      {\input{img/examples/partial-observability/moved-block.tex}};
    \node[graphicalworld, scale=0.6, label=below:{$(v_1, \mathit{nil})~~$}] (v2) [above left=of v1]  {\input{img/examples/single-agent-bw/single-agent-bw-1.tex}};
    \node[graphicalworld, scale=0.6, label=below:{$~~(v_2, \mathit{nil})$}] (v3) [above right=of v1] {\input{img/examples/single-agent-bw/single-agent-bw-2.tex}};

    \path
        (v1) edge[->,  thick, loop, out=195, in= 165, looseness=2] node[left] {$\agentStyle{l}$} (v1)
        (v2) edge[->,  thick, loop, out=105, in= 75, looseness=4] node[above] {$\agentStyle{a},\agentStyle{l},\agentStyle{r}$} (v2)
        (v3) edge[->,  thick, loop, out=105, in= 75, looseness=4] node[above] {$\agentStyle{a},\agentStyle{l},\agentStyle{r}$} (v3)
        (v1) edge[->,  thick] node[left]  {$\agentStyle{a},\agentStyle{r}$} (v2)
        (v1) edge[->,  thick] node[right] {$\agentStyle{a},\agentStyle{r}$} (v3)
        (v2) edge[<->, thick] node[above] {$\agentStyle{a},\agentStyle{r}$} (v3)
    ;
\end{tikzpicture}
        }
        \caption{Private ontic action $\mymathit{privMove}(l, b_2, b_1, b_3)$ and state $s''_l = s'_l \otimes \mymathit{privMove}(l, b_2, b_1, b_3)$. The preconditions and postconditions of the events are described in Example~\ref{ex:private-ontic}.}
        \label{fig:private-ontic}
    \end{figure}

    Private ontic actions represent actions where some agent knows that something about the world is being changed, while some other agents think that nothing is happening, as the next example shows.

    \begin{example}\label{ex:private-ontic}
        Let $\mymathit{privMove}(i, b, x, y)$ denote the epistemic action where agent $i$ moves block $b$ from position $x$ to position $y$ while the remaining agents are not looking.
        Then, $\mymathit{privMove}(i, b, x, y) = ((E, Q, \mymathit{pre}, \mymathit{post}), E_d)$, where:
        \begin{itemize}
            \item $E = \{ e, \mymathit{nil} \}$, $E_d = \{ e \}$;
            \item $Q_i = \{(f, f) \mid f \in E\}$, and for all $j \neq i$, $Q_j = \{ (f, \mymathit{nil}) \mid f \in E \}$;
            \item $\mymathit{pre}(e) = \B{i} (\mymathit{On}(b, x) \land \mymathit{Clear}(b) \land \mymathit{Clear}(y))$: agent $i$ needs to know/believe that block $b$ is on top of $x$ and that both $b$ and $y$ are clear;
            \item $\mymathit{post}(e, \mymathit{On}(b, x)) = \mymathit{post}(e, \mymathit{Clear}(y)) = \bot$;
            \item $\mymathit{post}(e, \mymathit{On}(b, y)) = \mymathit{post}(e, \mymathit{Clear}(x)) = \top$;
            \item $\mymathit{post}(e, p) = p$, for all other atoms $p$; and
            \item $\mymathit{nil}$ has trivial pre- and postconditions.
        \end{itemize}
        Event $e$ represents the move of the block (observed by agent $i$), and $\mymathit{nil}$ represents the null event (observed by the remaining agents).
        Figure~\ref{fig:private-ontic} (left) shows the case of $\mymathit{privMove}(l, b_2, b_1, b_3)$, \ie the case of agent $l$ privately moving block $b_2$ from $b_1$ to $b_3$.

        Let $s'_r = ((W', R', L'), W'_d)$ be the epistemic state of Figure~\ref{fig:public-announcement-update}, which is a local state for agents $a$ and $r$.
        Consider now the state $s'_l = ((W', R', L'), \{ v_1 \})$, which instead represents $l$'s perspective.
        Note that $\mymathit{privMove}(l, b_2, b_1, b_3)$ is applicable in $s'_l$, but not in $s'_r$ as $v_2$ does not satisfy the precondition of the only designated event $e$.
        The state $s''_l = s'_l \otimes \mymathit{privMove}(l, b_2, b_1, b_3) = ((W'', R'', L''), W''_d)$ is shown on the right-hand side of Figure~\ref{fig:private-ontic}, and it is computed as follows.
        The precondition of event $e$ only holds in world $v_1$ of $s'_l$, while $\mymathit{nil}$ holds in both $v_1$ and $v_2$.
        Thus, the worlds of $s''_l$ are $(v_1, e)$, $(v_1, \mymathit{nil})$ and $(v_2, \mymathit{nil})$.
        The label of $(v_1, e)$ is obtained by removing $\mymathit{On}(\objectStyle{b}_2, \objectStyle{b}_1)$ and $\mymathit{Clear}(\objectStyle{b}_3)$ from the label of $v_1$, and by adding $\mymathit{On}(\objectStyle{b}_2, \objectStyle{b}_3)$ and $\mymathit{Clear}(\objectStyle{b}_1)$.
        The only designated world is $(v_1, e)$, as both $v_1$ and $e$ are designated.

        Reasoning as in Example~\ref{ex:public-announcement}, the labeled edges outgoing from $(v_1, \mymathit{nil})$ and $(v_2, \mymathit{nil})$ are the same as those of $v_1$ and $v_2$.
        Since $e Q_l e$ and $v_1 R'_l v_1$, from Definition~\ref{def:product-update} we have $(v_1, e) R''_l (v_1, e)$.
        Similarly, we also get $(v_1, e) R''_a (v_1, \mymathit{nil})$, $(v_1, e) R''_r (v_1, \mymathit{nil})$, $(v_1, e) R''_a (v_2, \mymathit{nil})$ and $(v_1, e) R''_r (v_2, \mymathit{nil})$.
        Note that the application of the null event causes the entire state $s'_l$ to be ``copied'' into $s''_l$.
        As a consequence, any formula that is believed by agents $a$ and $r$ in the previous state is going to be believed by them in the new one.
        Moreover, after the update we have $s''_l \models \B{l} \mymathit{On}(\objectStyle{b}_2, \objectStyle{b}_3)$.

        Finally, it is not hard to check that the accessibility relations of agents $a$ and $r$ are no longer symmetric.
        Moreover, all relations are serial, transitive, and Euclidean.
        Therefore, $s''_l$ is not an S5$_n$-state like $s'_l$, but rather is a KD45$_n$-state.
    \end{example}

    Sensing actions allow one or more agents to learn whether some atom or formula holds or not.
    These actions can be private, meaning that some agent does not know that the action is taking place, semi-private, meaning that some agents know that the action is being performed, but they do not know what is being learned, or quasi-private, if both kinds of agents are present.
    The next example shows a quasi-private sensing action.

    \begin{example}\label{ex:quasi-private-sensing}
        Let $\mymathit{quasiPrivPeek}(i, j, b, x)$ denote the epistemic action where agent $i$ peeks under block $b$ to check whether it is on position $x$, while $j$ observes the action and the remaining agent, call it $k$, is distracted.
        Agent $i$ is called a \emph{fully observant} agent, as it knows what is being learned; agent $j$ is called a \emph{partially observant} agent, as it only knows that the action is taking place, but not its effects; and agent $k$ is called \emph{oblivious}, as it believes that no action is taking place.
        Then, $\mymathit{quasiPrivPeek}(i, j, b, x) = ((E, Q, \mymathit{pre}, \mymathit{post}), E_d)$, where:
        \begin{itemize}
            \item $E = \{ e, f, \mymathit{nil} \}$, $E_d = \{ e, f \}$;
            \item $Q_i = \{ (e, e), (f, f), (\mymathit{nil}, \mymathit{nil}) \}$;
            \item $Q_j = \{ (e, e), (e, f), (f, e), (f, f), (\mymathit{nil}, \mymathit{nil}) \}$;
            \item $Q_k = \{ (e, \mymathit{nil}), (f, \mymathit{nil}), (\mymathit{nil}, \mymathit{nil}) \}$;
            \item $\mymathit{pre}(e) = \mymathit{Clear}(b) \land      \mymathit{On}(b, x)$;
            \item $\mymathit{pre}(f) = \mymathit{Clear}(b) \land \neg \mymathit{On}(b, x)$;
            \item All events have trivial postconditions and the precondition of $\mymathit{nil}$ is also trivial.
        \end{itemize}
        Event $e$ represents the case where agent $i$ learns that $b$ is on top of $x$, event $f$ the one in which $i$ learns that $b$ is not on top of $x$, and $\mymathit{nil}$ represents the null event.
        In general, we do not know a priori what agent $i$ will learn, as this depends on the state in which the action is applied.
        Therefore, both events $e$ and $f$ must be designated.
        As agent $i$ is fully observant, meaning that it distinguishes between events $e$ and $f$ (\ie there are no $i$-edges between $e$ and $f$), the agent will come to know whether $b$ is on top of $x$.
        Agent $j$ is only partially observant, so it does not distinguish between $e$ and $f$ (\ie it considers both to be possible in both events), so $j$ will not come to know what $i$ has learned, but it will know \emph{that} $i$ knows whether $\mymathit{On}(b, x)$.
        Agent $k$ is oblivious, so it believes that nothing is happening.
        Figure~\ref{fig:quasi-private-sensing} (left) shows the case of $\mymathit{quasiPrivPeek}(a, r, b_2, b_1)$, \ie the case of agent $a$ peeking under block $b_2$ to check whether it is on top of $b_1$, while agent $r$ observes and $l$ is oblivious.

        Letting $s'_r = ((W', R', L'), W'_d)$ be the epistemic of Figure~\ref{fig:public-announcement-update}, the state $s''_r = s'_r \otimes \mymathit{quasiPrivPeek}(a, r, b_2, b_1) = ((W'', R'', L''), W''_d)$ is shown on the right-hand side of Figure~\ref{fig:quasi-private-sensing}, and it is computed as follows.
        The precondition of event $e$ only holds in world $v_1$ of state $s'_r$, the precondition of $f$ holds only in $v_2$, while $\mymathit{nil}$ holds in both $v_1$ and $v_2$.
        Thus, the worlds of $s''_r$ are $(v_1, e)$, $(v_2, f)$, $(v_1, \mymathit{nil})$ and $(v_2, \mymathit{nil})$.
        As all events have trivial postconditions, the labels of the updated worlds are the same as their corresponding original worlds.
        The designated worlds are $(v_1, e)$ and $(v_2, f)$, as $v_1$, $v_2$, $e$ and $f$ are all designated.

        Reasoning as in Example~\ref{ex:public-announcement}, the labeled edges outgoing from $(v_1, \mymathit{nil})$ and $(v_2, \mymathit{nil})$ are the same as those of $v_1$ and $v_2$.
        Since $e Q_a e$, $f Q_a f$, $v_1 R'_a v_1$ and $v_2 R'_a v_2$, from Definition~\ref{def:product-update} we have $(v_1, e) R''_a (v_1, e)$ and $(v_2, f) R''_a (v_2, f)$.
        Similarly, for all $x, y \in \{ (v_1, e), (v_2, f) \}$ we have $x R''_r y$, $x R''_l (v_1, \mymathit{nil})$ and $x R''_l (v_2, \mymathit{nil})$.
        
        After the update we have the following: $s''_r \models \Kw{a} \mymathit{On}(\objectStyle{b}_2, \objectStyle{b}_1)$, $s''_r \models \neg \Kw{r} \mymathit{On}(\objectStyle{b}_2, \objectStyle{b}_1) \land \B{r} \Kw{a} \mymathit{On}(\objectStyle{b}_2, b_1)$ and $s''_r \models \B{l} \neg \Kw{\{a, r\}} \mymathit{On}(\objectStyle{b}_2, \objectStyle{b}_1)$:
        Agent $a$ knows whether $b_2$ is on top of $b_1$ (as the agent is fully observant), agent $r$ does not, but believes that $a$ does (as it is partially observant), and agent $l$ believes that both $a$ and $r$ don't know whether $b_2$ is on top of $b_1$ (as the agent is oblivious).
        Note that agent $l$ has a \emph{false belief} about the perspective of agent $a$.
        Moreover, it is not hard to check that the accessibility relation of agent $l$ is not symmetric, while all relations are serial, transitive, and Euclidean.
        Therefore, $s''_r$ is a KD45$_n$-state.
    \end{example}

    Last, we look at a simple instance of a non-deterministic action.
    
\begin{figure}
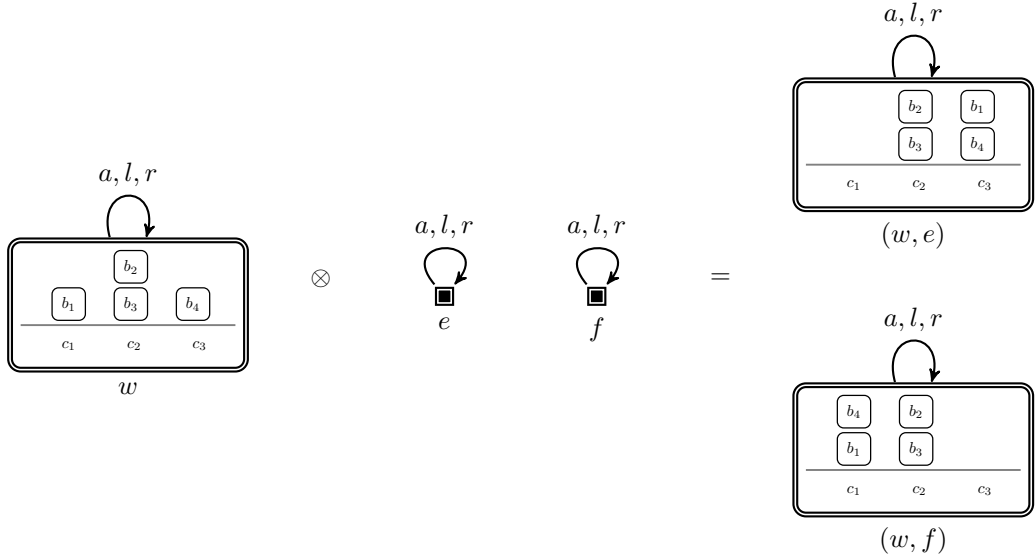

        \centering
        \begin{tikzpicture}[node distance=16pt, >=stealth']
    \node                  (s) {\input{img/examples/non-det-public-ontic/ndpo-es.tex}};
    \node[right=of s]      (otimes) {$\otimes$};
    \node[right=of otimes] (a) {\input{img/examples/non-det-public-ontic/ndpo.tex}};
    \node[right=of a]      (eq) {$=$};
    \node[right=of eq]     (s') {\input{img/examples/non-det-public-ontic/ndpo-es_.tex}};
\end{tikzpicture}
        \caption{Epistemic state $s$ (left), non-deterministic action $a$ (middle) and state $s' = s \otimes a$ (right) of Example~\ref{ex:non-det-public-ontic}. Event $e$ represents the move of $b_1$ on top of $b_4$ and $f$ the move of $b_4$ on top of $b_1$.}
        \label{fig:non-det-public-ontic}
    \end{figure}

    \begin{example}\label{ex:non-det-public-ontic}
        Let $s$ be the epistemic state in Figure~\ref{fig:non-det-public-ontic}.
        It is a state where block $b_1$ is on the left column, $b_2$ is stacked on top of $b_3$ on the middle column, $b_4$ is on the right one, and this is commonly known to all agents, \ie $s \models \CK{\{ a, l, r \}} \mymathit{On}(b_1, c_1) \land \mymathit{On}(b_2, c_2) \land \mymathit{On}(b_3, b_2) \land \mymathit{On}(b_4, c_3)$.

        Suppose the agents flip a coin to decide what block to move next: if the coin lands on heads (denoted by the fresh propositional atom $\mymathit{heads}$), the agents will move block $b_1$ from $c_1$ to $b_4$, and if it lands on tails they will move $b_4$ from $c_3$ to $b_1$.
        We assume both the result of the coin toss and the results of the moves are publicly observed by all agents.
        Since the outcome of the coin toss can not be known in advance, this is a non-deterministic action.
        We can model this as the epistemic action $a = ((E, Q, \mymathit{pre}, \mymathit{post}), E_d)$, shown in Figure~\ref{fig:non-det-public-ontic} (middle):
        \begin{itemize}
            \item $E = E_d = \{ e, f \}$;
            \item For all $i \in \agentSet$, we have $Q_i = \{ (e, e), (f, f) \}$;
            \item $\mymathit{pre}(e) = \mymathit{heads} \land \mymathit{On}(b_1, c_1) \land \mymathit{Clear}(b_1) \land \mymathit{Clear}(b_4)$;
            \item $\mymathit{pre}(f) = \neg\mymathit{heads} \land \mymathit{On}(b_4, c_3) \land \mymathit{Clear}(b_1) \land \mymathit{Clear}(b_4)$;
            \item $\mymathit{post}(e, \mymathit{On}(b_1, c_1)) = \mymathit{post}(e, \mymathit{Clear}(b_4)) = \bot$;
            \item $\mymathit{post}(e, \mymathit{On}(b_1, b_4)) = \mymathit{post}(e, \mymathit{Clear}(c_1)) = \top$;
            \item $\mymathit{post}(e, p) = p$, for all other atoms $p$;
            \item $\mymathit{post}(f, \mymathit{On}(b_4, c_3)) = \mymathit{post}(f, \mymathit{Clear}(b_1)) = \bot$;
            \item $\mymathit{post}(f, \mymathit{On}(b_4, b_1)) = \mymathit{post}(f, \mymathit{Clear}(c_3)) = \top$; and
            \item $\mymathit{post}(f, p) = p$, for all other atoms $p$.
        \end{itemize}
        Event $e$ represents the case where the coin lands on heads, triggering the move of $b_1$ to $b_4$, and event $f$ represents the opposite instance.
        Preconditions and postconditions for moving blocks are the standard ones.
        As the result of the coin toss can not be known a priori, both $e$ and $f$ must be designated.
        However, independent of the result of the coin toss, all agents will observe the consequent block move, and thus all agents distinguish between $e$ and $f$ (meaning that no edge connects them), making this a public action.
        Moreover, action $a$ can be defined as the disjoint union of two epistemic actions: the action where the coin lands on heads and $b_1$ is moved to $b_4$ (represented by the sub-model of $a$ rooted in event $e$), and the one where the coin lands on tails and $b_4$ is moved to $b_1$ (represented by the sub-model of $a$ rooted in $f$).

        Applying $a$ to state $s$ yields the epistemic state $s' = s \otimes a = ((W', R', L'), W'_d)$ shown in Figure~\ref{fig:non-det-public-ontic} (right), as we now describe.
        It is not hard to check that both events $e$ and $f$ are applicable in the only world $w$ of $s$.
        Thus, we get $W' = \{ (w, e), (w, f) \}$.
        Moreover, as $w$, $e$, and $f$ are all designated, we immediately get $W'_d = W'$.
        As all accessibility relations of all agents in $s$ and $a$ are reflexive, so will be those of $s'$, \ie for all $i \in \agentSet$ we get $R'_i = \{ (w', w') \mid w' \in W' \}$.
        Finally, the labels of the new worlds are computed as we did in Example~\ref{ex:private-ontic}.

        It is not hard to check that the updated state $s'$ is a non-deterministic epistemic state.
        The new state represents the effects of both block moves at once.
        Note that this is different than partial observability, as agents distinguish between worlds $(w, e)$ and $(w, f)$.
    \end{example}

    \subsection{Epistemic Planning Tasks}
    We recall the notions of epistemic planning tasks and solutions~\cite{conf/ijcai/Aucher2013}.
    For a sequence $\pi = \actionStyle{a}_1, \dots, \actionStyle{a}_l$ of actions and $1 \leq k \leq l$, $\pi_{\leq k}$ denotes the prefix $\actionStyle{a}_1, \dots, \actionStyle{a}_k$ of $\pi$, and $s \otimes \pi$ the state $s \otimes \actionStyle{a}_1 \dots \otimes \actionStyle{a}_l$ (if $\pi$ is empty, this is just $s$).
    We say that $\pi$ is \emph{applicable} in $s$ if for all $k$, $\actionStyle{a}_k$ is applicable in $s \otimes \pi_{\leq k-1}$.

    \begin{definition}[Epistemic Planning Task]\label{def:epistemic-planning-task}
        An \emph{(epistemic) planning task} of $\Lang[C]$ is a triple $T = (s_0, \actionSet, \phi_g)$, where $s_0$ is a state of $\Lang[C]$ (the \emph{initial state}), $\actionSet$ is a finite set of actions of $\Lang[C]$, and $\phi_g \in \Lang[C]$ is the \emph{goal formula}.
        A \emph{solution} (or \emph{plan}) to $T$ is a finite sequence $\pi$ of actions of $\actionSet$ such that:
        \begin{enumerate}
            \item $\pi$ is applicable in $s_0$; and
            \item $s_0 \otimes \pi \models \phi_g$.
        \end{enumerate}
    \end{definition}

    An epistemic planning task represents an instance of a problem.
    The initial state describes the knowledge/beliefs of agents at the beginning of the problem, while epistemic actions allow agents to modify the environment and each other's perspective.
    The goal represents the condition the agents want to achieve, which may describe both physical conditions (\eg the position of some block) and epistemic ones (\eg the fact that some agent knows the position of a block, or that some other falsely believes its position).

    \begin{example}[Epistemic Planning Task]\label{ex:epistemic-planning-task}
        Let $T = (s_0, \actionSet, \phi_g)$ be an epistemic planning task such that:
        \begin{itemize}
            \item $s_0$ is the epistemic state of Figure~\ref{fig:multi-agent-bw};
            \item $\actionSet = \{ \mymathit{ann}(\B{i} \mymathit{On}(b, x)), \mymathit{privMove}(i, b, x, y), \mymathit{quasiPrivPeek}(i, j, b, x) \mid i, j \in \agentSet, b$ is a block and $x, y$ are blocks/columns$\}$; and
            \item $\phi_g = \mymathit{On}(b_4, b_1) \land \Kw{a} \mymathit{On}(b_2, b_1) \land \B{r} \Kw{a} \mymathit{On}(b_2, b_1) \land \B{l} \CK{\{ a, l, r\}} \neg \Kw{\{ a, r\}} \mymathit{On}(b_2, b_1)$.
        \end{itemize}
        The goal is that $b_4$ is on top of $b_1$, that $a$ knows whether $b_2$ is on top of $b_1$ and $r$ believes this, and that $l$ believes that it is common belief that agents $a$ and $r$ don't know whether $b_2$ is on top of $b_1$.

        A solution to $T$ is the following action sequence:
        \begin{enumerate}
            \item $\mymathit{ann}(\B{r} \mymathit{On}(b_4, c_3))$;
            \item $\mymathit{quasiPrivPeek}(a, r, b_2, b_1)$;
            \item $\mymathit{privMove}(a, b_4, c_3, b_1)$.
        \end{enumerate}
        With the first action, agent $r$ publicly announces that $b_4$ is on top of $c_3$.
        In the resulting state (which is the same as that of Figure~\ref{fig:public-announcement-update}) we have that $\CK{\{ a, l, r \}} \mymathit{On}(b_4, c_3)$ holds.
        Moreover, the state satisfies $\CK{\{ a, l, r \}} \neg \Kw{\{ a, r\}} \mymathit{On}(b_2, b_1)$, as agents $a$ and $r$ still consider to be possible both that block $b_1$ is under $b_2$ and under $b_3$.
        Executing this action is necessary to make $\mymathit{privMove}(a, b_4, c_3, b_1)$ applicable.
        With the second action, agent $a$ peeks under $b_2$ and learns whether it is stacked on top of $b_1$ while $r$ observes and $l$ does not.
        In the resulting state, we thus have $\Kw{a} \mymathit{On}(b_2, b_1) \land \B{r} \Kw{a} \mymathit{On}(b_2, b_1)$.
        Moreover, since agent $l$ is oblivious, the new state also satisfies the formula $\B{l} \CK{\{ a, l, r\}} \neg \Kw{\{ a, r\}} \mymathit{On}(b_2, b_1)$.
        Finally, the last action makes $\mymathit{On}(b_4, b_1)$ true, satisfying the last part of the goal.
    \end{example}

    We conclude this section with the following remark.
    In DEL-based epistemic planning, non-determinism is modeled within epistemic states and actions (see Definitions~\ref{def:non-det-state} and~\ref{def:non-det-action}), rather than being handled at the algorithmic level~\cite{journals/jancl/Bolander2011}.
    We note that, because of this, any \emph{conformant planning task}~\cite{conf/aaai/SmithW98} can be represented by an equivalent epistemic planning task (namely, one that admits the same set of solutions), as we now informally show.
    A conformant planning task is a triple $T_\mymathit{conf} = (S, \actionSet, \phi_g)$, where $S = \{s_1, \dots, s_k\}$ is a set of possible initial epistemic states, $\actionSet$ is a finite set of epistemic actions and $\phi_g$ is the goal formula.
    An action sequence $\pi$ is a solution/plan to $T_\mymathit{conf}$ iff for all $s \in S$, we have that:
    \begin{inparaenum}
        \item $\pi$ is applicable in $s$, and
        \item $s \otimes \pi \models \phi_g$.
    \end{inparaenum}
    Consider now the epistemic planning task $T = (\bigsqcup_{s \in S} s, \actionSet, \phi_g)$, where the initial state of $T$ is the disjoint union of the states in $S$.
    By Definition~\ref{def:truth}, a formula holds in an epistemic state iff it holds in all of its designated worlds, implying that:
    \begin{enumerate}
        \item $\pi$ is applicable in $\bigsqcup_{s \in S} s$ iff $\pi$ is applicable in all $s \in S$, and
        \item $\bigsqcup_{s \in S} s \otimes \pi \models \phi_g$ iff for all $s \in S$, $s \otimes \pi \models \phi_g$.
    \end{enumerate}
    Therefore, a solution to $T$ is also a solution to $T_\mymathit{conf}$, and vice versa.

    \section{Abstract Epistemic Actions}\label{sec:abstract-epistemic-action}
    So far we have covered all the basic notions of DEL-based epistemic planning.
    We introduced different modalities to talk about various kinds of knowledge/beliefs of agents and groups of agents.
    With epistemic states we provided a semantics for our logical language of multi-agent epistemic logic, and described several non-trivial types of scenarios that can be modeled with them (\eg local states and non-deterministic states).
    As anticipated, in Section~\ref{sec:epddl-semantics} we are going to use the language $\Lang[C]$ and epistemic states of such language to provide a semantics for formulas and epistemic states of \textsc{epddl}, respectively.
    But what about epistemic actions?
    In the previous section, we described multi-pointed event models as a semantics for epistemic actions, and showcased a wide set of types of epistemic actions that can be modeled by such objects (\eg private ontic actions and quasi-private sensing actions).
    As seen, event models can represent various levels of observability, providing a very rich formalism for epistemic actions.
    Nonetheless, event models are quite low-level objects, as they describe specific instances of more generic action types.
    For example, a private announcement witnessed by agents $\agentStyle{i}$ and $\agentStyle{j}$, but not by $\agentStyle{k}$, and the same announcement witnessed by $\agentStyle{i}$ and $\agentStyle{k}$, but not by $\agentStyle{j}$, are two distinct actions, despite both being actions of the same type, namely a private announcement.
    Since our goal is to devise a language for representing epistemic planning tasks, it is thus desirable to avoid repeating the description of the same action for all possible combinations of edge labels.
    To this end, we introduce \emph{abstract epistemic actions}, a variation of epistemic actions that abstractly describe different degrees of observability without explicitly mentioning specific agents.
    In Section~\ref{sec:epddl-semantics}, we are going to use abstract epistemic actions to provide the semantics of actions of \textsc{epddl}.

    Abstract epistemic actions are inspired by the notion of \emph{frame of reference}, introduced by~\cite{journals/ai/BaralGPS22}, which essentially characterises a fixed set of perspectives that agents might have towards an action.
    More precisely, the frame of reference of an action partitions agents into three groups: fully observant, partially observant, and oblivious agents.
    Both fully and partially observant agents know about the fact that the action is being executed, while only fully observant agents know the effects of the action; oblivious agents do not know that the action is taking place, and instead believe that nothing is happening.
    We here generalise this idea by allowing actions to be defined on a generic set of such perspectives, that we call \emph{observability types}.
    Let $\obsTypeSet$ be a finite set of observability types.
    We can see each type as an abstract placeholder that denotes a particular perspective of some agent about an action.
    For instance, in a private announcement, we have two observability types, one for fully observant agents ($f$) and one for oblivious agents ($o$), so $\obsTypeSet = \{f, o\}$.
    In abstract epistemic actions (Definition~\ref{def:abstract-event-model}), we use observability types instead of agents, as we now proceed to show.

    \begin{figure}
        \centering
        \begin{tikzpicture}[>=stealth']
    \node[devent,                 label=below:{$~e$}] (e) {};
    \node[event,  right=5em of e, label=below:{$~\mathit{nil}$}] (nil) {};

    \path
        (e) edge[->, thick, loop, out=130, in=50, looseness=8] node[above] {$\agentStyle{f}$} (e)
        (nil) edge[->, thick, loop, out=130, in=50, looseness=8] node[above] {$\agentStyle{f},\agentStyle{o}$} (nil)
        (e) edge[->, thick] node[above] {$\agentStyle{o}$} (nil)
    ;
\end{tikzpicture}
        \caption{Pointed abstract frame for private actions, where $f$ and $o$ are observability types denoting fully observant and oblivious agents, respectively.}
        \label{fig:private-abstract-frame}
    \end{figure}
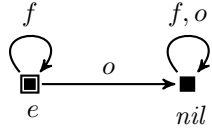

    \begin{definition}[Abstract Frames]\label{def:abstract-frame}
        Let $\obsTypeSet$ be a finite, non-empty set of observability types.
        An \emph{abstract frame} on $\obsTypeSet$ is a pair $F = (E, Q)$, where:
        \begin{itemize}
            \item $E \neq \varnothing$ is a finite set of events; and
            \item $Q : \obsTypeSet \rightarrow 2^{E \times E}$ assigns to each observability type $t \in \obsTypeSet$ an \emph{abstract accessibility relation} $Q_t$.
        \end{itemize}
        A \emph{multi-pointed abstract frame} is a pair $(F, E_d)$, where $F$ is an abstract frame on $\obsTypeSet$ and $E_d \subseteq E$ is a non-empty set of designated events.
    \end{definition}

    Abstract frames represent a set of perspectives on an epistemic action that can be held by one or more agents.
    For instance, Figure~\ref{fig:private-abstract-frame} shows the pointed abstract frame for private actions, where the abstract accessibility relations $Q_f$ and $Q_o$ represent the perspectives of fully observant and oblivious agents, respectively.

    \begin{definition}[Abstract Event Models and Epistemic Actions]\label{def:abstract-event-model}
        Let $\obsTypeSet$ be a finite, non-empty set of observability types.
        An \emph{abstract event model} of $\Lang[C]$ on $\obsTypeSet$ is a tuple $A = (E, Q, \mymathit{pre}, \mymathit{post}, \mymathit{obs})$, where $(E, Q)$ is an abstract frame on $\obsTypeSet$, $\mymathit{pre}$ and $\mymathit{post}$ are as in Definition \ref{def:event-model}, and $\mymathit{obs} : \agentSet \rightarrow (\obsTypeSet \rightarrow \Lang[C])$ assigns to each agent $\agentStyle{i} \in \agentSet$ an \emph{observability function} $\mymathit{obs}_\agentStyle{i}$ that maps to each observability type an \emph{observability condition} such that:
        \begin{itemize}
            \item Observability conditions are mutually inconsistent, \ie for all distinct $t, t' \in \obsTypeSet$ we have $\models \mymathit{obs}_\agentStyle{i}(t) \land \mymathit{obs}_\agentStyle{i}(t') \rightarrow \bot$; and
            \item Observability conditions cover the logical space, \ie $\models \bigvee_{t \in \obsTypeSet} \mymathit{obs}_\agentStyle{i}(t)$.
        \end{itemize}
        An \emph{abstract (epistemic) action} of $\Lang[C]$ on $\obsTypeSet$ is a pair $a = (A, E_d)$, where $A$ is an abstract event model of $\Lang[C]$ on $\obsTypeSet$ and $E_d \subseteq E$ is a non-empty set of designated events.
    \end{definition}

    Abstract epistemic actions do not explicitly provide the accessibility relation of each agent, but rather they specify a set of conditions that determine the perspective of each agent on the action (\ie its observability type).
    Note that the two requirements we imposed on observability functions guarantee that each agent is assigned a unique observability type.
    Nonetheless, the observability type of an agent is not fixed, as it varies based on the state where we evaluate the observability conditions.
    More precisely, given a state $s$ the $\mymathit{obs}$ function of an abstract action $\actionStyle{a}$ induces a function $\mymathit{type}_{\actionStyle{a},s} : \agentSet \rightarrow \obsTypeSet$ such that for each agent $\agentStyle{i} \in \agentSet$ we have that $\obsType[a,s]{\agentStyle{i}}$ is the unique observability type $t \in \obsTypeSet$ such that $s \models \mymathit{obs}_\agentStyle{i}(t)$.
    When $\actionStyle{a}$ and $s$ are clear from the context, we simply write $\obsType{\agentStyle{i}}$ for $\obsType[a,s]{\agentStyle{i}}$.

    \begin{example}\label{ex:abstract-action}
        Consider the standard epistemic action $\mymathit{privMove}(i, b, x, y)$ defined in Example~\ref{ex:private-ontic}.
        This is a private ontic action where agent $i$ moves block $b$ from position $x$ to $y$ while all remaining agents remain oblivious.
        We can generalise this action by making other agents oblivious only if, say, they are distracted, as we now show.
        Let $\mymathit{Distracted}(i)$ (for all $i \in \agentSet$) denote the fact that agent $i$ is distracted.
        We define the abstract epistemic action $\mymathit{absPrivMove}(i, b, x, y) = ((E, Q, \mymathit{pre}, \mymathit{post}, \mymathit{obs}), E_d)$, as follows:
        \begin{itemize}
            \item The abstract frame of the action is the one shown in Figure~\ref{fig:private-abstract-frame};
            \item The precondition of $\mymathit{nil}$ and the postconditions of events $e$ and $\mymathit{nil}$ are exactly as in Example~\ref{ex:private-ontic};
            \item $\mymathit{pre}(e) = \neg \mymathit{Distracted}(i) \land \psi_\mymathit{blocks} \land \psi_\mymathit{obs}$, where $\psi_\mymathit{blocks} = \B{i} (\mymathit{On}(b, x) \land \mymathit{Clear}(b) \land \mymathit{Clear}(y))$ and $\psi_\mymathit{obs} = \bigwedge_{j \in \agentSet} (\neg \mymathit{Distracted}(j) \rightarrow \B{j} \neg \mymathit{Distracted}(i))$;
            \item $\mymathit{obs}_i(f) = \top$ and $\mymathit{obs}_i(o) = \bot$; and
            \item For all $j \neq i$, $\mymathit{obs}_j(f) = \neg \mymathit{Distracted}(j)$ and $\mymathit{obs}_j(o) = \mymathit{Distracted}(j)$.
        \end{itemize}
        The precondition of event $e$ requires that agent $i$ is not distracted, that the agent knows/believes that the block can be effectively moved from $x$ to $y$, and that all non-distracted agents know/believe that agent $i$ is not distracted.
        The last condition ensures that attentive agents do not hold false beliefs about the attentiveness of the moving agent.
        The observability conditions for agent $i$ immediately give us that $\obsType[]{i} = f$: The moving agent knows that the action is taking place and what its effects are.
        The observability conditions of the remaining agents state that an agent is fully observant iff it is not distracted, so the observability type of the remaining agent will depend on the state on which the action is applied.
    \end{example}

    The next definition generalises global and local standard actions.
    \begin{definition}[Global and Local Abstract Epistemic Actions]\label{def:global-local-abstract-action}
        An abstract action $\actionStyle{a} = ((E, Q,$ $\mymathit{pre}, \mymathit{post}, \mymathit{obs}), E_d)$ of $\Lang[C]$ on $\obsTypeSet$ is called \emph{global} if $E_d$ is a singleton.
        It is called a \emph{local abstract (epistemic) action} for an observability type $t \in \obsTypeSet$, if $E_d$ is closed under $Q_t$.
        A \emph{local abstract (epistemic) action} $\actionStyle{a}$ is an abstract epistemic action that is local for some observability type.
    \end{definition}

    We now provide a definition of update for abstract epistemic actions.
    Applicability of abstract actions on states is defined as for standard epistemic actions.

    \begin{definition}[Abstract Product Update]\label{def:abstract-product-update}
        Let $\actionStyle{a} = ((E, Q, \mymathit{pre}, \mymathit{post}, \mymathit{obs}), E_d)$ be an abstract epistemic action applicable in an epistemic state $s = ((W, R, L), W_d)$.
        The \emph{abstract product update} of $s$ with $\actionStyle{a}$ is the epistemic state $s \odot \actionStyle{a} = ((W', R', L'), W'_d)$, where $W'$, $L'$ and $W'_d$ are as in Definition~\ref{def:product-update} and
        \begin{equation*}
            R'_\agentStyle{i} = \{ ((w, e), (v, f)) \in W' {\times} W' \mid w R_\agentStyle{i} v \text{ and } e Q_{\obsType{\agentStyle{i}}} f \}.
        \end{equation*}
    \end{definition}

    The only difference to the standard product update is that $R'_\agentStyle{i}$ is determined from the abstract accessibility relation $Q_{\obsType{\agentStyle{i}}}$ of the observability type of $\agentStyle{i}$ induced by state $s$.
    There is an $\agentStyle{i}$-edge from $(w, e)$ to $(v, f)$ iff there is an $\agentStyle{i}$-edge from $w$ to $v$ in $s$, and there is a $\agentStyle{t}$-edge from $e$ to $f$ in $\actionStyle{a}$, where $\agentStyle{t} = \obsType{\agentStyle{i}}$.

    \begin{figure}
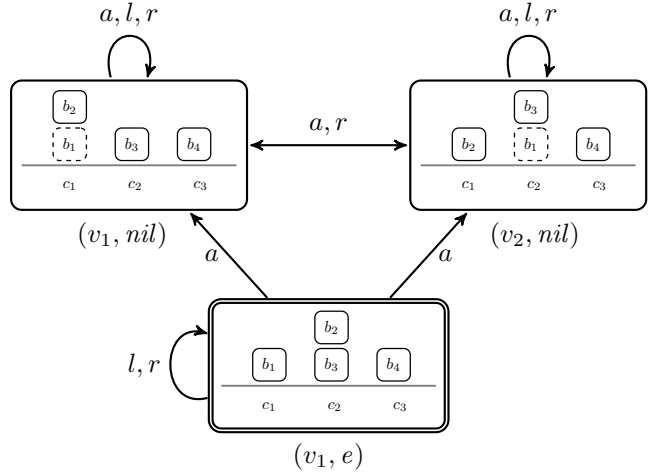

        \centering
        \begin{tikzpicture}[node distance=32pt and -16pt, >=stealth']
    \node[graphicaldworld, scale=0.6, label=below:{$(v_1, \mathit{e})$}] (v1)                      {\input{img/examples/partial-observability/moved-block.tex}};
    \node[graphicalworld, scale=0.6, label=below:{$(v_1, \mathit{nil})~~$}] (v2) [above left=of v1]  {\input{img/examples/single-agent-bw/single-agent-bw-1.tex}};
    \node[graphicalworld, scale=0.6, label=below:{$~~(v_2, \mathit{nil})$}] (v3) [above right=of v1] {\input{img/examples/single-agent-bw/single-agent-bw-2.tex}};

    \path
        (v1) edge[->,  thick, loop, out=195, in= 165, looseness=2] node[left] {$\agentStyle{l},\agentStyle{r}$} (v1)
        (v2) edge[->,  thick, loop, out=105, in= 75, looseness=4] node[above] {$\agentStyle{a},\agentStyle{l},\agentStyle{r}$} (v2)
        (v3) edge[->,  thick, loop, out=105, in= 75, looseness=4] node[above] {$\agentStyle{a},\agentStyle{l},\agentStyle{r}$} (v3)
        (v1) edge[->,  thick] node[left]  {$\agentStyle{a}$} (v2)
        (v1) edge[->,  thick] node[right] {$\agentStyle{a}$} (v3)
        (v2) edge[<->, thick] node[above] {$\agentStyle{a},\agentStyle{r}$} (v3)
    ;
\end{tikzpicture}
        \caption{State $s'_l \odot \mymathit{absPrivMove}(l, b_2, b_1, b_3)$, where $s'_l$ is the epistemic state from Example~\ref{ex:private-ontic} and the action is from Example~\ref{ex:abstract-action}.}
        \label{fig:abstract-private-ontic}
    \end{figure}
    
    \begin{example}[Abstract Product Update]\label{ex:abstract-product-update}
        Let $s'_l = ((W', R', L'), W'_d)$ be the epistemic state from Example~\ref{ex:private-ontic}.
        It is a local state for agent $l$ representing the fact that it is commonly known to all agents that both $a$ and $r$ don't know whether block $b_1$ is under $b_2$ or $b_3$.
        Moreover, assume that both worlds of $s'_l$ satisfy $\mymathit{Distracted}(a) \land \neg \mymathit{Distracted}(l) \land \neg \mymathit{Distracted}(r)$, \ie only agent $a$ is distracted, and this is commonly known by everyone.

        The abstract product update $s'_l \odot \mymathit{absPrivMove}(l, b_2, b_1, b_3) = ((W'', R'', L''), W''_d)$ is computed as follows.
        The sets of updated worlds and designated worlds, and the labels of the new worlds are calculated following the same rules of standard product update as in Example~\ref{ex:private-ontic}, and they are shown in Figure~\ref{fig:abstract-private-ontic}.
        We now turn to the accessibility relations of the new state.
        Since $s'_l \models \mymathit{Distracted}(a) \land \neg \mymathit{Distracted}(l) \land \neg \mymathit{Distracted}(r)$, we immediately get by definition of action $\mymathit{absPrivMove}(l, b_2, b_1, b_3)$ that $\obsType[]{l} = \obsType[]{r} = f$ and $\obsType[]{a} = o$, as $a$ is the only agent being distracted.
        We thus obtain the following:
        \begin{itemize}
            \item Since $v_1 R'_l v_1$, $v_1 R'_r v_1$, $e Q_f e$ and $\obsType[]{l} = \obsType[]{r} = f$, we get $(v_1, e) R''_i (v_1, e)$ for $i \in \{ l, r \}$;
            \item Since $v_1 R'_a v_1$, $v_1 R'_a v_2$, $e Q_o \mymathit{nil}$ and $\obsType[]{a} = o$, we get both $(v_1, e) R''_a (v_1, \mymathit{nil})$ and $(v_1, e) R''_a (v_2, \mymathit{nil})$; and
            \item Since $\mymathit{nil} Q_f \mymathit{nil}$ and $\mymathit{nil} Q_o \mymathit{nil}$, all edges outgoing from $(v_1, \mymathit{nil})$ and $(v_2, \mymathit{nil})$ are the same as those outgoing from $v_1$ and $v_2$, respectively.
        \end{itemize}
    \end{example}

    For a sequence $\pi = \actionStyle{a}_1, \dots, \actionStyle{a}_l$ of abstract actions and $1 \leq k \leq l$, $\pi_{\leq k}$ denotes the prefix $\actionStyle{a}_1, \dots, \actionStyle{a}_k$ of $\pi$, and $s \odot \pi$ denotes the state $s \odot \actionStyle{a}_1 \dots \odot \actionStyle{a}_l$ (if $\pi$ is empty, this is just $s$).
    We say that $\pi$ is \emph{applicable} in a state $s$ if for all $k$, $\actionStyle{a}_k$ is applicable in $s \odot \pi_{\leq k-1}$.

    \begin{definition}[Abstract Epistemic Planning Task]\label{def:abstract-planning-task}
        \label{def:solution}
        An \emph{abstract (epistemic) planning task} of $\Lang[C]$ is a triple $T = (s_0, \actionSet, \phi_g)$, where $s_0$ is a state of $\Lang[C]$ (the \emph{initial state}), $\actionSet$ is a finite set of abstract actions of $\Lang[C]$, and $\phi_g \in \Lang[C]$ is the \emph{goal formula}.
        A \emph{solution} (or \emph{plan}) to $T$ is a finite sequence $\pi$ of abstract actions of $\actionSet$ such that:
        \begin{enumerate}
            \item $\pi$ is applicable in $s_0$; and
            \item $s_0 \odot \pi \models \phi_g$.
        \end{enumerate}
    \end{definition}

    We call \emph{ADEL} the framework of \emph{Dynamic Epistemic Logic with Abstract actions}.
    In Section~\ref{sec:epddl-semantics}, we are going to build the semantics of \textsc{epddl} on the ADEL framework; namely, we are going to show how to construct an abstract planning task of ADEL from a given \textsc{epddl} specification.
    In the remainder of this section, we show that abstract and standard planning tasks are equally expressive.

    \subsection{Expressivity of Abstract Actions}
        Abstract epistemic actions can be systematically converted into standard actions (Definition~\ref{def:event-model}), and vice-versa.
        We first consider the transformation from abstract to standard actions.
        The function $\mymathit{type}_{\actionStyle{a},s}$, induced by an abstract action $\actionStyle{a}$ in an epistemic state $s$, determines the assignment of observability types to agents.
        Since this assignment depends on the state $s$, different states may yield different $\mymathit{type}_{\actionStyle{a},s}$ functions, and thus induce different standard epistemic actions from the same abstract action, as the next definition shows.

        \begin{definition}[Induced Epistemic Action]\label{def:induced-action}
            Let $\actionStyle{a} = ((E, Q, \mymathit{pre}, \mymathit{post}, \mymathit{obs}), E_d)$ be an abstract epistemic action applicable to an epistemic state $s$.
            The \emph{action induced from $\actionStyle{a}$ by $s$} is the epistemic action $\inducedAct{a}{s} = ((E, Q', \mymathit{pre}, \mymathit{post}), E_d)$, where for all $i \in \agentSet$, we let $Q'_\agentStyle{i} = \{ (e, f) \in E \times E \mid e Q_{\obsType{i}} f \}$.
        \end{definition}

        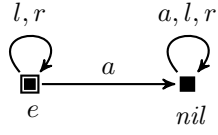
\begin{figure}
            \centering
            \begin{tikzpicture}[>=stealth']
    \node[devent,                 label=below:{$~e$}] (e) {};
    \node[event,  right=5em of e, label=below:{$~\mathit{nil}$}] (nil) {};

    \path
        (e) edge[->, thick, loop, out=130, in=50, looseness=8] node[above] {$l, r$} (e)
        (nil) edge[->, thick, loop, out=130, in=50, looseness=8] node[above] {$a, l, r$} (nil)
        (e) edge[->, thick] node[above] {$a$} (nil)
    ;
\end{tikzpicture}
            \caption{Induced epistemic action $\inducedAct{\mymathit{absPrivMove}(l, b_2, b_1, b_3)}{s'_l}$ from Example~\ref{ex:induced-action}.}
            \label{fig:induced-action}
        \end{figure}

        \begin{example}[Induced Epistemic Action]\label{ex:induced-action}
            Let $\mymathit{absPrivMove}(l, b_2, b_1, b_3)$ be the abstract epistemic action of Example~\ref{ex:abstract-action}, and let $s'_l$ be the epistemic state from Example~\ref{ex:private-ontic}.
            As we showed in the previous example, since $s'_l \models \mymathit{Distracted}(a) \land \neg \mymathit{Distracted}(l) \land \neg \mymathit{Distracted}(r)$, we immediately get by definition of action $\mymathit{absPrivMove}(l, b_2, b_1, b_3)$ that $\obsType[]{l} = \obsType[]{r} = f$ and $\obsType[]{a} = o$.
            Following Definition~\ref{def:induced-action}, we then have that the induced epistemic action $\inducedAct{\mymathit{absPrivMove}(l, b_2, b_1, b_3)}{s'_l} = ((E, Q', \mymathit{pre}, \mymathit{post}), E_d)$ is such that $E$, $\mymathit{pre}$, $\mymathit{post}$ and $E_d$ are as in $\mymathit{absPrivMove}(l, b_2, b_1, b_3)$, and the induced accessibility relations are as follows:
            \begin{itemize}
                \item $Q'_a = \{ (e, \mymathit{nil}), (\mymathit{nil}, \mymathit{nil}) \}$; and
                \item $Q'_l = Q'_r = \{ (e, e), (\mymathit{nil}, \mymathit{nil}) \}$.
            \end{itemize}
            The induced action is shown in Figure~\ref{fig:induced-action}.
        \end{example}

        We now consider the transformation from standard to abstract actions.
        For a standard epistemic action $\actionStyle{a} = ((E, Q, \mymathit{pre}, \mymathit{post}), E_d)$, let $\sim_\actionStyle{a}$ be an equivalence relation on agents such that $\agentStyle{i} \sim_\actionStyle{a} \agentStyle{j}$ iff $Q_\agentStyle{i} = Q_\agentStyle{j}$, and let $T_\actionStyle{a} = \{ [\agentStyle{i}]_{\sim_\actionStyle{a}} \mid \agentStyle{i} \in \agentSet \}$ be the set of $\sim_\actionStyle{a}$-equivalence classes.
        Two agents are $\sim_\actionStyle{a}$-equivalent if their accessibility relations in $a$ are equal.
        Intuitively, $\sim_\actionStyle{a}$-equivalent agents share the same perspective on the action.
        We are thus going to use the $\sim_\actionStyle{a}$-equivalence classes of $\obsTypeSet$ as observability types such that $[i]_a$ is the observability type of agent $j$ iff $i \sim_a j$.

        \begin{definition}[Abstraction of an Epistemic Action]\label{def:abstraction}
            Let $\actionStyle{a} = ((E, Q, \mymathit{pre}, \mymathit{post}), E_d)$ be an epistemic action.
            The \emph{abstraction of $\actionStyle{a}$} is the abstract action $\abstraction{a} = ((E, Q', \mymathit{pre}, \mymathit{post}, \mymathit{obs}), E_d)$, where $(E, Q')$ is an abstract frame on $T_\actionStyle{a}$ and:
            \begin{itemize}
                \item For all $t \in T_\actionStyle{a}$, $Q'_t = \{ (e, f) \in E \times E \mid \exists \agentStyle{i} \in t \text{ such that } e Q_\agentStyle{i} f \}$; and
                \item For all $\agentStyle{i} \in \agentSet$ and $t \in T_\actionStyle{a}$, $\mymathit{obs}_\agentStyle{i}(t) = \top$, if $\agentStyle{i} \in t$, and $\mymathit{obs}_\agentStyle{i}(t) = \bot$, otherwise.
            \end{itemize}
        \end{definition}

        For an observability type $t = \{\agentStyle{i}_1, \dots, \agentStyle{i}_n\} \in T_\actionStyle{a}$, the abstract accessibility relation $Q'_t$ contains the pairs $(e, f)$ of events that are linked in $\actionStyle{a}$ by an edge of some agent $\agentStyle{i}_k \in t$.
        For all $\agentStyle{i}_k \in t$ we have that $\mymathit{obs}_\agentStyle{i}(t) = \top$, and for all $\agentStyle{i}_k \in \agentSet \setminus t$ that $\mymathit{obs}_\agentStyle{i}(t) = \bot$.
        It is easy to check that each observability function $\mymathit{obs}_\agentStyle{i}$ satisfies the two requirements of Definition~\ref{def:abstract-event-model}, and that for any state $s$ we have $\obsType[\abstraction{a},s]{i} = t$ iff $\agentStyle{i} \in t$.
        This immediately gives the next result.

        \begin{lemma}\label{lem:induced-abstraction-identity}
            Let $s$ be a state and $\actionStyle{a}$ be an action.
            Then, $\inducedAct{(\abstraction{a})}{s} = a$.
        \end{lemma}

        We now show that abstract and standard product updates are equivalent.

        \begin{proposition}\label{prop:abstract-to-standard-update}
            Let $s$ be a state and let $\actionStyle{a}$ be an abstract action.
            Then, $s \odot a = s \otimes \inducedAct{a}{s}$.
        \end{proposition}

        \begin{proof}
            Let $s = ((W, R, L), W_d)$, $a = ((E, Q, \mymathit{pre}, \mymathit{post}, \mymathit{obs}), E_d)$, $\inducedAct{a}{s} = ((E, Q', \mymathit{pre}, \mymathit{post}),$ $E_d)$, $s \odot a = ((W', R', L'), W_d')$ and $s \otimes \inducedAct{a}{s} = ((W'', R'', L''), W_d'')$.
            By Definition~\ref{def:abstract-product-update}, we immediately have that $W' = W''$, $L' = L''$ and $W'_d = W''_d$, so we only need to show that $R'_\agentStyle{i} = R''_\agentStyle{i}$ for all $\agentStyle{i} \in \agentSet$.
            We have:
            \begin{equation*}
                \begin{array}{llll}
                    (w, e) R''_\agentStyle{i} (v, f) & \text{iff} & w R_\agentStyle{i} v \text{ and } e Q'_\agentStyle{i} f                 & \quad\text{(Def.~\ref{def:product-update})} \\
                                        & \text{iff} & w R_\agentStyle{i} v \text{ and } e Q_{\obsType[a,s]{i}} f & \quad\text{(Def.~\ref{def:induced-action})} \\
                                        & \text{iff} & (w, e) R'_\agentStyle{i} (v, f)                            & \quad\text{(Def.~\ref{def:abstract-product-update})}
                \end{array}
            \end{equation*}
        \end{proof}

        \begin{proposition}\label{prop:standard-to-abstract-update}
            Let $s$ be a state and let $\actionStyle{a}$ be an action.
            Then, $s \odot \abstraction{a} = s \otimes a$.
        \end{proposition}

        \begin{proof}
            Immediate from Lemma~\ref{lem:induced-abstraction-identity} and Proposition~\ref{prop:abstract-to-standard-update}: $s \odot \abstraction{a} = s \otimes \inducedAct{(\abstraction{a})}{s} = s \otimes a$.
        \end{proof}

        The propositions establish that product update and abstract product update yield the same result given a starting state $s$.
        Thus, starting from the same initial state, abstract and standard planning tasks are equally expressive.
        As a result, abstract planning tasks preserve the rich expressive power of standard planning tasks of DEL, while providing a cleaner and more natural representation of epistemic planning problems.
        We are going to use abstract planning tasks as a semantics for \textsc{epddl}, as we show in Section~\ref{sec:epddl-semantics}.

    \section{Syntax of \textnormal{\bf\textsc{epddl}}}\label{sec:epddl-syntax}
    Over the decades, \textsc{pddl}~\cite{misc/GhallabHKMRVWW1998} has become the de facto standard language for classical planning.
    Since its inception, several versions and expansions of the language have been proposed for different subfields of automated planning.
    All such languages share the same LISP-based syntax, which makes the \textsc{pddl} variations easier to approach and to understand for researchers and practitioners across all areas of automated planning.
    We here follow the same convention, and we base the syntax of \textsc{epddl} on the style of \textsc{pddl}, with statements surrounded by round parentheses, prefix operators/connectives, keywords prefixed by colons `:', and inline comments by semicolons `;'.

    In \textsc{pddl}, a classical planning task is described on two levels of abstraction.
    At the lower level we find the \emph{problem}, which contains the specific aspects of the planning task (objects, initial state, and goal), while on the upper level we have the \emph{domain}, which contains its universal aspects (predicates, types, actions, etc.).
    Although in principle we could adopt the same structure in \textsc{epddl}, this would result in a rather involved and error-prone syntax.
    In fact, since epistemic actions in DEL are represented by multi-pointed event models, attempting to capture all relevant information about actions within a single action declaration would easily result in code that is difficult to read, maintain, and debug.
    Moreover, it is often the case that different epistemic planning tasks contain actions that share the same perspectives of agents on the events (\ie the same accessibility relations), which if repeated in all action declarations would result in redundant code.
    
    To solve these issues, on top of problems and domains we introduce a third major component to \textsc{epddl} called \emph{action type libraries}, where we can define a set of action types.
    More precisely, by an action type we mean a multi-pointed abstract frame (Definition~\ref{def:abstract-frame}), and we say that two actions have the same type if they have the same multi-pointed abstract frame.
    Then, in domains we allow to import one or more action type libraries, meaning that the actions in those domains can be declared with one of the types described in the imported libraries.
    In this way, the definition of an abstract epistemic action is split into two parts: the types (multi-pointed abstract frames) of actions in the action type libraries, and pre-, post-, and observability conditions in the domains.
    This decomposition allows for reducing the amount of information within each action declaration, and avoids redundant definitions of abstract frames denoting the same action type, ultimately providing a clean and concise definition of epistemic planning tasks.
    
    Note that, much like domains describe aspects that are common to a set of problems, an action type library contains aspects that can be found in different domains.
    In other words, action type libraries constitute a third level of abstraction along with domains and problems.
    Furthermore, as we show in Section~\ref{sec:syntax-libraries}, action type libraries can be used to describe a wide range of DEL fragments.
    Given the richness of the DEL framework, one can also show that several epistemic planning formalisms not based on DEL can be translated into equivalent DEL-planning fragments, and thus that they are representable by appropriate action type libraries (we plan on providing a detailed account of this in future work).
    As a result, action type libraries provide a clear way to measure and compare the expressivity of epistemic planners, including those that are not based on fragments of DEL.

    In this section, we describe the syntax of problems, domains, and action type libraries in \textsc{epddl}.
    In Section~\ref{sec:syntax-ebnf}, we introduce the meta-syntax that we use to define the syntax of \textsc{epddl}.
    In Section~\ref{sec:syntax-common}, we describe formulas and list comprehensions, a novel feature introduced in \textsc{epddl}.
    Sections~\ref{sec:syntax-problems}, \ref{sec:syntax-libraries}, and \ref{sec:syntax-domains} cover problems, action type libraries, and domains, respectively.
    Finally, in Section~\ref{sec:syntax-requirements} we describe the requirements of \textsc{epddl}.
    
    \subsection{EBNF Meta-Syntax}\label{sec:syntax-ebnf}
    We describe the syntax of \textsc{epddl} using an \emph{Extended Backus-Naur Form} (\emph{EBNF}) meta-language with the following conventions.
    A \emph{meta-symbol} in EBNF is either a \emph{terminal} or a \emph{non-terminal}.
    Non-terminals can be replaced by a sequence of meta-symbols, while terminals can not.
    Non-terminals are surrounded by angled parentheses `$\langle$' and `$\rangle$', \eg the non-terminal \ebnfDecl{formula} can be used to represent formulas of our language.
    Non-terminals can also be \emph{parametrised}, denoted by \ebnfDecl{$n (x_1, \dots, x_k)$}, where each parameter $x_i$ is a meta-symbol.
    For instance, the parametrised non-terminal \ebnfDecl{list$($term$)$} can be used to represent a list of terms.
    We also make use of standard regular expressions (\emph{regex}) notation to denote patterns of characters, and we assume the reader's familiarity with such notation.
    
    An EBNF \emph{specification} is a list of \emph{production rules}, being meta-statements of the form
    \begin{equation*}
        \text{\ebnfDecl{$n (x_1, \dots, x_h)$} \ebnfProdArrow ~$\mymathit{expr}_1 \mid \dots \mid \mymathit{expr}_k$}
    \end{equation*}
    where $h \geq 0$ and $k \geq 1$, and each (possibly parametrised) \emph{expression} $\mymathit{expr}_i$ is a finite sequence of meta-symbols.
    The \emph{head} $n (x_1, \dots, x_h)$ of the production rule denotes the non-terminal being defined, the expressions $\mymathit{expr}_i$ following the production meta-symbols, `\texttt{\ebnfProdArrow}' and `\texttt{\ebnfProdOr}', denote the possible expansions of the head.
    An empty non-terminal sequence is denoted by $\epsilon$.
    For a non-terminal \ebnfDecl{$n$}, we let \ebnfDeclList{$n$}, \ebnfDeclNonEmptyList{$n$}, and \ebnfDeclOptional{$n$} denote a possibly empty list, a non-empty list, and an optional occurrence of \ebnfDecl{$n$}, respectively.
    
    Following the conventions of \textsc{pddl}, we use \emph{requirements} to specify the required set of features for a given problem, as we now explain.
    First, we define a baseline fragment of \textsc{epddl} that constitutes the baseline semantics that an epistemic planner must be able to handle, and we represent this by a ``default'' requirement (in \textsc{pddl}, the default requirement is \texttt{:strips}).
    Second, we introduce a set of \emph{requirements} to specify the set of features that extend the baseline and that are required of a planner to solve a given problem (see~Section~\ref{sec:syntax-requirements} for a complete list).
    For a requirement $r$, we write \ebnfProdArrow[$r$] ~and \ebnfProdOr[$r$] ~to denote that the expansion of the expression following the production meta-symbol is possible only if $r$ is declared among the requirements.
    We define the \textsc{epddl} baseline fragment by the following set of constraints:
    \begin{enumerate}
        \item\label{itm:baseline-1} All formulas (pre-, post-, observability conditions, goal) are conjunctions of positive literals (atoms);
        \item\label{itm:baseline-2} All models (initial state, actions) are single-pointed (and thus deterministic) S5$_n$-models;
        \item\label{itm:baseline-3} All actions are fully observable;
        \item\label{itm:baseline-4} All actions are purely epistemic.
    \end{enumerate}
    Note that the combination of items \ref{itm:baseline-2}, \ref{itm:baseline-3}, and \ref{itm:baseline-4} implies that the only type of action allowed in the baseline fragment is public announcements.
    The DEL fragment that only comprises such actions is called \emph{Public Announcement Logic} (\emph{PAL})~\cite{conf/ismis/Plaza1989,journals/synthese/Plaza2007}, one of the simplest and most investigated DEL fragments in the literature.
    Combined with item \ref{itm:baseline-1}, we obtain a fragment of PAL as our baseline fragment, namely one that allows propositional conjunction.
    The baseline isolates a rather narrow fragment of DEL: in this way, we hope to make \textsc{epddl} usable by as many planners as possible, and to facilitate the implementation of new ones.
    We use the default requirement \texttt{:pal} to denote such a baseline.
    If no requirements are provided, the default baseline requirement is assumed.

    \subsection{Formulas, Lists and List Comprehensions}\label{sec:syntax-common}
    Before introducing the main components of \textsc{epddl}, we present the syntax of the constructs that occur throughout the entire language: \emph{formulas} (Section~\ref{sec:syntax-common-formulas}), \emph{lists}, and \emph{list comprehensions} (Section~\ref{sec:syntax-common-lists}).

    \subsubsection{Formulas}\label{sec:syntax-common-formulas}
        The syntax of \textsc{epddl} formulas is as follows:

        \begin{ebnf}
            \ebnfDecl{name}                & \ebnfProdArrow                    & \ebnfToken{$[$a-zA-Z$][$a-zA-Z0-9\_\textbackslash-$]^*$} \\
            [0.4em]
            \ebnfDecl{variable}            & \ebnfProdArrow                    & \ebnfToken{?$[$a-zA-Z$][$a-zA-Z0-9\_\textbackslash-$]^*$} \\
            [0.4em]
            \ebnfDecl{term}                & \ebnfProdArrow                    & \ebnfDecl{name} \\
                                           & \ebnfProdOr                       & \ebnfDecl{variable} \\
            [0.4em]
            \ebnfDecl{predicate$(x)$}      & \ebnfProdArrow                    & \ebnfLeftPar ~\ebnfDecl{name} \ebnfDeclList{$x$} \ebnfRightPar \\
            [0.4em]
            \ebnfDecl{atomic-formula$(x)$} & \ebnfProdArrow                    & \ebnfLeftPar ~\ebnfToken{true} \ebnfRightPar \\
                                           & \ebnfProdOr                       & \ebnfLeftPar ~\ebnfToken{false} \ebnfRightPar \\
                                           & \ebnfProdOr                       & \ebnfDecl{predicate$(x)$} \\
                                           & \ebnfProdOr[:equality]            & \ebnfLeftPar ~\ebnfToken{=} \ebnfToken{$x$} \ebnfToken{$x$} \ebnfRightPar \\
                                           & \ebnfProdOr[:equality]            & \ebnfLeftPar ~\ebnfToken{/=} \ebnfToken{$x$} \ebnfToken{$x$} \ebnfRightPar \\
            [0.4em]
            \ebnfDecl{formula}             & \ebnfProdArrow                    & \ebnfDecl{atomic-formula$($term$)$} \\
            [0.4em]
                                           & \ebnfProdOr[$\mymathit{negative}\textnormal{-}\mymathit{req}$]
                                                                               & \ebnfLeftPar ~\ebnfToken{not} \ebnfDecl{formula} \ebnfRightPar \\
                                           & \ebnfProdOr                       & \ebnfLeftPar ~\ebnfToken{and} \ebnfDeclNonEmptyList{formula} \ebnfRightPar \\
                                           & \ebnfProdOr[$\mymathit{disjunctive}\textnormal{-}\mymathit{req}$]
                                                                               & \ebnfLeftPar ~\ebnfToken{or} \ebnfDeclNonEmptyList{formula} \ebnfRightPar \\
                                           & \ebnfProdOr[$\mymathit{disjunctive}\textnormal{-}\mymathit{req}$]
                                                                               & \ebnfLeftPar ~\ebnfToken{imply} \ebnfDecl{formula} \ebnfDecl{formula} \ebnfRightPar \\
            [0.4em]
                                           & \ebnfProdOr[$\mymathit{universal}\textnormal{-}\mymathit{req}$]
                                                                               & \ebnfLeftPar ~\ebnfToken{forall} \ebnfDecl{formal-params} \ebnfDecl{formula} \ebnfRightPar \\
                                           & \ebnfProdOr[$\mymathit{existential}\textnormal{-}\mymathit{req}$]
                                                                               & \ebnfLeftPar ~\ebnfToken{exists} \ebnfDecl{formal-params} \ebnfDecl{formula} \ebnfRightPar \\
            [0.4em]
                                           & \ebnfProdOr[$\mymathit{modal}\textnormal{-}\mymathit{req}$]
                                                                               & \ebnfLeftPar ~\ebnfDecl{modality} \ebnfDecl{formula} \ebnfRightPar \\
            [0.4em]
            \ebnfDecl{modality}            & \ebnfProdArrow                    & \ebnfToken{[} \ebnfDeclOptional{modality-name} \ebnfDecl{modality-index} \ebnfToken{]} \\
                                           & \ebnfProdOr                       & \ebnfToken{<} \ebnfDeclOptional{modality-name} \ebnfDecl{modality-index} \ebnfToken{>} \\
            [0.4em]
            \ebnfDecl{modality-name}       & \ebnfProdArrow[:knowing-whether]  & \ebnfToken{Kw.} \\
                                           & \ebnfProdOr[%
                                           $\begin{array}{r}
                                               \texttt{:common-knowledge }
                                               \textnormal{\emph{or}} \\
                                               \texttt{:static-common-knowledge}
                                           \end{array}$]
                                                                               & \ebnfToken{C.} \\
            [0.4em]
            \ebnfDecl{modality-index}      & \ebnfProdArrow                    & \ebnfDecl{term} \\
                                           & \ebnfProdOr[:group-modalities]    & \ebnfDecl{agent-group} \\
                                           & \ebnfProdOr[:group-modalities]    & \ebnfToken{All} \\
            [0.4em]
            \ebnfDecl{agent-group}         & \ebnfProdArrow                    & \ebnfDecl{list$($non-empty-term-seq$)$} \\
            [0.4em]
            \ebnfDecl{non-empty-term-seq}  & \ebnfProdArrow                    & \ebnfLeftPar ~\ebnfDeclNonEmptyList{term} \ebnfRightPar
        \end{ebnf}

        A \emph{name} is defined by a regex starting either with a letter, followed by letters, digits, underscores, and hyphens.\footnote{Reserved \textsc{epddl} keywords can not be used as names.}
        Names in \textsc{epddl} are case-sensitive, so \eg \texttt{foo} and \texttt{Foo} are different names.
        A \emph{variable} name starts with a question mark, followed by the same regex of names.
        Note that a regex is considered an atomic lexical unit, meaning that `\texttt{?x}' is a syntactically correct variable name, but `\texttt{?~x}' is not.
        A \emph{term} is either a name or a variable.
        A \emph{predicate on $x$} is a name followed by a possibly empty sequence of $x$ elements, all surrounded by parentheses.
        An \emph{atomic formula} is either \texttt{(true)}, \texttt{(false)} (representing $\top$ and $\bot$, respectively), a predicate on terms, or an (in)equality check between terms (requires \texttt{:equality}).
        \emph{Formulas} are built from atomic formulas using the standard propositional connectives \texttt{not}, \texttt{and}, \texttt{or} and \texttt{imply}, propositional quantifiers \texttt{forall} and \texttt{exists} (formal parameters are described below), and \emph{modalities}, with the appropriate requirements.
        Negative formulas need the $\mymathit{negative}\textnormal{-}\mymathit{req}$ requirement, being one of the following, depending on the context in which the formula occurs: \texttt{:negative-preconditions}, \texttt{:negative-postconditions}, \texttt{:negative-obs-conditions}, \texttt{:negative-goals}, \texttt{:negative-list-formulas}.
        Note that $\mymathit{negative}\textnormal{-}\mymathit{req}$ implies $\mymathit{disjunctive}\textnormal{-}\mymathit{req}$, as disjunction is defined from a combination of negation and conjunction.
        Similarly to negations, disjunctions need $\mymathit{disjunctive}\textnormal{-}\mymathit{req}$, implications need $\mymathit{negative}\textnormal{-}\mymathit{req}$, universally and existentially quantified formulas need $\mymathit{universal}\textnormal{-}\mymathit{req}$ and $\mymathit{existential}\textnormal{-}\mymathit{req}$, respectively, and modal formulas need $\mymathit{modal}\textnormal{-}\mymathit{req}$.

        A modality is an optional \emph{modality name} followed by a \emph{modality index}, all surrounded either by squared brackets (for the primal modalities $\B{}$, $\Kw{}$ and $\CK{}$), or by angled brackets (for the dual modalities $\D{}$, $\Kwd{}$ and $\CKd{}$).
        A modality name is either \texttt{Kw.}, for knowing whether formulas (requires \texttt{:knowing-whether}), or \texttt{C.}, for common knowledge formulas (requires either \texttt{:common-knowledge} or \texttt{:static-common-knowledge}).
        Static common knowledge refers to common knowledge of formulas whose truth value does not change over time, for instance formulas that describe the disposition of rooms in a building.
        The requirement for static common knowledge is separated from the one for standard common knowledge as different epistemic planners in the literature support the former, but not the latter, \eg~\cite{conf/aips/KominisG15,conf/ijcai/HuangFW017}.
        We provide more details about static common knowledge in Sections~\ref{sec:syntax-common-lists} and~\ref{sec:semantics-init-goal}.
        If the modality name is omitted, then we assume that the standard modalities ($\B{}$ or $\D{}$) are being referred to.
        In the future, we plan to extend \textsc{epddl} with additional modalities and to include user-defined modalities.
        A modality index is either a term, an \emph{agent group}, described below, or the reserved name \ebnfToken{All}, denoting the set of all agents (the last two require \texttt{:group-modalities}).
        An agent group is a list of non-empty sequences of terms.
        We now give some examples.
        Formulas of the form $\B{i} \phi$ and $\D{i} \phi$ are represented by \texttt{([i] phi)} and \texttt{(<i> phi)}, respectively (where \texttt{phi} denotes the \textsc{epddl} representation of $\phi$).
        For instance, the formula $\B{l} \mymathit{on}(b_2, b_1)$ can be represented by \texttt{([l] (on b2 b1))}.
        Similarly, \texttt{([Kw.\ i] phi)} and \texttt{(<Kw.\ i> phi)} represent $\Kw{i} \phi$ and $\Kwd{i} \phi$, respectively.
        If $G$ is a group of agents, the \textsc{epddl} formulas \texttt{([G] phi)}, \texttt{(<G> phi)}, \texttt{([Kw.\ G] phi)}, \texttt{(<Kw.\ G> phi)}, \texttt{([C.\ G] phi)} and \texttt{(<C.\ G> phi)} represent logical formulas $\B{G} \phi$, $\D{G} \phi$, $\Kw{G} \phi$, $\Kwd{G} \phi$, $\CK{G} \phi$ and $\CKd{G} \phi$, respectively.
        Note that $\B{}$, $\D{}$, $\Kw{}$ and $\Kwd{}$ can be both single-agent and group modalities, while $\CK{}$ and $\CKd{}$ can only be used as group modalities.

    \subsubsection{Lists and Lists Comprehensions}\label{sec:syntax-common-lists}
        The syntax of lists and list comprehensions is as follows:

        \begin{ebnf}
            \ebnfDecl{list$(x)$}                      & \ebnfProdArrow                    & \ebnfToken{$x$} \\
                                                      & \ebnfProdOr[:lists]               & \ebnfLeftPar ~\ebnfToken{:and} \ebnfDeclNonEmptyList{list$(x)$} \ebnfRightPar \\
                                                      & \ebnfProdOr[:lists]               & \ebnfLeftPar ~\ebnfToken{:forall} \ebnfDecl{formal-params} \ebnfDecl{list$(x)$} \ebnfRightPar \\
            [0.4em]
            \ebnfDecl{formal-params}                  & \ebnfProdArrow                    & \ebnfLeftPar ~\ebnfDecl{typed-list$($variable, type$)$} \ebnfRightPar \\
                                                      & \ebnfProdOr[:list-comprehensions] & \ebnfLeftPar ~\ebnfDecl{typed-list$($variable, type$)$} \\
                                                      &                                   & \quad \ebnfToken{|} \ebnfDecl{propositional-formula} \ebnfRightPar \\
            [0.4em]
            \ebnfDecl{typed-list$(x, \mymathit{type})$} & \ebnfProdArrow                    & \ebnfList{$x$} \\
                                                      & \ebnfProdOr                       & \ebnfNonEmptyList{$x$} \ebnfToken{-} $\mymathit{type}$ \ebnfDecl{typed-list$(x, \mymathit{type})$} \\
            [0.4em]
            \ebnfDecl{type}                           & \ebnfProdArrow                    & \ebnfDecl{primitive-type} \\
                                                      & \ebnfProdOr[:typing]              & \ebnfLeftPar ~\ebnfToken{either} \ebnfDeclNonEmptyList{primitive-type} \ebnfRightPar \\
            [0.4em]
            \ebnfDecl{primitive-type}                 & \ebnfProdArrow                    & \ebnfToken{entity} \\
                                                      & \ebnfProdOr                       & \ebnfToken{object} \\
                                                      & \ebnfProdOr                       & \ebnfToken{agent} \\
                                                      & \ebnfProdOr                       & \ebnfToken{agent-group} \\
                                                      & \ebnfProdOr                       & \ebnfToken{world} \\
                                                      & \ebnfProdOr                       & \ebnfToken{event} \\
                                                      & \ebnfProdOr                       & \ebnfToken{obs-type} \\
                                                      & \ebnfProdOr[:typing]              & \ebnfDecl{name} \\
            [0.4em]
            \ebnfDecl{propositional-formula}          & \ebnfProdArrow                    & \emph{\small See below.}
        \end{ebnf}
    
        A \emph{list of $x$} is either a \emph{singleton} list containing an $x$ element, a \emph{concatenation} (\texttt{:and}) of lists of $x$, or by an \emph{universally quantified list} (\texttt{:forall}).
        The latter comprises the \emph{formal parameters} of the quantification followed by a list of $x$.
        Formal parameters are either a \emph{typed list} of variables (as in standard \textsc{pddl}), or one followed by a \emph{propositional formula}\footnote{Propositional \textsc{epddl} formulas only contain propositional connectives and/or propositional quantifiers, and they are defined similarly to standard \textsc{epddl} formulas. For this reason, we omit their EBNF representation.} specifying the condition required of the elements of the list (requires \texttt{:list-comprehensions}).
        We call the latter expression a \emph{list comprehension}, defined below.
        Lists are also present in conditional effects of standard \textsc{pddl}~\cite{misc/GhallabHKMRVWW1998}, although with a slightly different syntax, where terminals \texttt{and} and \texttt{forall} are used instead of \texttt{:and} and \texttt{:forall}.
        Here, we part from \textsc{pddl} lists, while simultaneously maintaining a similar syntax, for two main reasons.
        First, we obtain a clear differentiation between propositional conjunction and universal quantification, and list concatenation and quantification.
        Second, we avoid syntactic ambiguity in statements involving lists of formulas (appearing in Section~\ref{sec:syntax-init-goal}).
        In fact, if terminal \texttt{and} were used both for \textsc{epddl} formulas and lists, then it would be non-trivial, or even unfeasible, to distinguish between the conjunction of formulas and lists of formulas.
        For instance, the statement \texttt{(and (p) (q) (r))} could be used to represent both the formula $\mathtt{p} \land \mathtt{q} \land \mathtt{r}$, and the list $\langle \mathtt{p}, \mathtt{q}, \mathtt{r} \rangle$.
        A similar observation can also be made for terminal \texttt{forall}.

        A typed list on $x$ and $\mymathit{type}$ is a sequence of $x$ elements, each possibly followed by a \emph{type} element, representing their type.
        A \emph{type} is either a \emph{primitive type}, or a \emph{composite type} of the form \texttt{(either t1 ... tk)}, where \texttt{t1}, \dots, \texttt{tk} are primitive types.
        A primitive type is either a name, or one of the following reserved types: \texttt{entity}, \texttt{object}, \texttt{agent}, \texttt{agent-group}, \texttt{world}, \texttt{event}, \texttt{obs-type}.
        Both non-reserved primitive types and composite types require \texttt{:typing}.
        Without this requirement, only reserved primitive types are allowed.
        Type \texttt{entity} is equivalent to \texttt{(either object agent agent-group)}, and types \texttt{world}, \texttt{event}, and \texttt{obs-type} can not be specialised.
        More details on types can be found in Section~\ref{sec:syntax-types-predicates-const}.

        List comprehensions are a novel feature of \textsc{epddl}, inspired by functional languages like Haskell.
        List comprehensions provide a concise and powerful syntax for defining lists of syntactic elements satisfying complex conditions, here also called \emph{list formulas}, a feature that can be useful, or even necessary, in many situations.
        In \textsc{epddl}, list comprehensions are used in different parts of a specification, \eg in quantified formulas, and actions' parameters (see Section~\ref{sec:syntax-actions}).
        In the former case they are used to define the ground lists of parameters of a formula, and in the latter to specify the ground parameters to instantiate an action schema.
        Note that, as they are intended to specify which ground values can be assigned to parameters, lists formulas need to be evaluated at \emph{grounding time}.
        For instance, consider the quantified formula \texttt{(forall (?b - block | (/= ?b b1)) (clear ?b))}, expressing the fact that all blocks except for $b_1$ are clear.
        Evaluating the condition \texttt{(/= ?b b1)} on the set of blocks of Example~\ref{ex:multi-agent-bw} induces the ground list \texttt{(b2 b3 b4)} of objects of type \texttt{block}, from which we obtain the ground formula $\mathtt{clear}_{\mathtt{b_2}} \land \mathtt{clear}_{\mathtt{b_3}} \land \mathtt{clear}_{\mathtt{b_4}}$ (see Section~\ref{sec:semantics-formulas} for more details).

        To enforce that list comprehensions denote concrete, objective collections, we only allow them to express factual, objective conditions.
        For this reason, we introduce \emph{facts}, a novel feature of \textsc{epddl}: facts are objectively true/false predicates (\ie their truth value does not depend on any agent's perspective) that are both static (\ie truth value can not be changed by actions' effects) and commonly known by all agents.
        Facts can be used to represent public and immutable information, like the disposition of rooms in a building, or the graph describing a map where the agents can move.
        List formulas must be built solely from such facts (and the usual propositional connectives/quantifiers over them).
        Note that modalities are therefore forbidden inside list formulas, since they express subjective beliefs rather than objective facts.
        Requirements for list formulas are defined as for standard formulas (except for modal formulas, which are not allowed).
        Restricting to factual conditions is paramount to ensure that list formulas are well-defined.
        For instance, if in the list comprehension above we replaced the condition \texttt{(/= ?b b1)} with \texttt{(on ?b c1)}, we would obtain that the generation of the quantified formula is dependent on the position of the blocks, which is not fixed.
        As a result, the list induced by the modified list comprehension would not be well-defined.
        Moreover, facts allow for a clean and precise characterization of static common knowledge:
        A common-knowledge formula is called \emph{static} if the only predicates occurring in it are static ones.
        Hence, if common knowledge is used in an \textsc{epddl} problem or domain, we can distinguish whether requirement \texttt{:common-knowledge} or \texttt{:static-common-knowledge} is needed.
        If only static common-knowledge formulas are used, then the latter requirement is sufficient, while if some standard (non-static) predicate occurs in some common-knowledge formulas, the former is necessary.

        Facts can be declared in an \textsc{epddl} domain, as shown in Section~\ref{sec:syntax-types-predicates-const}, akin to standard predicates, and the set of true facts can be declared in a problem (see Section~\ref{sec:syntax-init-goal}).
        On this set of propositional atoms are evaluated the (propositional) conditions of list comprehensions at grounding time, as detailed in Section~\ref{sec:semantics-formulas}.

        We now show some examples.
        Suppose we are modelling a planning domain involving students, teachers and different kinds of courses.
        In this scenario, we might be interested in specifying the fact that professor Smith and all computer science students know that the algorithms course is taught by professor Smith.
        This can be represented by the formula:
        \begin{lstlisting}[language=EPDDL,label={lst:formula},caption={Example of \textsc{epddl} formula where a group modality is defined via a list comprehension. Reserved keywords are colored in blue.},captionpos={b}]
([
  (:and
    (Smith)
    (:forall (?i - student | (enrolled-in ?i comp-science))
      (?i)
    )
  )
] (teaches Smith algorithms)
)
        \end{lstlisting}
        The list of agents between squared brackets is built by concatenating the singleton list \texttt{(Smith)} with a list produced by the list comprehension of agents of type \texttt{student} that are enrolled in the computer science program.
        Here we assume that \texttt{enrolled-in} is a fact.
        Although list comprehensions provide a clean and powerful syntax for defining lists, using them for specifying agent groups directly within formulas modalities can lead to less readable code.
        To avoid this issue, we introduce agent groups definition in \textsc{epddl} problems, as we show in the next section.

    \subsection{Problems}\label{sec:syntax-problems}
    As in standard \textsc{pddl}, problems in \textsc{epddl} describe the specific elements of a planning task.
    These are objects, agents, agent groups, initial state, initialization of facts, and goal formula:

    \begin{ebnf}
        \ebnfDecl{problem}         & \ebnfProdArrow                  & \ebnfLeftPar ~\ebnfToken{define} \ebnfLeftPar ~\ebnfToken{problem} \ebnfDecl{name} \ebnfRightPar \\
                                   &                                 & \quad \ebnfLeftPar ~\ebnfToken{:domain} \ebnfDecl{name} \ebnfRightPar \\
                                   &                                 & \quad \ebnfDeclList{problem-item} \ebnfRightPar \\
        [0.4em]
        \ebnfDecl{problem-item}    & \ebnfProdArrow                  & \ebnfDecl{require-decl} \\
                                   & \ebnfProdOr                     & \ebnfDecl{objects-decl} \\
                                   & \ebnfProdOr                     & \ebnfDecl{agents-decl} \\
                                   & \ebnfProdOr[:agent-groups]      & \ebnfDecl{agent-groups-decl} \\
                                   & \ebnfProdOr                     & \ebnfDecl{init} \\
                                   & \ebnfProdOr[:facts]             & \ebnfDecl{facts-init} \\
                                   & \ebnfProdOr                     & \ebnfDecl{goal}
    \end{ebnf}

    Problem items may appear in any order in the problem description, and multiple declarations of requirements, objects, agents, agent groups and goals are allowed.
    In Sections~\ref{sec:semantics-formulas} and~\ref{sec:semantics-init-goal} we describe how multiple declarations are handled.
    There must be a unique declaration of the initial epistemic state, and at most one facts initialization.
    Objects, agents and agent groups are discussed in Section~\ref{sec:syntax-objects-agents-groups}, initial states and goals in Section~\ref{sec:syntax-init-goal}, and facts initialization in Section~\ref{sec:syntax-facts-init}.
    Requirements are discussed in Section~\ref{sec:syntax-requirements}.

    \subsubsection{Objects, Agents and Agent Groups}\label{sec:syntax-objects-agents-groups}
    The declarations of objects, agents and agent groups in \textsc{epddl} are as follows:
    
    \begin{ebnf}
        \ebnfDecl{objects-decl}                             & \ebnfProdArrow & \ebnfLeftPar ~\ebnfToken{:objects} \\
                                                            & \ebnfProdOr    & \quad \ebnfDecl{typed-list$($name, primitive-type$)$} \ebnfRightPar \\
        [0.4em]
        \ebnfDecl{agents-decl}                              & \ebnfProdArrow & \ebnfLeftPar ~\ebnfToken{:agents} \\
                                                            & \ebnfProdOr    & \quad \ebnfDecl{non-empty-typed-list$($name, primitive-type$)$} \ebnfRightPar \\
        [0.4em]
        \ebnfDecl{non-empty-typed-list$(x, \mymathit{type})$} & \ebnfProdArrow & \ebnfNonEmptyList{$x$} \\
                                                            & \ebnfProdOr    & \ebnfNonEmptyList{$x$} \ebnfToken{-} $\mymathit{type}$ \ebnfDecl{non-empty-typed-list$(x, \mymathit{type})$} \\
        [0.4em]
        \ebnfDecl{agent-groups-decl}                        & \ebnfProdArrow & \ebnfLeftPar ~\ebnfToken{:agent-groups} \ebnfDeclList{agent-group-decl} \ebnfRightPar \\
        [0.4em]
        \ebnfDecl{agent-group-decl}                         & \ebnfProdArrow & \ebnfLeftPar ~\ebnfDecl{name} \ebnfDeclOptional{agent-group-type} \ebnfDecl{agent-group} \ebnfRightPar \\
        [0.4em]
        \ebnfDecl{agent-group-type}                         & \ebnfProdArrow & \ebnfToken{-} \ebnfDecl{primitive-type}
    \end{ebnf}

    The definition of objects is the standard one in \textsc{pddl}.
    Agents are defined in a similar fashion, and we require that a problem specifies at least one agent.
    If an object is declared with no type, then it is assumed that its type is \texttt{object}.
    Similarly, agents declared with no type are implicitly considered of type \texttt{agent} (see Section~\ref{sec:syntax-types-predicates-const} for more details).
    Both objects and agents can be declared with their own types (see Section~\ref{sec:syntax-types-predicates-const} for more details on types).
    Agent groups are defined by assigning a name (the group name) to a list of names (the agents in the group), and possibly by specifying their type.
    If no type is provided, it is assumed to be \texttt{agent-group}.
    Continuing the toy example from the previous section, we now show a (partial) instance of an \textsc{epddl} problem:
    \begin{lstlisting}[language=EPDDL,label={lst:dummy-problem},caption={Example of declaration of objects, agents and agent groups in an \textsc{epddl} problem.},captionpos={b}]
(define (problem courses-1)
  (:domain courses)

  (:requirements :typing :list-comprehensions :agent-groups)

  (:objects
    lecture-room1 lecture-room2 plenary-room room
    comp-science engineering economics - programme
    algorithms analysis planning - courses
  )

  (:agents
    Anne Bob Carl Daphne - student
    Smith Jones - professor
  )

  (:agent-groups
    (CS-students - students
      (:forall (?i - student | (enrolled-in ?i comp-science))
        (?i)
      )
    )
    (CS-students-Smith
      (:and (CS-students) (Smith))
    )
  )
  ...
)
    \end{lstlisting}

    The problem defines several objects and agents, together with their types.
    For instance, \texttt{comp-science} is an object of type \texttt{programme} and \texttt{Anne} is an agent of type \texttt{student}.
    We adopt and encourage the following naming convention: names beginning with a lower-case letter are used for objects, while those starting with an upper-case letter are used for agents and agent groups.
    The problem then defines two agent groups: \texttt{CS-students}, which contains all and only those entities of type \texttt{student} that are enrolled in the \texttt{comp-science} course (again, we assume that \texttt{enrolled-in} is a fact), and \texttt{CS-students-Smith}, which contains all the entities from \texttt{CS-students}, together with \texttt{Smith}.
    With these new definitions, we can rewrite the formula from Listing~\ref{lst:formula} in the following simplified version:
    \begin{lstlisting}[language=EPDDL,label={lst:formula-simplified},caption={Simplified \textsc{epddl} formula.},captionpos={b}]
( [CS-students-Smith] (teaches Smith algorithms) )
    \end{lstlisting}

    \subsubsection{Initial State and Goal}\label{sec:syntax-init-goal}
    Initial states and goal formulas are defined as follows:

    \begin{ebnf}
        \ebnfDecl{goal}             & \ebnfProdArrow                     & \ebnfLeftPar ~\ebnfToken{:goal} \ebnfDecl{formula} \ebnfRightPar \\
        [0.4em]
        \ebnfDecl{init}             & \ebnfProdArrow                     & \ebnfLeftPar ~\ebnfToken{:init} \ebnfDecl{epistemic-state} \ebnfRightPar \\
                                    & \ebnfProdOr[:finitary-S5-theories] & \ebnfLeftPar ~\ebnfToken{:init} \ebnfDecl{list$($finitary-s5-formula$)$} \ebnfRightPar \\
    \end{ebnf}

    Goals in \textsc{epddl} are defined as general formulas.
    Initial epistemic states can be defined in two distinct ways.
    The first one is an explicit definition, where the initial state is described in terms of (designated) worlds, accessibility relations and labels, as follows:

    \begin{ebnf}
        \ebnfDecl{epistemic-state}     & \ebnfProdArrow                     & \ebnfDecl{worlds} \\
                                       &                                    & \ebnfDecl{relations$($term$)$} \\
                                       &                                    & \ebnfDecl{labels} \\
                                       &                                    & \ebnfDecl{designated$($name$)$} \\
        [0.4em]
        \ebnfDecl{worlds}              & \ebnfProdArrow                     & \ebnfToken{:worlds} \ebnfLeftPar ~\ebnfDeclNonEmptyList{name} \ebnfRightPar \\
        [0.4em]
        \ebnfDecl{relations$(x)$}      & \ebnfProdArrow                     & \ebnfToken{:relations} \ebnfLeftPar ~\ebnfDeclList{agent-relation$(x)$} \ebnfRightPar \\
        [0.4em]
        \ebnfDecl{agent-relation$(x)$} & \ebnfProdArrow                     & \ebnfDecl{name} \ebnfDecl{list$($pair$(x))$} \\
        [0.4em]
        \ebnfDecl{pair$(x)$}           & \ebnfProdArrow                     & \ebnfLeftPar ~\ebnfToken{$x$} \ebnfToken{$x$} \ebnfRightPar \\
        [0.4em]
        \ebnfDecl{labels}              & \ebnfProdArrow                     & \ebnfToken{:labels} \ebnfLeftPar ~\ebnfDeclList{world-label} \ebnfRightPar \\
        [0.4em]
        \ebnfDecl{world-label}         & \ebnfProdArrow                     & \ebnfDecl{name} \ebnfDecl{list$($predicate$($name$))$} \\
        [0.4em]
        \ebnfDecl{designated$(x)$}     & \ebnfProdArrow                     & \ebnfToken{:designated} \ebnfLeftPar ~\ebnfToken{$x$} \ebnfRightPar \\
                                       & \ebnfProdOr[:multi-pointed-models] & \ebnfToken{:designated} \ebnfLeftPar ~\ebnfNonEmptyList{$x$} \ebnfRightPar \\
    \end{ebnf}

    In an explicit definition of an initial state, we specify a non-empty sequence of world names, a sequence of \emph{agent accessibility relations}, the \emph{world labels}, and either a single designated world name, or a non-empty sequence of designated world names (the latter requires \texttt{:multi-pointed-models}).
    An agent accessibility relation is an agent name followed by a \emph{list of term-pairs}.
    A world label is a world name followed by a list of predicates (the predicates being true in the world).
    The following problem contains an explicit definition of the epistemic state of Example~\ref{ex:multi-agent-bw-local}:

    \begin{lstlisting}[language=EPDDL,label={lst:init-explicit},caption={Explicit definition of the epistemic state of Example~\ref{ex:multi-agent-bw-local}.},captionpos={b}]
(define (problem ebw1)
  (:domain epistemic-blocks-world)

  (:requirements :typing :agent-groups)

  (:objects
    b1 b2 b3 b4 - block
    c1 c2 c3 - column
  )

  (:agents A L R)

  ; (:agent-groups (All (A L R)))

  (:init
    :worlds (w1 w2 w3)
    :relations ( A (:forall (?w ?v - world) (?w ?v))
                 L (:and (w1 w1) (w2 w2) (w2 w3) (w3 w2) (w3 w3))
                 R (:and (w1 w1) (w1 w2) (w2 w1) (w2 w2) (w3 w3)) )
    :labels ( w1 (:and (on b1 c1) (on b2 b1) (on b3 c2) (on b4 c3)
                       (clear b2) (clear b3) (clear b4))
              w2 (:and (on b1 c2) (on b2 c1) (on b3 b1) (on b4 c3)
                       (clear b2) (clear b3) (clear b4))
              w3 (:and (on b1 c3) (on b2 c1) (on b3 c2) (on b4 b1)
                       (clear b2) (clear b3) (clear b4)) )
    :designated (w1 w2)
  )
  
  (:goal
    ([C. All] (on b2 b1))
  )
)
    \end{lstlisting}

    An explicit encoding of an initial epistemic state can be tedious and error-prone for large models.
    To ease modelling, \textsc{epddl} supports a compact syntactic alternative called \emph{finitary S5-theories} (requires \texttt{:finitary-S5-theories})~\cite{conf/jelia/SonPBG14}.
    A finitary S5-theory is a finite set of formulas of restricted form from which an S5 epistemic state is constructed.
    Compared with an explicit description, finitary S5-theories are typically more concise and more intuitive, but they capture only a subclass of S5-states; specifically, those in which all worlds have distinct labels (see~\cite{conf/jelia/SonPBG14}).
    This restriction notwithstanding, finitary S5-theories are sufficient for many common benchmark initial states.
    Explicit state definitions remain available (and are fully general), but they scale poorly as model size grows.
    We plan to extend the syntactic representation in future work to cover a broader class of initial states.

    Formally, a (primitive) finitary S5-theory $\Phi$ is a finite set of formulas such that:
    \begin{itemize}
        \item Formulas are of the forms:
        \begin{inparaenum}
            \item\label{itm:s5-theory-1} $\phi$;
            \item\label{itm:s5-theory-2} $\CK{\agentSet}(\B{i}\phi)$;
            \item\label{itm:s5-theory-3} $\CK{\agentSet}(\Kw{i}\phi)$; and
            \item\label{itm:s5-theory-4} $\CK{\agentSet}(\Kwd{i}\phi)$
        \end{inparaenum}
        (where $\phi$ is a propositional formula).
        \item For all agents $\agentStyle{i} \in \agentSet$, we have that each atom $\fluentStyle{p} \in \atomSet$ occurs in at least a formula of types \ref{itm:s5-theory-2}, \ref{itm:s5-theory-3}, or \ref{itm:s5-theory-4}.
    \end{itemize}
    We assume that theories are consistent: for all $\phi,\phi' \in \Phi$ the conjunction $\phi \land \phi'$ is satisfiable.
    Since finitary S5-theories represent S5-states, following \cite{conf/jelia/SonPBG14} formulas of type~\ref{itm:s5-theory-2} are often abbreviated as $\CK{\agentSet}\phi$ (in S5 the T axiom ensures that $\B{i}\phi$ implies $\phi$).
    Intuitively, type~\ref{itm:s5-theory-1} formulas specify the propositional facts that hold in designated worlds of the state; types~\ref{itm:s5-theory-2} and~\ref{itm:s5-theory-3} describe what agents know; and type~\ref{itm:s5-theory-4} captures what agents are uncertain about.
    For compactness, if for an agent $\agentStyle{i}$ an atom $\fluentStyle{p}$ does not occur in any of that agent's formulas of types~\ref{itm:s5-theory-2}-\ref{itm:s5-theory-4}, we assume $\CK{\agentSet}(\neg\Kw{i}\fluentStyle{p}) \in \Phi$, \ie agents are by default uncertain about atoms that are not mentioned for them.
    Similarly, if an atom does not appear in any type~\ref{itm:s5-theory-1} formula, we assume $\neg\fluentStyle{p} \in \Phi$ (the atom is false).
    
    The syntax for formulas of finitary S5-theories is as follows, where \emph{predicate formulas} are propositional formulas where (in)equalities and \texttt{(true)}/\texttt{(false)} formulas do \emph{not} occur:

    \begin{ebnf}
        \ebnfDecl{finitary-s5-formula} & \ebnfProdArrow & \ebnfDecl{predicate-formula} \\
                                       & \ebnfProdOr    & \ebnfLeftPar ~\ebnfToken{[} \ebnfToken{C.} \ebnfToken{All} \ebnfToken{]} \ebnfDecl{predicate-formula} \ebnfRightPar \\
                                       & \ebnfProdOr    & \ebnfLeftPar ~\ebnfToken{[} \ebnfToken{C.} \ebnfToken{All} \ebnfToken{]} \ebnfLeftPar ~\ebnfToken{[} \ebnfDecl{term} \ebnfToken{]} \ebnfDecl{predicate-formula} \ebnfRightPar ~\ebnfRightPar \\
                                       & \ebnfProdOr    & \ebnfLeftPar ~\ebnfToken{[} \ebnfToken{C.} \ebnfToken{All} \ebnfToken{]} \ebnfLeftPar ~\ebnfToken{[} \ebnfToken{Kw.} \ebnfDecl{term} \ebnfToken{]} \ebnfDecl{predicate-formula} \ebnfRightPar ~\ebnfRightPar \\
                                       & \ebnfProdOr    & \ebnfLeftPar ~\ebnfToken{[} \ebnfToken{C.} \ebnfToken{All} \ebnfToken{]} \ebnfLeftPar ~\ebnfToken{<} \ebnfToken{Kw.} \ebnfDecl{term} \ebnfToken{>} \ebnfDecl{predicate-formula} \ebnfRightPar ~\ebnfRightPar \\
        [0.4em]
        \ebnfDecl{predicate-formula}   & \ebnfProdArrow & \emph{\small See below.}
    \end{ebnf}

    The epistemic state of Example~\ref{ex:multi-agent-bw-local} can be defined in \textsc{epddl} by the following finitary S5-theory.

    \begin{lstlisting}[language=EPDDL,label={lst:init-finitary-s5-theory},caption={Altertative definition of the epistemic state of Example~\ref{ex:multi-agent-bw-local} via a finitary S5-theory.},captionpos={b}]
  (:init
    (:and
      ; 0. Actual configurations of the blocks
      (and (clear b2) (clear b3) (clear b4)
        (or (and (on b1 c1) (on b2 b1) (on b3 c2) (on b4 c3))
            (and (on b1 c2) (on b2 c1) (on b3 b1) (on b4 c3)) ))
      ; It is commonly known that:
      ; 1. Only the following configurations are possible
      ([C. All] (and
        (clear b2) (clear b3) (clear b4)
        (or (and (on b1 c1) (on b2 b1) (on b3 c2) (on b4 c3))
            (and (on b1 c2) (on b2 c1) (on b3 b1) (on b4 c3))
            (and (on b1 c3) (on b2 c1) (on b3 c2) (on b4 b1)) )))
      ; 2. Agent A doesn't know where block b1 is (implicit)
      ; (:forall (?b - block | (/= ?b b1))
      ;   ([C. All] (<Kw. A> (on ?b b1))) )
      ; 3. Agent L knows whether b2 is on b1
      ([C. All] ([Kw. L] (on b2 b1)))
      ; 4. Agent R knows whether b4 is on b1
      ([C. All] ([Kw. R] (on b4 b1)))
    )
  )
    \end{lstlisting}

    We now briefly describe how state $s_r$ of Example~\ref{ex:multi-agent-bw-local} is described by the above finitary S5-theory as follows.
    See Section~\ref{sec:semantics-init-goal} and Example~\ref{ex:semantic-init-finitary} for a complete explanation of the semantics of finitary S5-theories.
    Recall that state $s_r$ is a local state for agent $r$, namely it represents the situation from agent $r$'s perspective.
    First, the possible actual configurations of the blocks that are consistent with the perspective of agent $r$ are defined by formula $0$, which states that all blocks except for $b_1$ are clear and that $b_1$ is either on the first or second column.
    Note that formula $0$ is satisfied in the actual worlds $w_1$ and $w_2$ of $s_r$.
    Second, the theory specifies that it is commonly known by all agents that blocks $\objectStyle{b}_2$, $\objectStyle{b}_3$, and $\objectStyle{b}_4$ are clear, and that block $\objectStyle{b}_1$ is under either remaining blocks (formula $1$).
    This describes the set of possible configurations of the blocks, \ie the set of worlds of the initial state.
    As per our convention, we do not need to explicitly specify that agent $\agentStyle{a}$ does not know where block $\objectStyle{b}_1$, as this can be represented by a conjunction of type~\ref{itm:s5-theory-4} formulas: $\bigwedge_{b \in \{\objectStyle{b}_2, \objectStyle{b}_3, \objectStyle{b}_4\}} \CK{\agentSet}( \Kwd{a} \mymathit{On}(b, \objectStyle{b}_1) )$.
    As a result, the theory states that the accessibility relation $R_a$ is the universal relation $W \times W$.
    Finally, the theory states that agents $l$ and $r$ respectively know whether block $\objectStyle{b}_1$ is under blocks $\objectStyle{b}_2$ (formula $3$), and $\objectStyle{b}_4$ (formula $4$).
    The last two formulas describe the accessibility relations of agents $l$ and $r$, respectively.

    \subsubsection{Facts Initialization}\label{sec:syntax-facts-init}
    Facts initialization is defined as follows:

    \begin{ebnf}
        \ebnfDecl{facts-init}       & \ebnfProdArrow      & \ebnfLeftPar ~\ebnfToken{:facts-init} \ebnfDeclList{predicate$($name$)$} \ebnfRightPar \\
    \end{ebnf}

    A facts-initialization specifies the set of (true) facts for a problem.
    Facts that do not occur in a facts-initialization statement are implicitly considered to be false.
    We recall that a fact is an objectively true/false, commonly known, static predicate (see Section~\ref{sec:syntax-common-lists}).
    Since facts are commonly known by all agents, facts initialised here hold in all possible worlds of the initial epistemic state.
    We thus separate the initialization of facts from the initial state declaration to avoid verbose and error-prone specifications, since the same (potentially long) list of facts would have to be repeated for every world label.
    When constructing the initial epistemic state, true facts are automatically inserted into the label of all worlds (see Section~\ref{sec:semantics-init-goal}).
    Consider the following example, where \texttt{(adj ?x ?y)} is a binary fact stating that rooms \texttt{?x} and \texttt{?y} are adjacent:
    \begin{lstlisting}[language=EPDDL,label={lst:facts-init},caption={Example of facts initialization.},captionpos={b}]
(:facts-init
  (adj room1 room2)
  (adj room2 room3)
)
    \end{lstlisting}
    Here are initialised the facts \texttt{(adj room1 room2)} \texttt{(adj room2 room3)}, which are both going to hold in all possible worlds of the initial state.
    All remaining facts left uninitialised (\eg \texttt{(adj room1 room3)}) are implicitly considered not to hold in any world.
    Further details are provided in Section~\ref{sec:semantics-init-goal}.

    \subsection{Action Type Libraries}\label{sec:syntax-libraries}
    Action type libraries are a novel component to \textsc{epddl}, where we define a set of action types.
    An \emph{action type} represents a multi-pointed abstract frame (see Definition~\ref{def:abstract-frame}) plus some semantic conditions.
    A library contains the definitions of one or more action types, which can be used across different \textsc{epddl} domains to define different instances of actions of those types.
    In this way, we simplify the definition of actions of the same type, as we only need to define their abstract frame once.
    Furthermore, we can use action type libraries to define different fragments of DEL-based epistemic planning: the action types described in a library are the abstract frames that are allowed in the epistemic actions of the fragment.
    In this way, we establish a unified and flexible syntax for representing a wide variety of epistemic planning formalisms, facilitating direct and meaningful comparisons between different solvers.
    The syntax of action types in \textsc{epddl} is the following:

    \begin{ebnf}
        \ebnfDecl{library}           & \ebnfProdArrow                     & \ebnfLeftPar ~\ebnfToken{define} \ebnfLeftPar \ebnfToken{ action-type-library} \ebnfDecl{name} \ebnfRightPar \\
                                     &                                    & \quad \ebnfDeclList{library-item} \ebnfRightPar \\
        [0.4em]
        \ebnfDecl{library-item}      & \ebnfProdArrow                     & \ebnfDecl{require-decl} \\
                                     & \ebnfProdOr                        & \ebnfDecl{action-type-decl} \\
        [0.4em]
        \ebnfDecl{action-type-decl}  & \ebnfProdArrow                     & \ebnfLeftPar ~\ebnfToken{:action-type} \ebnfDecl{name} \\
                                     &                                    & \quad \ebnfDecl{events} \\
                                     &                                    & \quad \ebnfDecl{obs-types} \\
                                     &                                    & \quad \ebnfDecl{relations$($variable$)$} \\
                                     &                                    & \quad \ebnfDecl{designated$($variable$)$} \\
                                     &                                    & \quad \ebnfDecl{conditions} \ebnfRightPar \\
        [0.4em]
        \ebnfDecl{events}            & \ebnfProdArrow                     & \ebnfToken{:events} \ebnfLeftPar ~\ebnfDeclNonEmptyList{variable} \ebnfRightPar \\
        [0.4em]
        \ebnfDecl{obs-types}         & \ebnfProdArrow                     & \ebnfToken{:observability-types} \ebnfDeclNonEmptyList{name} \\
        [0.4em]
        \ebnfDecl{conditions}        & \ebnfProdArrow                     & $\epsilon$ \\
                                     & \ebnfProdOr[:events-conditions]    & \ebnfToken{:conditions} \ebnfLeftPar ~\ebnfDeclList{event-conditions} \ebnfRightPar \\
        [0.4em]
        \ebnfDecl{event-conditions}  & \ebnfProdArrow                     & \ebnfDecl{variable} \ebnfDeclNonEmptyList{event-cond} \\
        [0.4em]
        \ebnfDecl{event-cond}        & \ebnfProdArrow                     & \ebnfToken{:propositional-precondition} \\
                                     & \ebnfProdOr                        & \ebnfToken{:propositional-postconditions} \\
                                     & \ebnfProdOr                        & \ebnfToken{:propositional-event} \\
                                     & \ebnfProdOr                        & \ebnfToken{:trivial-precondition} \\
                                     & \ebnfProdOr                        & \ebnfToken{:trivial-postconditions} \\
                                     & \ebnfProdOr                        & \ebnfToken{:trivial-event} \\
                                     & \ebnfProdOr                        & \ebnfToken{:non-trivial-precondition} \\
                                     & \ebnfProdOr                        & \ebnfToken{:non-trivial-postconditions} \\
                                     & \ebnfProdOr                        & \ebnfToken{:non-trivial-event}
    \end{ebnf}

    In an action type library we can define requirements (discussed in Section~\ref{sec:syntax-requirements}) and action types.
    An action type contains a non-empty sequence of \emph{event variables}, a non-empty sequence of observability type names, a sequence of abstract accessibility relations (defined as agent accessibility relations in Section~\ref{sec:syntax-init-goal}), one or more designated event variables, and possibly a sequence of \emph{event conditions} (requires \texttt{:events-conditions}).
    Event variables act as parameters (of type \texttt{event}) for an action type, so an action type functions as a template for constructing multi-pointed abstract frames (Definition~\ref{def:abstract-frame}).
    To instantiate a frame, a domain action (see Section~\ref{sec:syntax-actions}) provides a list of concrete events (Section~\ref{sec:syntax-events}), each of which is bound to one of the event variables declared by the action type.
    The instantiated frame therefore combines the template's structure (events, observability types, abstract accessibility relations, and designated events) with the specific events supplied by the action.
    It is also possible to declare some conditions for event variables.
    When an event variable \texttt{?e} is bound to a concrete event \texttt{e} during action instantiation, \texttt{e} must satisfy all event conditions declared for \texttt{?e}.
    These conditions restrict which concrete events may be used to instantiate the action type.
    Specifically, \texttt{:propositional-precondition} means that the precondition of the event must be propositional, \texttt{:propositional-postconditions} means that all of its postconditions must be propositional, and \texttt{:propositional-event} is a syntactic sugar to include both conditions.
    Similarly, \texttt{:trivial-precondition} means that the precondition of the event must be trivial (\ie \texttt{(true)}, or omitted), \texttt{:trivial-postconditions} that no effects can be declared for the event, and \texttt{:trivial-event} includes both conditions (\texttt{:non-trivial-precondition}, \texttt{:non-trivial-postconditions} and \texttt{:non-trivial-events} are defined analogously).
    Below, we give several examples of action types.

    \begin{lstlisting}[language=EPDDL,label={lst:action-types},caption={Examples of \textsc{epddl} action types.},captionpos={b}]
(define (action-type-library my-library)
  (:requirements
    :partial-observability :multi-pointed-models :events-conditions
  )

  (:action-type public-ontic
    :events     (?e)
    :observability-types (Fully)
    :relations  (Fully (?e ?e))
    :designated (?e)
    :conditions (?e :non-trivial-postconditions)
  )

  (:action-type private
    :events     (?e ?nil)
    :observability-types (Fully Oblivious)
    :relations  (Fully     (:and (?e ?e  ) (?nil ?nil))
                 Oblivious (:and (?e ?nil) (?nil ?nil)) )
    :designated (?e)
    :conditions (?nil :trivial-event)
  )

  (:action-type semi-private-sensing
    :events     (?e ?f)
    :observability-types (Fully Partially)
    :relations  (Fully     (:and (?e ?e) (?f ?f))
                 Partially (:forall (?x ?y - event) (?x ?y)) )
    :designated (?e ?f)
    :conditions (?e :trivial-postconditions
                 ?f :trivial-postconditions )
  )

  (:action-type quasi-private-announcement
    :events     (?e ?f ?nil)
    :observability-types (Fully Partially Oblivious)
    :relations  (Fully     (:forall (?x    - event) (?x ?x))
                 Partially (:and (?e ?e) (?e ?f) (?f ?e) (?f ?f))
                 Oblivious (:forall (?x    - event) (?x ?nil)) )
    :designated (?e)
    :conditions (?e   :trivial-postconditions
                 ?f   :trivial-postconditions
                 ?nil :trivial-event )
  )
)
    \end{lstlisting}

    A \texttt{public-ontic} action type represents an abstract frame with a single event \texttt{?e}, one observability type \texttt{Fully}, and with reflexive accessibility relations, \ie $Q_{\texttt{Fully}} = (\texttt{?e}, \texttt{?e})$ (see Section~\ref{sec:epistemic-action-types}).
    Furthermore, condition \texttt{:non-trivial-postconditions} requires that \texttt{?e} binds to an event that defines some effects, thus ensuring that the action is ontic.
    A \texttt{private} action type has an extra event variable, \texttt{?nil}, representing the null event, and an additional observability type, \texttt{Oblivious}.
    The \texttt{Fully} abstract relation is reflexive, while $Q_{\texttt{Oblivious}}$ links all events to the \texttt{?nil} event.
    Here, the only required condition is that \texttt{?nil} binds to a trivial event, ensuring that oblivious agents do not know that the action is taking place.
    A \texttt{semi-private-sensing} action type has two event variables, \texttt{?e} and \texttt{?f}, representing the two possible outcomes of the action, and two observability types, \texttt{Fully} and \texttt{Partially}.
    Fully observant agents know the outcome of the action ($Q_{\texttt{Fully}}$ is reflexive), while partially observant agents do not ($Q_{\texttt{Partially}}$ is the universal relation).
    As both \texttt{?e} and \texttt{?f} are designated, the action type induces local abstract actions for \texttt{Partially}.
    Furthermore, condition \texttt{:trivial-postconditions} is required for both event variables, ensuring that the action is purely epistemic.
    Finally, \texttt{quasi-private-announcement} action types combine the frames of of private and semi-private action types, and by letting \texttt{?e} be the only designated event variable.
    
    As mentioned above, with action type libraries we can define fragments of DEL-based epistemic planning by specifying a set of allowed action types and a set of requirements.
    For instance, the following listing shows a fragment that admits public, possibly ontic, atomic actions and semi-private, purely epistemic, global actions, with generic preconditions and goals, and quantified post- and observability conditions.

\begin{lstlisting}[language=EPDDL,label={lst:fragment},caption={\textsc{epddl} description of an DEL-based epistemic planning fragment.},captionpos={b}]
(define (action-type-library my-fragment)
  (:requirements
    :partial-observability :ontic-actions
    :general-preconditions :general-goals
    :quantified-postconditions :quantified-obs-conditions
  )

  (:action-type public-atomic
    :events     (?e)
    :observability-types (Fully)
    :relations  (Fully (?e ?e))
    :designated (?e)
  )

  (:action-type semi-private-epistemic
    :events     (?e ?f)
    :observability-types (Fully Partially)
    :relations  (Fully     (:forall (?x    - event) (?x ?x))
                 Partially (:forall (?x ?y - event) (?x ?y)) )
    :designated (?e)
    :conditions (?e :trivial-postconditions
                 ?f :trivial-postconditions )
  )
)
\end{lstlisting}

    The \texttt{:general-preconditions} and \texttt{:general-goals} requirements are an abbreviation for negative, disjunctive, quantified, and modal preconditions and goals, respectively, while \texttt{:quantified-postconditions} and \texttt{:quantified-obs-conditions} are an abbreviation for universal and existential postconditions and observability conditions, respectively (see Section~\ref{sec:syntax-requirements} for more details).

    As seen, by combining action types and requirements we can describe with a unified syntax a rich variety of DEL-based epistemic planning fragments.
    Furthermore, it can be shown that several epistemic planning formalisms not based on DEL can be translated into corresponding DEL-planning fragments, and thus that they are representable by appropriate action type libraries.
    This enables direct and meaningful comparisons of a wide range of solvers based on different epistemic planning formalisms.
    We believe this is one of the major strengths of \textsc{epddl}, as it allows us to create different sets of epistemic planning benchmarks, each based on a fragment of DEL-planning, by employing a unified language.

    \subsection{Domains}\label{sec:syntax-domains}
    Domains describe the universal aspects of a planning task.
    In \textsc{epddl}, these are types, predicates, constants, events and actions.

    \begin{ebnf}
        \ebnfDecl{domain}           & \ebnfProdArrow                      & \ebnfLeftPar ~\ebnfToken{define} \ebnfLeftPar ~\ebnfToken{domain} \ebnfDecl{name} \ebnfRightPar \\
                                    &                                     & \quad \ebnfDeclList{domain-item} \ebnfRightPar\\
        [0.4em]
        \ebnfDecl{domain-item}      & \ebnfProdArrow                      & \ebnfDecl{domain-libraries} \\
                                    & \ebnfProdOr                         & \ebnfDecl{require-decl} \\
                                    & \ebnfProdOr[:typing]                & \ebnfDecl{types-decl} \\
                                    & \ebnfProdOr                         & \ebnfDecl{predicates-decl} \\
                                    & \ebnfProdOr                         & \ebnfDecl{constants-decl} \\
                                    & \ebnfProdOr                         & \ebnfDecl{event-decl} \\
                                    & \ebnfProdOr                         & \ebnfDecl{action-decl} \\
        [0.4em]
        \ebnfDecl{domain-libraries} & \ebnfProdArrow                      & $\epsilon$ \\
                                    & \ebnfProdOr[:partial-observability] & \ebnfLeftPar ~\ebnfToken{:action-type-libraries} \ebnfDeclNonEmptyList{name} \ebnfRightPar
    \end{ebnf}

    Just like a problem specifies the domain it refers to, a domain may specify a list of action type libraries (requires \texttt{:partial-observability}).
    If a library is included in a domain specification, then all of its action types can be used in the definition of the domain's actions, as we show in Section~\ref{sec:syntax-actions}.
    Otherwise, if no library is specified, then it is assumed that all actions in the domain have a ``baseline'' type following the \textsc{epddl} called \texttt{basic} (note that this is a reserved keyword).
    The \texttt{basic} action type refers to the type of actions allowed in the baseline fragment of \textsc{epddl} described in Section~\ref{sec:syntax-ebnf}, namely fully observable, deterministic, purely epistemic, and atomic actions.
    The following is a possible definition of the \texttt{basic} action type (note that this is equivalent to the \texttt{public-atomic} action type shown in Section~\ref{sec:syntax-libraries}):
    \begin{lstlisting}[language=EPDDL,label={lst:basic-action-type},caption={The \texttt{basic} action type.},captionpos={b}]
(:action-type basic
  :events     (?e)
  :observability-types (Fully)
  :relations  (Fully (?e ?e))
  :designated (?e)
  :conditions (?e :trivial-postconditions)
)
    \end{lstlisting}

    Domain items may appear in any order in the domain description, and multiple declarations of requirements, types, predicates, and constants are allowed.
    In Section~\ref{sec:semantics-actions} we describe how multiple declarations are handled.
    Types, predicates, and constants are discussed in Section~\ref{sec:syntax-types-predicates-const}, events in Section~\ref{sec:syntax-events}, and actions in Section~\ref{sec:syntax-actions}.
    Requirements are discussed in Section~\ref{sec:syntax-requirements}.

    \subsubsection{Types, Predicates and Constants}\label{sec:syntax-types-predicates-const}
    The declarations of types, predicates, and constants in \textsc{epddl} are as follows:

    \begin{ebnf}
        \ebnfDecl{types-decl}              & \ebnfProdArrow      & \ebnfLeftPar ~\ebnfToken{:types} \ebnfDecl{typed-list$($name, primitive-type$)$} \ebnfRightPar \\
        [0.4em]
        \ebnfDecl{predicates-decl}         & \ebnfProdArrow      & \ebnfLeftPar ~\ebnfToken{:predicates} \ebnfDeclNonEmptyList{predicate-decl} \ebnfRightPar \\
        [0.4em]
        \ebnfDecl{predicate-decl}          & \ebnfProdArrow      & \ebnfLeftPar ~\ebnfDecl{standard-predicate-decl} \ebnfRightPar \\
                                           & \ebnfProdOr[:facts] & \ebnfLeftPar ~\ebnfDecl{fact-decl} \ebnfRightPar \\
        [0.4em]
        \ebnfDecl{standard-predicate-decl} & \ebnfProdArrow      & \ebnfDecl{name} \ebnfDecl{typed-list$($variable, type$)$} \\
        [0.4em]
        \ebnfDecl{fact-decl}               & \ebnfProdArrow      & \ebnfToken{:fact} \ebnfDecl{standard-predicate-decl} \\
        [0.4em]
        \ebnfDecl{constants-decl}          & \ebnfProdArrow      & \ebnfLeftPar ~\ebnfToken{:constants} \ebnfDecl{typed-list$($name, primitive-type$)$} \ebnfRightPar
    \end{ebnf}

    A \emph{type declaration} is a type name, possibly followed by a second type name, separated by a dash.
    For instance, an expression of the form $\mathtt{t_1}$ \texttt{-} $\mathtt{t_2}$ defines a new type $\mathtt{t_1}$ and with \emph{supertype} $\mathtt{t_2}$.
    We call $\mathtt{t_1}$ a \emph{subtype} of $\mathtt{t_2}$, and we say that $\mathtt{t_1}$ \emph{specialises} $\mathtt{t_2}$.
    It is also possible to specialise more than one type at a time.
    For instance, an expression of the form $\mathtt{t_1} \mathtt{t_2} \mathtt{t_3}$ \texttt{-} $\mathtt{t_4}$ defines the new types $\mathtt{t_1}$, $\mathtt{t_2}$ and $\mathtt{t_3}$, all of which specialise $\mathtt{t_4}$.
    Only \texttt{object}, \texttt{agent}, \texttt{agent-group} and and their subtypes can be specialised, while \texttt{entity}, \texttt{world}, \texttt{event} and \texttt{obs-type} can not.
    If no supertype is specified, then we assume that the supertype is \texttt{object}.
    Types \texttt{object}, \texttt{agent}, and \texttt{agent-group} are all subtypes of \texttt{entity}.
    
    A \emph{predicate declaration} is either a standard declaration, \ie a name followed by a typed list of variables, or a \emph{fact declaration}, being a standard declaration preceded by the \texttt{:fact} keyword (requires \texttt{:facts}).
    Predicate declarations marked as facts must be initialised in the facts-initialization statement (Section~\ref{sec:syntax-facts-init}).
    Standard predicate declarations, by contrast, are used in the initial state initialization (Section~\ref{sec:syntax-init-goal}).

    \emph{Constants} describe entities (\ie objects, agents, and agent groups) that are common to all problem instances of a domain, and they are defined akin to problem objects (Section~\ref{sec:syntax-objects-agents-groups}).
    Note that this is slightly different than \textsc{pddl}, where only constant objects can be defined.
    If a constant is declared with no type, then we assume that its type is \texttt{object}, so constant entities of type \texttt{agent} or \texttt{agent-group} must be explicitly declared with the appropriate type.
    The following listing shows the types and predicates of our Epistemic Blocks World domain.

    \begin{lstlisting}[language=EPDDL,label={lst:types-predicates},caption={Types and predicates for the Epistemic Blocks World domain.},captionpos={b}]
(define (domain epistemic-blocks-world)
  (:action-type-libraries my-library)

  (:requirements :typing)

  (:types block column)

  ; No constants declared
  ; (:constants)

  (:predicates
    (on ?b - block ?x - (either block column))
    (clear ?x - (either block column))
  )
  ...
)
    \end{lstlisting}

    \subsubsection{Events, Preconditions and Effects}\label{sec:syntax-events}
    Events in \textsc{epddl} are defined as follows:

    \begin{ebnf}
        \ebnfDecl{event-decl}          & \ebnfProdArrow                    & \ebnfLeftPar ~\ebnfToken{:event} \ebnfDecl{name} \\
                                       &                                   & \quad \ebnfDeclOptional{event-formal-params} \\
                                       &                                   & \quad \ebnfDeclOptional{precondition} \\
                                       &                                   & \quad \ebnfDeclOptional{effects} \ebnfRightPar \\
        [0.4em]
        \ebnfDecl{event-formal-params} & \ebnfProdArrow                    & \ebnfToken{:parameters} \ebnfLeftPar ~\ebnfDecl{typed-list$($variable, type$)$} \ebnfRightPar \\
        [0.4em]
        \ebnfDecl{precondition}        & \ebnfProdArrow                    & \ebnfToken{:precondition} \ebnfLeftPar ~\ebnfDecl{formula} \ebnfRightPar \\
        [0.4em]
        \ebnfDecl{effects}             & \ebnfProdArrow                    & \ebnfToken{:effects} \ebnfLeftPar ~\ebnfRightPar \\
                                       & \ebnfProdOr[:ontic-actions]       & \ebnfToken{:effects} \ebnfDecl{list$($cond-effect$)$} \\
        [0.4em]
        \ebnfDecl{cond-effect}         & \ebnfProdArrow                    & \ebnfDecl{literal} \\
                                       & \ebnfProdOr[:conditional-effects] & \ebnfLeftPar ~\ebnfToken{when} \ebnfDecl{formula} \ebnfDecl{list$($literal$)$} \ebnfRightPar \\
                                       & \ebnfProdOr[:conditional-effects] & \ebnfLeftPar ~\ebnfToken{iff} \ebnfDecl{formula} \ebnfDecl{list$($literal$)$} \ebnfRightPar \\
        [0.4em]
        \ebnfDecl{literal$(x)$}        & \ebnfProdArrow                    & \ebnfDecl{predicate$(x)$} \\
                                       & \ebnfProdOr                       & \ebnfLeftPar ~\ebnfToken{not} \ebnfDecl{predicate$(x)$} \ebnfRightPar
    \end{ebnf}

    In an event definition we specify the name of the event, a typed list of variables (the formal parameters of the event), a precondition, and the effects.
    The reader familiar with \textsc{pddl} will notice the similarity of events to classical actions definition.
    All components of events are optional: Empty parameters can be omitted and if the precondition (respectively, the effects) is not specified, then we assume that the event has a trivial precondition (respectively, effects).
    Events parameters are slightly different than actions parameters (defined in Section~\ref{sec:syntax-actions}).
    While the latter describes all possible combinations of parameters used at grounding time to generate the actions of a planning task, the former specifies the variables that must be passed to an event when it is referred to in an action definition.
    For this reason, events parameters are defined by a typed list of variables, and not by a list comprehension.
    A \emph{precondition} is a formula and the \emph{effects} are either an empty declaration (denoting trivial postconditions), or a list of \emph{conditional effects}.
    A conditional effect is either a \emph{literal}, a \emph{when-effect}, or an \emph{iff-effect}.
    A literal is either a predicate or its negation.
    When-effects are a formula (the condition of the effect) followed by a list of literals.
    The semantics is the same as in \textsc{pddl}: If the condition of the effect is not satisfied, then no change is produced by the event; otherwise, the event will make true all the predicates of positive literals occurring in the list, and will make false those appearing in negative literals.
    Iff-effects are similar, but their semantics is defined by a double implication, and not by a single implication as the previous type of effects.
    Iff-effects are included in \textsc{epddl} to better align the language with the semantics of DEL, where effects in the product update are expressed using double implications (see Definition~\ref{def:product-update}).
    This syntactic construct allows for a more direct and faithful representation of DEL-style event model updates within the planning language.
    We now provide an example.

    \begin{lstlisting}[language=EPDDL,label={lst:event},caption={Example of events in \textsc{epddl}: the \texttt{e-move} event represents the move of a block, the \texttt{e-tell} event represents the communication that an agent knows/believes the position of a block, and the \texttt{nil} event represents the null event.},captionpos={b}]
(define (domain epistemic-blocks-world)
  ...

  (:requirements :typing :modal-preconditions :ontic-actions)
  
  (:event e-move
    :parameters (?b - block ?x ?y - (either block column))
    :precondition (and (on ?b ?x) (clear ?b) (clear ?y))
    :effects (:and
           (on ?b ?y)
      (not (on ?b ?x))
           (clear ?x)
      (not (clear ?y)) )
  )
  
  (:event e-tell
    :parameters (?ag - agent ?b - block ?x - (either block column))
    :precondition ([?ag] (on ?b ?x))
    ; :effects ()
  )
  
  (:event nil
    ; :parameters ()
    ; :precondition (true)
    ; :effects ()
  )
  ...
)
    \end{lstlisting}

    The \texttt{e-move} event takes three parameters: \texttt{?b}, of type \texttt{block}, and \texttt{?x} and \texttt{?y}, either of type \texttt{block} or \texttt{column}.
    The precondition requires that \texttt{?b} is on top of \texttt{?x}, and that both \texttt{?b} and \texttt{?y} are clear.
    This event describes the move of a block:
    The effects specify that after the event is applied to a world, in the updated world \texttt{?b} will be on top of \texttt{?y} (and no longer on \texttt{?x}), and \texttt{?x} becomes clear, while \texttt{?y} is no longer clear.

    The \texttt{e-tell} event takes three parameters: \texttt{?ag}, of type \texttt{agent}, \texttt{?b}, of type \texttt{block}, and \texttt{?x}, either of type \texttt{block} or \texttt{column}.
    The precondition requires that agent \texttt{?ag} knows/believes that \texttt{?b} is on top of \texttt{?x}, and the effects are trivial.
    This event describes the announcement of the fact that the agent knows/believes where the block is.
    
    Finally, the \texttt{nil} event is a trivial event, as both its precondition and its effects are trivial.
    This event describes the perspective of oblivious agents, who believe that no action is taking place.

    \subsubsection{Actions and Observability Conditions}\label{sec:syntax-actions}
    So far, we have seen events, which describe preconditions and effects, and action types, which describe in observability types, (designated) events and abstract accessibility relations.
    Events and action types are combined together in action declarations to describe abstract epistemic actions, as the next EBNF shows:

    \begin{ebnf}
        \ebnfDecl{action-decl}           & \ebnfProdArrow                      & \ebnfLeftPar ~\ebnfToken{:action} \ebnfDecl{name} \\
                                         &                                     & \quad \ebnfDecl{action-formal-params} \\
                                         &                                     & \quad \ebnfDecl{action-type-sign} \\
                                         &                                     & \quad \ebnfDecl{obs-conditions} \ebnfRightPar \\
        [0.4em]
        \ebnfDecl{action-formal-params}  & \ebnfProdArrow                      & \ebnfToken{:parameters} \ebnfDecl{formal-params} \\
        [0.4em]
        \ebnfDecl{action-type-sign}      & \ebnfProdArrow                      & \ebnfToken{:action-type} \\
                                         &                                     & \quad \ebnfLeftPar ~\ebnfDecl{act-type-name} \ebnfDeclNonEmptyList{event-sign} \ebnfRightPar \\
        [0.4em]
        \ebnfDecl{act-type-name}         & \ebnfProdArrow                      & \ebnfToken{basic} \\
                                         & \ebnfProdOr[:partial-observability] & \ebnfDecl{name} \\
        [0.4em]
        \ebnfDecl{event-sign}            & \ebnfProdArrow                      & \ebnfLeftPar ~\ebnfDecl{name} \ebnfDeclList{term} \ebnfRightPar \\
        [0.4em]
        \ebnfDecl{obs-conditions}        & \ebnfProdArrow                      & $\epsilon$ \\
                                         & \ebnfProdOr[:partial-observability] & \ebnfToken{:observability-conditions} \ebnfDecl{list$($obs-cond$)$} \\
        [0.4em]
        \ebnfDecl{obs-cond}              & \ebnfProdArrow                      & \ebnfLeftPar ~\ebnfDecl{term} \ebnfDecl{name} \ebnfRightPar \\
                                         & \ebnfProdOr                         & \ebnfLeftPar ~\ebnfDecl{term} \ebnfDecl{if-then-else-obs-cond} \ebnfRightPar \\
                                         & \ebnfProdOr                         & \ebnfLeftPar ~\ebnfToken{default} \ebnfDecl{name} \ebnfRightPar \\
        [0.4em]
        \ebnfDecl{if-then-else-obs-cond} & \ebnfProdArrow                      & \ebnfLeftPar ~\ebnfDecl{if-obs-cond} \\
                                         &                                     & \quad \ebnfDeclList{else-if-obs-cond} \\
                                         &                                     & \quad \ebnfDeclOptional{else-obs-cond} \ebnfRightPar \\
        [0.4em]
        \ebnfDecl{if-obs-cond}           & \ebnfProdArrow                      & \ebnfToken{if} \ebnfDecl{formula} \ebnfDecl{name} \\
        [0.4em]
        \ebnfDecl{else-if-obs-cond}      & \ebnfProdArrow                      & \ebnfToken{else-if} \ebnfDecl{formula} \ebnfDecl{name} \\
        [0.4em]
        \ebnfDecl{else-obs-cond}         & \ebnfProdArrow                      & \ebnfToken{else} \ebnfDecl{name}
    \end{ebnf}

    An \emph{action declaration} is an action name, followed by its \emph{formal parameters}, \emph{action type signature}, and \emph{observability conditions}.
    The formal parameters of an action are specified either by a typed list of variables or by a list comprehension (Section~\ref{sec:syntax-common-lists}).
    Note that, when list comprehensions are used, we specify conditions that explicitly control the set of ground actions generated at grounding time.
    For instance, suppose that a domain uses the static predicates \texttt{(edge ?x ?y - node)} to represent the graph of a map where agents can move.
    Using list comprehensions, we can tell the grounder to only generate meaningful grounded instances of the \texttt{move} action by defining its parameters with the list comprehension \texttt{(?x ?y - node | (edge ?x ?y))}.
    In this way, the grounder will only generate a ground action $\mathtt{move_{m,n}}$ if \texttt{m} and \texttt{n} are objects of type \texttt{node} such that \texttt{(edge m n)} holds.
    While it is not strictly necessary to use list comprehensions, since the existence of an edge could alternatively be enforced in the action's preconditions, doing so can significantly reduce the number of ground actions generated, thus improving the efficiency of epistemic planners.
    In our example, this reduction can be particularly beneficial in domains with sparse graphs or a large number of nodes.
    Furthermore, the use of list comprehensions offers a conceptual advantage, as we clearly separate the conditions that determine the existence of actions from those that govern their applicability.

    An action type signature is an \emph{action type name} followed by a non-empty sequence of \emph{event signatures}.
    An action type name is either \texttt{basic}, or a general name (requires \texttt{:partial-observability}).
    The \texttt{basic} action type denotes the only action type allowed in the baseline configuration of \textsc{epddl} (Section~\ref{sec:syntax-ebnf}), namely purely epistemic, fully observable, deterministic, atomic actions (see Listing~\ref{lst:basic-action-type}).
    An event signature is an event name followed by a possibly empty sequence of terms (the actual parameters of the event).
    For instance, the expression \texttt{:action-type (private (e-move ?b ?x ?y) (nil))} declares an action of type \texttt{private}, and it binds the events \texttt{(e-move ?b ?x ?y)} and \texttt{(nil)} to the event parameters \texttt{?e} and \texttt{?nil}, respectively, declared in the \texttt{private} action type (see Listing~\ref{lst:action-types}).
    As we show in more detail in Section~\ref{sec:semantics-actions}, the abstract frame of the action is thus built according to the specification of the \texttt{private} action type, instantiated on the events \texttt{(e-move ?b ?x ?y)} and \texttt{(nil)}.

    An observability condition is either a \emph{static}, an \emph{if-then-else}, or a \emph{default} observability condition.
    If specified, observability conditions require \texttt{:partial-observability}.
    Note that the observability types available to an action are those defined in its action type.
    For instance, actions of type \texttt{private} can refer to the observability types \texttt{Fully} and \texttt{Oblivious}, as specified in Listing~\ref{lst:action-types}.
    A static observability condition is a term followed by an observability type name.
    For instance, \texttt{(Fully Alice)} means that Alice is a fully observant agent.
    An if-then-else observability condition allows the observability type to depend on the truth of one or more formulas, and their structure is similar to standard if-then-else constructs in programming languages.
    For example, \texttt{(?a (if (looking ?a) Fully else Oblivious))} specifies that agent \texttt{?a} is fully observant if it is looking, and otherwise oblivious.
    Default observability conditions specify the observability type to use when no other condition applies.

    The following constraints, verified at type-checking time, apply to observability conditions.
    \begin{inparaenum}
        \item\label{itm:obs-cond-rule-1} There must be an observability condition declaration for each agent, unless a default one is provided (in that case, the observability type of the agent will be the default one);
        \item\label{itm:obs-cond-rule-2} There can not be multiple observability conditions declarations for the same agent;
        \item\label{itm:obs-cond-rule-3} All if-then-else observability conditions must contain an ``else'' statement, unless a default observability condition is given; and
        \item\label{itm:obs-cond-rule-4} At most one default observability condition can be defined.
    \end{inparaenum}
    These conditions will later help us to ensure (see Section~\ref{sec:semantics-actions}) that the observability conditions satisfy the requirements imposed in Definition~\ref{def:abstract-event-model}.
    We now show some examples from the Epistemic Blocks World domain.

    \begin{lstlisting}[language=EPDDL,label={lst:actions},caption={Example of actions in \textsc{epddl}.},captionpos={b}]
(define (domain epistemic-blocks-world)
  (:action-type-libraries my-library)

  (:requirements
    :typing :list-comprehensions :partial-observability
    :ontic-actions :negative-preconditions
  )
  ...
  (:event e-move ...)
  (:event e-tell ...)
  (:event nil)
  
  (:event e-peek-pos
    :parameters (?b - block ?x - (either block column))
    :precondition (and (clear ?b) (on ?b ?x))
  )

  (:event e-peek-neg
    :parameters (?b - block ?x - (either block column))
    :precondition (and (clear ?b) (not (on ?b ?x)))
  )
  
  (:action move
    :parameters (?i    - agent
                 ?b    - block
                 ?x ?y - (either block column)
                   | (and (/= ?b ?x) (/= ?b ?y) (/= ?x ?y)) )
    :action-type (private (e-move ?b ?x ?y) (nil))
    :observability-conditions (:and
      (?i Fully)
      (:forall (?j - agent | (/= ?i ?j)) (?j Oblivious)) )
      ; Equivalently, one could instead use (default Oblivious)
  )

  (:action tell
    :parameters (?i - agent
                 ?b  - block
                 ?x  - (either block column) | (/= ?b ?x) )
    :action-type (basic (e-tell ?i ?b ?x))
    ; The basic action type can always be used
    :observability-conditions (default Fully)
  )

  (:action peek
    :parameters (?i - agent
                 ?b  - block
                 ?x  - (either block column) | (/= ?b ?x) )
    :action-type (semi-private-sensing (e-peek-pos ?b ?x)
                                       (e-peek-neg ?b ?x))
    :observability-conditions (:and
      (?i Fully)
      (default Partially) )
  )
)
    \end{lstlisting}

    Action \texttt{move} takes four parameters: an agent \texttt{?ag}, a block \texttt{?b}, and two blocks/columns \texttt{?x} and \texttt{?y} such that \texttt{?b}, \texttt{?x}, and \texttt{?y} are all different.
    The action is declared with type \texttt{private} (included from the action type library of Listing~\ref{lst:action-types}), which describes the perspective of two types of agents: fully observant agents, who know that the action is being executed and its effects, and oblivious agents, who believe no action is happening.
    In this action, the observability conditions specify that only \texttt{?ag} is fully observant, while all other agents are oblivious.
    We pass to \texttt{private} the events \texttt{e-move} and \texttt{nil}, which bind to the event variables \texttt{?e} and \texttt{?nil} of the action type, respectively, so we obtain the following frame for action \texttt{move}:
    \begin{itemize}
        \item $E = \{ \texttt{?e}, \texttt{?nil} \} = \{ \texttt{e-move}, \texttt{nil} \}$;
        \item $Q_{\texttt{Fully}} = \{ (\texttt{?e}, \texttt{?e}), (\texttt{?nil}, \texttt{?nil}) \} = \{ (\texttt{e-move}, \texttt{e-move}), (\texttt{nil}, \texttt{nil}) \}$;
        \item $Q_{\texttt{Oblivious}} = \{ (\texttt{?e}, \texttt{?nil}), (\texttt{?nil}, \texttt{?nil}) \} = \{ (\texttt{e-move}, \texttt{nil}), (\texttt{nil}, \texttt{nil}) \}$; and
        \item $E_d = \{ \texttt{?e} \} = \{ \texttt{e-move} \}$.
    \end{itemize}
    Events \texttt{e-move} and \texttt{nil} are described in Section~\ref{sec:syntax-events}, and they represent the actual move of the block and the null action, respectively.

    Action \texttt{tell} takes three parameters, an agent \texttt{?ag}, a block \texttt{?b} and a block/column \texttt{?x} such that \texttt{?b} $\neq$ \texttt{?x}.
    It is a \texttt{basic} action that is instantiated on the \texttt{e-tell} event defined in Section~\ref{sec:syntax-events}, meaning that \texttt{tell} is a public announcement.
    Specifically, since the precondition of \texttt{e-tell} is \texttt{([?ag] (on ?b ?x))}, the action is a public announcement of the fact that agent \texttt{?ag} knows/believes that block \texttt{?b} is on top of \texttt{?x}.

    Action \texttt{peek} takes three parameters, an agent \texttt{?ag}, a block \texttt{?b} and a block/column \texttt{?x} such that \texttt{?b} $\neq$ \texttt{?x}.
    It is a \texttt{semi-private-sensing} action that is instantiated on events \texttt{e-peek-pos} and \texttt{e-peek-neg}.
    Both events have no effect and their preconditions are \texttt{(and (clear ?b) (on ?b ?x))} and \texttt{(and (clear ?b) (not (on ?b ?x)))}, respectively.
    Hence, we have that the action is purely epistemic, globally deterministic, and its preconditions cover the logical space, so \texttt{peek} is a (local) semi-private sensing action (Section~\ref{sec:epistemic-action-types}).
    The action represents \texttt{?ag} learning whether block \texttt{?b} is on top of \texttt{?x} or not.
    Semi-private actions describe the perspective of two types of agents: fully observant agents, who know about both the action execution and its effects, and partially observant agents, who only know about the execution.
    In this action, the observability conditions specify that only \texttt{?ag} is fully observant, while all other agents are partially observant.
    Moreover, since the action is local, it describes the perspective of partially observant agents.


  



    \subsection{Requirements}\label{sec:syntax-requirements}
    Requirements specify the set of features of a problem, domain or action type library that extend the baseline fragment of \textsc{epddl} (Section~\ref{sec:syntax-ebnf}), and that are required of epistemic planners to solve a problem.
    Some requirements imply others, while others are abbreviations for common sets of requirements.
    If no requirement is provided, it is assumed the baseline requirement \texttt{:pal}.
    The following is a complete list of \textsc{epddl} requirements:

    \begin{ebnf}[lrlrl]
        \ebnfDecl{require-decl} & \ebnfProdArrow &  \ebnfLeftPar \ebnfToken{ :requirements} \ebnfDeclNonEmptyList{require-key} \ebnfRightPar \\
        [0.4em]
        \ebnfDecl{require-key}  & \ebnfProdArrow & \ebnfToken{:agent-groups}               & \ebnfProdOr & \ebnfToken{:list-comprehensions}       \\
                                & \ebnfProdOr    & \ebnfToken{:common-knowledge}           & \ebnfProdOr & \ebnfToken{:modal-formulas}            \\
                                & \ebnfProdOr    & \ebnfToken{:conditional-effects}        & \ebnfProdOr & \ebnfToken{:modal-goals}               \\
                                & \ebnfProdOr    & \ebnfToken{:del}                        & \ebnfProdOr & \ebnfToken{:modal-obs-conditions}      \\
                                & \ebnfProdOr    & \ebnfToken{:disjunctive-formulas}       & \ebnfProdOr & \ebnfToken{:modal-postconditions}      \\
                                & \ebnfProdOr    & \ebnfToken{:disjunctive-goals}          & \ebnfProdOr & \ebnfToken{:modal-preconditions}       \\
                                & \ebnfProdOr    & \ebnfToken{:disjunctive-list-formulas}  & \ebnfProdOr & \ebnfToken{:multi-pointed-models}      \\
                                & \ebnfProdOr    & \ebnfToken{:disjunctive-obs-conditions} & \ebnfProdOr & \ebnfToken{:negative-formulas}         \\
                                & \ebnfProdOr    & \ebnfToken{:disjunctive-postconditions} & \ebnfProdOr & \ebnfToken{:negative-goals}            \\
                                & \ebnfProdOr    & \ebnfToken{:disjunctive-preconditions}  & \ebnfProdOr & \ebnfToken{:negative-list-formulas}    \\
                                & \ebnfProdOr    & \ebnfToken{:equality}                   & \ebnfProdOr & \ebnfToken{:negative-obs-conditions}   \\
                                & \ebnfProdOr    & \ebnfToken{:events-conditions}          & \ebnfProdOr & \ebnfToken{:negative-postconditions}   \\
                                & \ebnfProdOr    & \ebnfToken{:existential-formulas}       & \ebnfProdOr & \ebnfToken{:negative-preconditions}    \\
                                & \ebnfProdOr    & \ebnfToken{:existential-goals}          & \ebnfProdOr & \ebnfToken{:ontic-actions}             \\
                                & \ebnfProdOr    & \ebnfToken{:existential-list-formulas}  & \ebnfProdOr & \ebnfToken{:pal}                       \\
                                & \ebnfProdOr    & \ebnfToken{:existential-obs-conditions} & \ebnfProdOr & \ebnfToken{:partial-observability}     \\
                                & \ebnfProdOr    & \ebnfToken{:existential-postconditions} & \ebnfProdOr & \ebnfToken{:quantified-formulas}       \\
                                & \ebnfProdOr    & \ebnfToken{:existential-preconditions}  & \ebnfProdOr & \ebnfToken{:quantified-goals}          \\
                                & \ebnfProdOr    & \ebnfToken{:finitary-S5-theories}       & \ebnfProdOr & \ebnfToken{:quantified-list-formulas}  \\
                                & \ebnfProdOr    & \ebnfToken{:facts}                      & \ebnfProdOr & \ebnfToken{:quantified-obs-conditions} \\
                                & \ebnfProdOr    & \ebnfToken{:general-formulas}           & \ebnfProdOr & \ebnfToken{:quantified-postconditions} \\
                                & \ebnfProdOr    & \ebnfToken{:general-frames}             & \ebnfProdOr & \ebnfToken{:quantified-preconditions}  \\
                                & \ebnfProdOr    & \ebnfToken{:general-goals}              & \ebnfProdOr & \ebnfToken{:static-common-knowledge}   \\
                                & \ebnfProdOr    & \ebnfToken{:general-list-formulas}      & \ebnfProdOr & \ebnfToken{:typing}                    \\
                                & \ebnfProdOr    & \ebnfToken{:general-obs-conditions}     & \ebnfProdOr & \ebnfToken{:universal-formulas}        \\
                                & \ebnfProdOr    & \ebnfToken{:general-postconditions}     & \ebnfProdOr & \ebnfToken{:universal-goals}           \\
                                & \ebnfProdOr    & \ebnfToken{:general-preconditions}      & \ebnfProdOr & \ebnfToken{:universal-list-formulas}   \\
                                & \ebnfProdOr    & \ebnfToken{:group-modalities}           & \ebnfProdOr & \ebnfToken{:universal-obs-conditions}  \\
                                & \ebnfProdOr    & \ebnfToken{:KD45-frames}                & \ebnfProdOr & \ebnfToken{:universal-postconditions}  \\
                                & \ebnfProdOr    & \ebnfToken{:knowing-whether}            & \ebnfProdOr & \ebnfToken{:universal-preconditions}   \\
                                & \ebnfProdOr    & \ebnfToken{:lists}                      &             &
    \end{ebnf}

    The \texttt{:negative-preconditions} requirement implies \texttt{:disjunctive-preconditions} (and similarly for postconditions, observability conditions, goals, and static formulas).
    The \texttt{:conditional-effects} requirement is implied by each postconditions requirement.
    The \texttt{:general-preconditions} requirement is an abbreviation for all precondition requirements, and the \texttt{:disjunctive-formulas} is an abbreviation for all disjunctive formulas requirements (the same applies to similar requirements).
    The \texttt{:del} requirement abbreviates all of the following: \texttt{:typing}, \texttt{:equality}, \texttt{:partial-observability}, \texttt{:ontic-actions}, \texttt{multi-pointed-models}, \texttt{:general-frames}, \texttt{:general-formulas}.
    Finally, \texttt{:finitary-S5-theories} implies \texttt{:common-knowledge} and \texttt{:knowing-whether};
    \texttt{:common-knowledge} implies \texttt{:group-modalities}; \texttt{:static-common-knowledge} implies \texttt{:group-modalities} and \texttt{:facts}; and
    both \texttt{:agent-groups} and \texttt{:group-modalities} imply \texttt{:lists}.

    \section{Semantics of \textsc{epddl}}\label{sec:epddl-semantics}
    In this section, we define the semantics of \textsc{epddl} specifications based on abstract epistemic planning tasks of ADEL (Definition~\ref{def:abstract-planning-task}).
    More precisely, letting $\mathtt{T} = (\mathtt{Prob}, \mathtt{Dom}, \mathtt{Libs})$ denote an \textsc{epddl} specification, where $\mathtt{Prob}$ is a problem, $\mathtt{Dom}$ is a domain and $\mathtt{Libs} = \{ \mathtt{Lib_1}, \dots, \mathtt{Lib_k} \}$ is a possibly empty, finite set of action type libraries, we define the abstract epistemic planning task $T = (s_0, \actionSet, \phi_g)$ determined from $\mathtt{T}$.
    In the remainder of the section, $\mathtt{T} = (\mathtt{Prob}, \mathtt{Dom}, \mathtt{Libs})$ denotes a fixed \textsc{epddl} specification.

    The section is structured as follows.
    Section~\ref{sec:semantics-formulas} defines a translation of \textsc{epddl} agents, agent groups, predicates and formulas to their respective logical counterparts.
    Section~\ref{sec:semantics-init-goal} determines the initial state $s_0$ and the goal formula $\phi_g$ of the planning task.
    Finally, Section~\ref{sec:semantics-actions} specifies the set $\actionSet$ of abstract actions of $T$.

    \subsection{Agents, Predicates and Formulas}\label{sec:semantics-formulas}
    Let $\objectSet$, $\agentSet$, and $\groupSet$ be the sets of object names, agent names, and agent-group names, respectively, declared in $\mathtt{Prob}$.
    The set $\agentSet$ will be the agent set of our logical language.
    Let $\mymathit{agents} : \groupSet \rightarrow 2^\agentSet$ be a function that associates to an agent-group name $\mathtt{G}$ the set $\mymathit{agents}(\mathtt{G})$ of agents in its corresponding declaration.
    Let $\primitiveTypeSet$ be the sets of type names declared in $\mathtt{Dom}$, together with the reserved types \texttt{object}, \texttt{agent}, \texttt{agent-group} and \texttt{entity}, and let $\typeSet = \bigcup_{k > 0} \{ (\mathtt{t_1}, \dots, \mathtt{t_k}) \in \primitiveTypeSet^k \}$ be the set of tuple of primitive types, where a tuple $(\mathtt{t_1}, \dots, \mathtt{t_k}) \in \typeSet$ ($k > 0$) denotes the composite type $\texttt{(}\mathtt{either\ t_1 \dots t_k}\texttt{)}$.
    Note that the (degenerate) composite type \texttt{(either t)} (where $\mathtt{t} \in \primitiveTypeSet$) is trivially equivalent to \texttt{t}, and hence we often consider a singleton tuple $(\mathtt{t}) \in \typeSet$ to represent the primitive type \texttt{t}.
    We let $\mathtt{p(x_1 : t_1, \dots, x_k : t_k)}$ denote a $\mathtt{k}$-ary \emph{predicate signature} for predicate $\mathtt{p}$ on variables $\mathtt{x_1}, \dots, \mathtt{x_k}$ with types $\mathtt{t_1}, \dots, \mathtt{t_k} \in \typeSet$, respectively, and we let $\predSet$ and $\constSet$ be the sets of predicate signatures and constant names, respectively, declared in $\mathtt{Dom}$.
    If multiple declarations of objects, agents, agent groups, types, predicates, or constants are present, we define the corresponding set of elements as the union of the elements in each declaration of the same kind.
    For instance, if $\mathtt{Prob}$ contains multiple object declarations, \eg \texttt{(:objects b1 b2 - block)} and \texttt{(:objects b3 b4 - block)}, then $\objectSet = \{ \mathtt{b1}, \mathtt{b2}, \mathtt{b3}, \mathtt{b4} \}$.

    We now determine the set $\atomSet$ of ground atoms.
    Let $\entitySet = \constSet \cup \objectSet \cup \agentSet \cup \groupSet$, let $\entityType : \entitySet \rightarrow \primitiveTypeSet$ be a function that maps to each entity the type specified in its declaration, or its implicit type as described in Sections~\ref{sec:syntax-objects-agents-groups} and~\ref{sec:syntax-types-predicates-const}.
    We define the following compatibility relation on types.
    \begin{definition}[Compatible Types]
        A primitive type $\mathtt{t}$ is \emph{compatible} with a primitive type $\mathtt{t'}$, denoted $\compTypes{t}{t'}$, iff $\mathtt{t}$ is a subtype of $\mathtt{t'}$ or $\mathtt{t} = \mathtt{t'}$.
        A composite type $\mathtt{(either\ t_1 \dots t_m)}$ is \emph{compatible} with a composite type $\mathtt{(either\ u_1 \dots u_n)}$ iff for all $\mathtt{t_i}$ there exists a $\mathtt{u_j}$ such that $\compTypes{t_i}{u_j}$.
    \end{definition}
    Note that, by construction, we always have $\compTypes{object}{entity}$, $\compTypes{agent}{entity}$ and $\compTypes{agent\textnormal{\texttt{-}}group}{entity}$.
    Furthermore, in the Epistemic Blocks World domain (Listing~\ref{lst:types-predicates}) we have $\compTypes{\compTypes{block}{object}}{entity}$, but not $\compTypes{block}{agent}$.

    For a primitive type $\mathtt{t}$, we let $\typedEntitySet{t} = \{ \mathtt{e} \in \entitySet \mid \compTypes{e}{t} \}$ be the set of entities whose type is compatible with \texttt{t}.
    We extend the notation to composite types and we let $\typedEntitySet{(t_1, ..., t_k)} = \bigcup_{1 \leq \mathtt{i} \leq \mathtt{k}} \typedEntitySet{t_i}$.
    We then let $\atomSet = \{ \mathtt{p_{e_1, \dots, e_k}} \mid \mathtt{p(x_1 : t_1, \dots, x_k : t_k)} \in \predSet$ and $\mathtt{e_i} \in \typedEntitySet{t_i}$ for all $1 \leq \mathtt{i} \leq \mathtt{k} \}$ be the set of ground atoms obtained by instantiating the predicate signatures in $\predSet$ with the entities of the appropriate types.
    Furthermore, we let $\factSet \subseteq \atomSet$ be the set of ground predicates obtained from facts declarations.

    \begin{example}[Entities, Types and Predicates]
        Consider the \textnormal{\texttt{ebw1}} problem (Listing~\ref{lst:init-explicit}) from the \textnormal{\texttt{epistemic-blocks-}} \textnormal{\texttt{world}} domain (Listing~\ref{lst:types-predicates}).
        The sets of objects, agents and agent-group names of \textnormal{\texttt{ebw1}} are $\objectSet = \{ \mathtt{b1}, \mathtt{b2}, \mathtt{b3}, \mathtt{b4}, \mathtt{c1},$ $\mathtt{c2}, \mathtt{c3} \}$, $\agentSet = \{ \mathtt{A}, \mathtt{L}, \mathtt{R} \}$ and $\groupSet = \varnothing$, respectively.
        The set of primitive types is $\primitiveTypeSet = \{ \mathtt{block}, \mathtt{column} \}$ and the set of constants is $\constSet = \varnothing$.
        As specified in Section~\ref{sec:syntax-types-predicates-const}, since their declarations do not specify a supertype, both $\mathtt{block}$ and $\mathtt{column}$ are implicitly assigned supertype $\mathtt{object}$.
        We thus have: $\compTypes{block\mathnormal{,}column}{object}$ and $\compTypes{object\mathnormal{,}agent}{entity}$.
        Moreover, as $\mathtt{A}$, $\mathtt{L}$ and $\mathtt{R}$ are declared without an explicit type, they are implicitly assigned with type $\mathtt{agent}$, as described in Section~\ref{sec:syntax-objects-agents-groups}.
        From this, we obtain the following sets of entities: $\typedEntitySet{block} = \{ \mathtt{b1}, \mathtt{b2}, \mathtt{b3}, \mathtt{b4} \}$, $\typedEntitySet{column} = \{ \mathtt{c1}, \mathtt{c2}, \mathtt{c3} \}$, $\typedEntitySet{object} = \typedEntitySet{block} \cup \typedEntitySet{column}$, $\typedEntitySet{agent} = \{ \mathtt{A}, \mathtt{L}, \mathtt{R} \}$, $\typedEntitySet{agent\textnormal{\texttt{-}}group} = \varnothing$, and $\typedEntitySet{entity} = \typedEntitySet{object} \cup \typedEntitySet{agent} \cup \typedEntitySet{agent\textnormal{\texttt{-}}group}$.
        The set of predicate signatures is $\predSet = \{ \mathtt{on(b : block, x : (block, column))}, \mathtt{clear(x : (block, column))} \}$, and the set of ground predicates is then $\atomSet = \{ \mathtt{on_{b, x}}, \mathtt{clear_x} \mid \mathtt{b} \in \typedEntitySet{block}$ and $\mathtt{x} \in \typedEntitySet{(block, column)} \}$.
        Finally, as no facts are declared, we have $\factSet = \varnothing$.
    \end{example}

    We now have defined the agent set $\agentSet$ and the atom set $\atomSet$, which give us the language $\Lang[C]$ of the formulas of our planning task.
    Note that, for generality, we consider the entire set of formulas representable in \textsc{epddl}.
    In practice, the specific language of formulas available is determined by the requirements of the given specification $\mathtt{T}$.

    We now introduce the following notion of substitution.
    \begin{definition}[Substitution]\label{def:substitution}
        Let $\mathtt{stmt}$ be a syntactically well-formed \textnormal{\textsc{epddl}} statement containing occurrences of variables $\mathtt{x_1, \dots, x_k}$, and let $(\mathtt{e_1}, \dots, \mathtt{e_k})$ be a tuple of entities such that, for all $1 \leq \mathtt{i} \leq \mathtt{k}$, the type of $\mathtt{e_i}$ is compatible with the type of $\mathtt{x_i}$.
        The \emph{substitution} of $\mathtt{x_1, \dots, x_k}$ with $(\mathtt{e_1}, \dots, \mathtt{e_k})$ in $\mathtt{stmt}$ is the statement $\mathtt{stmt_{[x_1/e_1, \dots, x_k/e_k]}}$ obtained from $\mathtt{stmt}$ by substituting each occurrence of a variable $\mathtt{x_i}$ with the corresponding entity $\mathtt{e_i}$.
    \end{definition}

    As we anticipated in Section~\ref{sec:syntax-common}, the conditions of list comprehensions are evaluated on the set of true facts declared in the \ebnfDecl{facts-init} statement of the problem.
    We denote this set with $\positiveFacts$ (if the statement is not declared, then we let $\positiveFacts = \varnothing$).
    Let $\mathtt{params} = \mathtt{\{ x_1 : t_1, \dots, x_k : t_k \mid phi \}}$ denote a list comprehension with variables $\mathtt{x_1, \dots, x_k}$ of types $\mathtt{t_1, \dots, t_k} \in \typeSet$, respectively, and with condition $\mathtt{phi}$ being a static, propositional \textsc{epddl} formula (if no condition is specified, we assume that $\mathtt{phi}$ is \texttt{(true)}).
    By mutual recursion we now define induced power sets $2^\mathtt{params}$ of list comprehensions, the expansion function $\mymathit{listExp}$ of \textsc{epddl} lists, and a translation function $\tau$ from \textnormal{\textsc{epddl}} formulas to formulas of $\Lang[C]$.
    The next definition will play a central role in the remainder of this section.

        \begin{definition}[Induced Power Set, List Expansion, and Formulas Translation]\label{def:power-set-expansion-translation}
            The \emph{induced power set} of a list comprehension $\mathtt{params} = \mathtt{\{ x_1 : t_1, \dots, x_k : t_k \mid phi \}}$ is the set
            \[
                2^\mathtt{params} \;=\; \{ (\mathtt{e_1, \dots, e_k}) \in \typedEntitySet{t_1} \times \dots \times \typedEntitySet{t_k} \mid \positiveFacts \models \tau(\mathtt{phi_{[x_1/e_1, \dots, x_k/e_k]}}) \},
            \]
            \ie all properly typed entity-tuples that make the translated condition true in $\positiveFacts$ (here standard propositional semantics is used).

            Let ~$\mathtt{list}$ be an \textnormal{\textsc{epddl}} list expression of ``$x$'' elements.
            We define its \emph{list expansion} $\mymathit{listExp}(\mathtt{list})$ by induction on the structure of the list:
            \begin{itemize}
                \item If ~$\mathtt{list}$ is a singleton list $x$, then $\mymathit{listExp}(\mathtt{list}) = \{ x \}$;
                \item If ~$\mathtt{list}$ is a concatenation \textnormal{\texttt{(:and $\mathtt{l_1}$ ... $\mathtt{l_k}$)}} of lists of $x$, then
                \begin{equation*}
                    \mymathit{listExp}(\mathtt{list}) = \bigcup_{\mathtt{i} = 1}^{\mathtt{k}} \mymathit{listExp}(\mathtt{l_i});
                \end{equation*}
                \item If ~$\mathtt{list}$ is a universally quantified list \textnormal{\texttt{(:forall params l)}}, with $\mathtt{params} = \mathtt{\{ x_1 : t_1, \dots, x_k : t_k \mid phi \}}$, then
                \begin{equation*}
                    \mymathit{listExp}(\mathtt{list}) = \displaystyle\bigcup_{(\mathtt{e_1,\dots,e_k}) \in 2^\mathtt{params}} \mymathit{listExp}(\mathtt{l_{[x_1/e_1,\dots,x_k/e_k]}}).
                \end{equation*}
            \end{itemize}

            The \emph{translation function} $\tau$ is defined by induction on the structure of \textnormal{\textsc{epddl}} formulas:
            \begin{itemize}
                \item Atomic formulas:
                    \begin{equation*}
                        \begin{array}{lll}
                            \tau(\textnormal{\texttt{(true)}})                                                & = & \top,                                                                                              \\[0.25em]
                            \tau(\textnormal{\texttt{(false)}})                                               & = & \bot,                                                                                              \\[0.25em]
                            \tau(\textnormal{\texttt{(}} \mathtt{p\ e_1\ ...\ e_k} \textnormal{\texttt{)}}) & = & \mathtt{p_{e_1,\dots,e_k}},                                                                        \\[0.25em]
                            \tau(\textnormal{\texttt{(= }}  \mathtt{t_1\ t_2} \textnormal{\texttt{)}})        & = & \top \textnormal{ \emph{if} } \mathtt{t_1} =    \mathtt{t_2}; \bot \textnormal{ \emph{otherwise}}, \\[0.25em]
                            \tau(\textnormal{\texttt{(/= }} \mathtt{t_1\ t_2} \textnormal{\texttt{)}})        & = & \top \textnormal{ \emph{if} } \mathtt{t_1} \neq \mathtt{t_2}; \bot \textnormal{ \emph{otherwise}}.
                        \end{array}
                    \end{equation*}
                \item Propositional connectives:
                    \begin{equation*}
                        \begin{array}{lll}
                            \tau(\textnormal{\texttt{(}} \mathtt{not\ phi}                 \textnormal{\texttt{)}}) & = & \neg\tau(\mathtt{phi}),                                   \\[0.25em]
                            \tau(\textnormal{\texttt{(}} \mathtt{and\ phi_1\ ...\ phi_k} \textnormal{\texttt{)}}) & = & \bigwedge_{\mathtt{i}=1}^\mathtt{k} \tau(\mathtt{phi_i}), \\[0.25em]
                            \tau(\textnormal{\texttt{(}} \mathtt{or\ phi_1\ ...\ phi_k}  \textnormal{\texttt{)}}) & = & \bigvee_{\mathtt{i}=1}^\mathtt{k} \tau(\mathtt{phi_i}),   \\[0.25em]
                            \tau(\textnormal{\texttt{(}} \mathtt{imply\ phi_1\ phi_2}      \textnormal{\texttt{)}}) & = & \tau(\mathtt{phi_1}) \rightarrow \tau(\mathtt{phi_2}).
                        \end{array}
                    \end{equation*}
                \item Quantifiers over a list comprehension $\mathtt{params} = \mathtt{\{ x_1 : t_1, \dots, x_k : t_k \mid psi \}}$:
                    \begin{equation*}
                        \begin{array}{l}
                            \tau(\mathtt{\textnormal{\texttt{(}} forall\ params\ phi \textnormal{\texttt{)}} }) = \displaystyle\bigwedge_{(\mathtt{e_1,\dots,e_k}) \in 2^\mathtt{params}} \tau(\mathtt{phi_{[x_1/e_1,\dots,x_k/e_k]}}),\\[0.75em]
                            \tau(\mathtt{\textnormal{\texttt{(}} exists\ params\ phi \textnormal{\texttt{)}} }) = \displaystyle\bigvee_{(\mathtt{e_1,\dots,e_k}) \in 2^\mathtt{params}} \tau(\mathtt{phi_{[x_1/e_1,\dots,x_k/e_k]}}).
                        \end{array}
                    \end{equation*}
                \item Modalities.
                Let $\mathtt{m}$ be an \textnormal{\textsc{epddl}} modality index.
                If $\mathtt{m}$ is a single entity, then we let the corresponding logical modality index be $\mymathit{m} = \mathtt{m}$.
                If $\mathtt{m}$ is an \textnormal{\textsc{epddl}} list of entities, then we let $\mymathit{m} = \mymathit{listExp}(\mathtt{m})$.
                Finally, if ~$\mathtt{m} = \mathtt{All}$, we let $\mymathit{m} = \agentSet$.
                For each modality type we then have the following translations:
                    \begin{equation*}
                        \begin{array}{lll}
                            \tau(\textnormal{\texttt{([}} \mathtt{     m \textnormal{\texttt{]}}\ phi} \textnormal{\texttt{)}} ) & = & \B   {\mymathit{m}} \tau(\mathtt{phi}), \\[0.25em]
                            \tau(\textnormal{\texttt{(<}} \mathtt{     m \textnormal{\texttt{>}}\ phi} \textnormal{\texttt{)}} ) & = & \D   {\mymathit{m}} \tau(\mathtt{phi}), \\[0.25em]
                            \tau(\textnormal{\texttt{([}} \mathtt{Kw.\ m \textnormal{\texttt{]}}\ phi} \textnormal{\texttt{)}} ) & = & \Kw  {\mymathit{m}} \tau(\mathtt{phi}), \\[0.25em]
                            \tau(\textnormal{\texttt{(<}} \mathtt{Kw.\ m \textnormal{\texttt{>}}\ phi} \textnormal{\texttt{)}} ) & = & \Kwd {\mymathit{m}} \tau(\mathtt{phi}), \\[0.25em]
                            \tau(\textnormal{\texttt{([}} \mathtt{C.\  m \textnormal{\texttt{]}}\ phi} \textnormal{\texttt{)}} ) & = & \CK  {\mymathit{m}} \tau(\mathtt{phi}), \\[0.25em]
                            \tau(\textnormal{\texttt{(<}} \mathtt{C.\  m \textnormal{\texttt{>}}\ phi} \textnormal{\texttt{)}} ) & = & \CKd {\mymathit{m}} \tau(\mathtt{phi}).
                        \end{array}
                    \end{equation*}
            \end{itemize}
        \end{definition}

    The induced power set of a list comprehension is the set of tuples of the appropriate type that satisfy the provided condition.
    The expansion of a list is obtained by recursively evaluating quantified list definitions and flattening concatenations, producing the concrete set of elements defined by the list.
    \textsc{epddl} predicates are translated into their corresponding ground predicate.
    Propositional connectives are replaced by their corresponding logical counterpart.
    Propositional universally (resp., existentially) quantified formulas are translated into finite conjunctions (resp., disjunctions) over the translations of the instantiated formulas for each tuple in the corresponding induced power set.
    Finally, modalities are translated by first evaluating the \textsc{epddl} modality index \texttt{m} to the corresponding logical modality index $\mymathit{m}$ (a single agent, the agent group resulting by the expansion of an entity list, or $\agentSet$ for \texttt{All}) and then applying the corresponding modal operator to the translation of the inner formula.

    \begin{example}
        Let $\mathtt{phi}$ be the following \textnormal{\textsc{epddl}} formula:
\begin{lstlisting}[language=EPDDL]
([C. (:forall (?i - agent) (?i))]
  (forall (?i - agent | (/= ?i A))
    (exists (?b - block | (and (/= ?b b1) (/= ?b b2)))
      ([Kw. ?i] (on ?b b1)))))
\end{lstlisting}
        The modality index of the common knowledge formula, call it $\mathtt{m}$, is defined by a universally quantified list.
        By Definition~\ref{def:power-set-expansion-translation}, the corresponding logical index is $\mymathit{m} = \mymathit{listExp}(\mathtt{m}) = \bigcup_{\mathtt{i} \in \agentSet} \mathtt{i} = \agentSet$.
        Let now $\mathtt{params} = \{ \mathtt{i} : \mathtt{agent} \mid \mathtt{i} \neq \mathtt{A} \}$ and $\mathtt{params'} = \{ \mathtt{b} : \mathtt{block} \mid \mathtt{b} \neq \mathtt{b_1} \land \mathtt{b} \neq \mathtt{b_2} \}$ denote the two list comprehensions appearing in the second and third line, respectively.
        By Definition~\ref{def:power-set-expansion-translation}, their induced power sets are $2^\mathtt{params} = \{ \mathtt{i} \in \typedEntitySet{agent} \mid \mathtt{i} \neq \mathtt{A} \} = \{ \mathtt{L}, \mathtt{R} \}$ and $2^\mathtt{params'} = \{ \mathtt{b} \in \typedEntitySet{block} \mid \mathtt{b} \neq \mathtt{b_1} \land \mathtt{b} \neq \mathtt{b_2} \} = \{ \mathtt{b_3}, \mathtt{b_4} \}$, respectively.
        Finally, the modality index of the knowing whether formula is the variable \textnormal{\texttt{?i}}, which will be instantiated into each agent $i \in 2^\mathtt{params}$, yielding the modal operators $\Kw{i}$ in the expansion.
        Then, by recursively applying $\tau$ we obtain the following translated logical formula:
        \begin{equation*}
            \begin{array}{lll}
                \tau(\mathtt{phi}) & = &
                \CK{\{\mathtt{A}, \mathtt{L}, \mathtt{R}\}}
                \left(
                    \bigwedge_{i \in \{ \mathtt{L}, \mathtt{R} \}}
                        \bigvee_{b \in \{ \mathtt{b_3}, \mathtt{b_4} \}} \Kw{i} \mathtt{on_{\mathnormal{b}, b_1}}
                \right)
                \\
                [0.75em]
                & = &
                \CK{\{\mathtt{A}, \mathtt{L}, \mathtt{R}\}}
                \left(
                    \left(
                        \Kw{\mathtt{L}} \mathtt{on_{b_3, b_1}} \lor \Kw{\mathtt{L}} \mathtt{on_{b_4, b_1}}
                    \right)
                    \land
                    \left(
                        \Kw{\mathtt{R}} \mathtt{on_{b_3, b_1}} \lor \Kw{\mathtt{R}} \mathtt{on_{b_4, b_1}}
                    \right)
                \right)
            \end{array}
        \end{equation*}
    \end{example}

    \subsection{Initial State and Goal}\label{sec:semantics-init-goal}
    We now determine the initial epistemic state $s_0$ and the goal formula $\phi_g$ of our abstract planning task $T = (s_0, \actionSet, \phi_g)$.
    If there is one goal declaration \texttt{(:goal $\mathtt{phi_g}$)} in the problem specification, then $\phi_g = \tau(\mathtt{phi_g})$.
    If multiple goal declarations are present, then $\phi_g$ is the conjunction of all translated goal formulas.
    For instance, the goal declared in Listing~\ref{lst:init-explicit} is \texttt{([C. All] (on b2 b1))}, so we have $\phi_g = \CK{\agentSet} \mathtt{on_{b_2, b_1}}$.


    As stated in Section~\ref{sec:syntax-init-goal}, an initial state may be either declared explicitly or via a finitary S5-theory.
    In the former case, let $\mathtt{s_0}$ denote the explicit initial state declaration, and let $\mathtt{s_0}.\mymathit{worlds}$, $\mathtt{s_0}.\mymathit{rel}$, $\mathtt{s_0}.\mymathit{labels}$, and $\mathtt{s_0}.\mymathit{design}$ denotes the lists of world names, accessibility relations, world labels, and designated-world names specified in $\mathtt{s_0}$, respectively.
    The initial state $s_0 = ((W, R, L), W_d)$ of $T$ is obtained as follows.
    The world set is $W = \mathtt{s_0}.\mymathit{worlds}$ (the set of world names defined in $\mathtt{s_0}$) and, similarly, the designated world set is $W_d = \mathtt{s_0}.\mymathit{design}$.
    For each agent relation declaration in $\mathtt{s_0}.\mymathit{rel}$ of the form $\mathtt{i}\ \mathtt{R_i}$, where $\mathtt{R_i}$ is an \textsc{epddl} list of pairs of world names, we let the accessibility relation $R_\mathtt{i} = \mymathit{listExp}(\mathtt{R_i})$ be the set of world-name pairs in the list expansion of $\mathtt{R_i}$.
    If no declaration is provided for $\mathtt{i}$, we then let $R_\mathtt{i} = \varnothing$.
    Finally, if \texttt{w label} is a world-label declaration in $\mathtt{s_0}.\mymathit{labels}$, where \texttt{w} is a world name and \texttt{label} is an \textsc{epddl} list of predicates,
    we let $L(\mathtt{w}) = \mymathit{listExp}(\mathtt{label}) \cup \positiveFacts$ (facts hold in all possible worlds).
    If no declaration is provided for $\mathtt{w}$, we then let $L(\mathtt{w}) = \positiveFacts$.
    We now show an example.
    
    \begin{example}[Explicitly Defined Initial State]\label{ex:semantic-init-explicit}
        Consider the explicit initial state declaration of Listing~\ref{lst:init-explicit}.
        The corresponding initial state is $s_0 = ((W, R, L), W_d)$, where:
        \begin{itemize}
            \item $W = \{ \mathtt{w_1}, \mathtt{w_2}, \mathtt{w_3} \}$;
            \item $R_\mathtt{A} = W \times W$;
            \item $R_\mathtt{L} = \{ (\mathtt{w_1}, \mathtt{w_1}), (\mathtt{w_2}, \mathtt{w_2}), (\mathtt{w_2}, \mathtt{w_3}), (\mathtt{w_3}, \mathtt{w_2}), (\mathtt{w_3}, \mathtt{w_3}) \}$;
            \item $R_\mathtt{R} = \{ (\mathtt{w_1}, \mathtt{w_1}), (\mathtt{w_1}, \mathtt{w_2}), (\mathtt{w_2}, \mathtt{w_1}), (\mathtt{w_2}, \mathtt{w_2}), (\mathtt{w_3}, \mathtt{w_3}) \}$;
            \item $L(\mathtt{w_1}) = \{ \mathtt{on_{b_1, c_1}}, \mathtt{on_{b_2, b_1}}, \mathtt{on_{b_3, c_2}}, \mathtt{on_{b_4, c_3}}, \mathtt{clear_{b_2}}, \mathtt{clear_{b_3}}, \mathtt{clear_{b_4}} \}$;
            \item $L(\mathtt{w_2}) = \{ \mathtt{on_{b_1, c_2}}, \mathtt{on_{b_2, c_1}}, \mathtt{on_{b_3, b_1}}, \mathtt{on_{b_4, c_3}}, \mathtt{clear_{b_2}}, \mathtt{clear_{b_3}}, \mathtt{clear_{b_4}} \}$;
            \item $L(\mathtt{w_3}) = \{ \mathtt{on_{b_1, c_3}}, \mathtt{on_{b_2, c_1}}, \mathtt{on_{b_3, c_2}}, \mathtt{on_{b_4, b_1}}, \mathtt{clear_{b_2}}, \mathtt{clear_{b_3}}, \mathtt{clear_{b_4}} \}$; and
            \item $W_d = \{ \mathtt{w_1}, \mathtt{w_2} \}$.
        \end{itemize}
        Note that this is exactly the epistemic state of Example~\ref{ex:multi-agent-bw-local}.
    \end{example}

    If instead the initial state is defined via a finitary S5-theory $\Phi$, then $s_0$ is computed from $\Phi$, as we now show.
    Let $M^\mymathit{univ} = (W^\mymathit{univ}, R^\mymathit{univ}, L^\mymathit{univ})$ be an epistemic model such that $W^\mymathit{univ} = \{ P' \in 2^\atomSet \mid s_\mymathit{facts} \subseteq P' $ and $P' \cap (\factSet \setminus s_\mymathit{facts}) = \varnothing \}$,
    $R^\mymathit{univ}_\agentStyle{i} = W \times W$ for all $\agentStyle{i} \in \agentSet$ and $L^\mymathit{univ}(w) = w$ for all $w \in W$.
    That is, $M^\mymathit{univ}$ contains one world per possible label that contains all the true facts and no false facts, the accessibility relations are the universal relation on such worlds, and the label of a world is the world itself.
    In this model, every possible valuation of the propositional atoms compatible with $s_\mymathit{facts}$ is represented as a distinct world, and the accessibility relations for all agents are universal, meaning each agent considers every world possible.
    Consequently, it is commonly known that all agents are maximally uncertain about the actual state of affairs; that is, for any agent $\agentStyle{i} \in \agentSet$ and any distinct worlds $w,v \in W^\mymathit{univ}$, $\agentStyle{i}$ cannot distinguish $w$ from $v$.
    However, by requiring that all labels are compatible with $s_\mymathit{facts}$, we force facts to be commonly known in $M^\mymathit{univ}$.
    In fact, it is not hard to check that for all $w \in W^\mymathit{univ}$ and all $p \in \factSet$ we have that $(M^\mymathit{univ}, w) \models \CK{\agentSet} p$ if $p \in s_\mymathit{facts}$, and $(M^\mymathit{univ}, w) \models \CK{\agentSet} \neg p$ otherwise.
    The \emph{induced state} $s_\Phi = (M, W_d)$ of $\Phi$, with $M = (W, R, L)$, is going to be computed by removing worlds and/or edges from $M^\mymathit{univ}$ in such a way that the resulting state satisfies all formulas in $\Phi$.
    Specifically, $s_\Phi$ is computed as follows:
    \begin{itemize}
        \item Let $\psi_\mymathit{worlds} = \bigwedge_{\left( \CK{\agentSet} \B{i}\phi \right) \in \Phi} \phi$ be the conjunction of all type~\ref{itm:s5-theory-2} formulas in $\Phi$ (including those with abbreviated form $\CK{\agentSet}\phi$).
        Then, we let $W$ be the set of worlds $w \in W^\mymathit{univ}$ such that $(M^\mymathit{univ}, w) \models \psi_\mymathit{worlds}$;
        \item
        For all agents $\agentStyle{i} \in \agentSet$, we let $R_\agentStyle{i} = W \times W \setminus \{ (w, v) \in W \times W \mid$ there exists $\left( \CK{\agentSet} \Kw{i}\phi \right) \in \Phi$ such that $(M^\mymathit{univ}, w) \models \phi$ iff $(M^\mymathit{univ}, v) \not\models \phi \}$;
        \item For all worlds $w$, we let $L(w) = w$;
        \item Finally, let $\psi_\mymathit{design} = \bigwedge_{\phi \in \Phi} \phi$ be the conjunction of all type~\ref{itm:s5-theory-1} formulas in $\Phi$.
        Then, we let $w \in W_d$ iff $(M^\mymathit{univ}, w) \models \psi_\mymathit{design}$.
    \end{itemize}
    Formula $\psi_\mymathit{worlds}$ represents the facts that are commonly known in $s_\Phi$ by all agents, and thus it identifies the worlds of the induced state.
    Similarly, $\psi_\mymathit{design}$ identifies the designated worlds.
    The $\agentStyle{i}$-edges of $s_\Phi$ are obtained by removing from the universal relation $W \times W$ all the world pairs $(w, v)$ such that $w$ and $v$ do not agree on some formula $\phi$ appearing in a type~\ref{itm:s5-theory-3} formula for agent $i$.
    This is because $(M, w) \models \Kw{i}\phi$ (\ie in world $w$, $\agentStyle{i}$ knows that $\phi$, or $\agentStyle{i}$ knows that $\neg\phi$) iff the $R_\agentStyle{i}$ contains only world pairs $(w, v)$ where $w$ and $v$ agree on $\phi$, and $s_\Phi \models \CK{\agentSet} \Kw{i}\phi$ iff $\Kw{i}\phi$ holds in all worlds of $s_\Phi$.
    Note that, as we begin from model $M^\mymathit{univ}$, we do not need to consider type~\ref{itm:s5-theory-4} formulas in the construction of $s_\Phi$.
    In fact, type~\ref{itm:s5-theory-4} formulas describe the uncertainty of some agent about a propositional formula, and $M^\mymathit{univ}$ captures a situation where all agents are maximally uncertain about the actual state of affairs.
    
    This construction is almost identical to the one by \cite{conf/jelia/SonPBG14}, the only difference being the treatment of type~\ref{itm:s5-theory-1} formulas.
    Specifically, the construction by \cite{conf/jelia/SonPBG14} yields a set $S = \{ (M, w_d) \mid w_d \in W_d \}$ of single-pointed models, one for each designated world, rather than a single multi-pointed model as presented here.
    This is because their framework only considers single-pointed epistemic models.
    Note that the states in $S$ share the same epistemic model, and only differ in the actual world.
    In contrast, our approach directly constructs a unique multi-pointed model, accommodating specifications with multiple designated worlds.
    It is not hard to check that the states $s_\Phi$ and $\bigsqcup_{s \in S} s$ (the disjoint union of the states in $S$) satisfy the same formulas of $\Lang[C]$, and thus that the two constructions are equivalent.

    \begin{example}[State Induced by a Finitary S5-Theory]\label{ex:semantic-init-finitary}
        Let $\Phi$ be the finitary S5-theory obtained from Listing~\ref{lst:init-finitary-s5-theory}.
        We first define the following abbreviations:
        \begin{itemize}
            \item $\phi_{\mymathit{clear}} = \mathtt{clear_{b_2}} \land \mathtt{clear_{b_3}} \land \mathtt{clear_{b_4}}$;
            \item $\phi_{blocks}^1 = \mathtt{on_{b_1, c_1}} \land \mathtt{on_{b_2, b_1}} \land \mathtt{on_{b_3, c_2}} \land \mathtt{on_{b_4, c_3}}$;
            \item $\phi_{blocks}^2 = \mathtt{on_{b_1, c_2}} \land \mathtt{on_{b_2, c_1}} \land \mathtt{on_{b_3, b_1}} \land \mathtt{on_{b_4, c_3}}$,;
            \item $\phi_{blocks}^3 = \mathtt{on_{b_1, c_3}}, \mathtt{on_{b_2, c_1}}, \mathtt{on_{b_3, c_2}}, \mathtt{on_{b_4, b_1}}$.
        \end{itemize}
        
        Following the procedure described above, the induced state $s_\Phi = ((W, R, L), W_d)$ is calculated as follows:
        \begin{itemize}
            \item The only type~\ref{itm:s5-theory-2} formula in $\Phi$ is $\CK{\{\mathtt{A}, \mathtt{L}, \mathtt{R}\}} \psi_\mymathit{worlds}$, where $\psi_\mymathit{worlds} = \phi_{\mymathit{clear}} \land (\phi_{blocks}^1 \lor \phi_{blocks}^2 \lor \phi_{blocks}^3)$.
            It is not hard to check that $\psi_\mymathit{worlds}$ is only satisfied by the following three worlds of model $M^\mymathit{univ}$:
            \begin{enumerate}
                \item $w_1 = \{ \mathtt{on_{b_1, c_1}}, \mathtt{on_{b_2, b_1}}, \mathtt{on_{b_3, c_2}}, \mathtt{on_{b_4, c_3}}, \mathtt{clear_{b_2}}, \mathtt{clear_{b_3}}, \mathtt{clear_{b_4}} \}$;
                \item $w_2 =
                \{ \mathtt{on_{b_1, c_2}}, \mathtt{on_{b_2, c_1}}, \mathtt{on_{b_3, b_1}}, \mathtt{on_{b_4, c_3}}, \mathtt{clear_{b_2}}, \mathtt{clear_{b_3}}, \mathtt{clear_{b_4}} \}$;
                \item $w_3 =
                \{ \mathtt{on_{b_1, c_3}}, \mathtt{on_{b_2, c_1}}, \mathtt{on_{b_3, c_2}}, \mathtt{on_{b_4, b_1}}, \mathtt{clear_{b_2}}, \mathtt{clear_{b_3}}, \mathtt{clear_{b_4}} \}$.
            \end{enumerate}
            We thus have $W = \{ w_1, w_2, w_3 \}$ and $L(w) = w$ for all $w \in W$.
            \item Consider agent $\mathtt{A}$. As there are no type~\ref{itm:s5-theory-3} formulas for agent $\mathtt{A}$ in $\Phi$, we immediately get $R_\mathtt{A} = W \times W$.
            \item Consider now agent $\mathtt{L}$.
            The only type~\ref{itm:s5-theory-3} formula for agent $\mathtt{L}$ in $\Phi$ is $\CK{\{\mathtt{A}, \mathtt{L}, \mathtt{R}\}} \Kw{\mathtt{L}} \mathtt{on_{b_2, b_1}}$.
            Following our construction, we need to remove from $R_\mathtt{L}^\mymathit{univ}$ all those pairs of worlds that do not agree on $\mathtt{on_{b_2, b_1}}$.
            We have that world $w_1$ does not agree with $w_2$ and $w_3$ on $\mathtt{on_{b_2, b_1}}$, while $w_2$ and $w_3$ do, meaning that there can be no $\mathtt{L}$-edges between $w_1$ and $w_2$, nor between $w_1$ and $w_3$.
            Thus, $R_\mathtt{L} = W \times W \setminus \{ (w_1, w_2), (w_2, w_1), (w_1, w_3), (w_3, w_1) \}$.
            \item Consider now agent $\mathtt{R}$.
            The only type~\ref{itm:s5-theory-3} formula for agent $\mathtt{R}$ in $\Phi$ is $\CK{\{\mathtt{A}, \mathtt{L}, \mathtt{R}\}} \Kw{\mathtt{R}} \mathtt{on_{b_4, b_1}}$.
            Since $\mathtt{on_{b_4, b_1}}$ is true in $w_3$, but false in $w_1$ and $w_2$, reasoning as above we get that there can be no $\mathtt{R}$-edges between $w_3$ and $w_1$, nor between $w_3$ and $w_2$.
            Thus, $R_\mathtt{R} = W \times W \setminus \{ (w_1, w_3), (w_3, w_1), (w_2, w_3), (w_3, w_2) \}$.
            \item The only type~\ref{itm:s5-theory-1} formula in $\Phi$ is $\psi_\mymathit{design} = \phi_{\mymathit{clear}} \land (\phi_{blocks}^1 \lor \phi_{blocks}^2)$, which is satisfied by worlds $w_1$ and $w_2$, so $W_d = \{ w_1, w_2 \}$, and we're done.
        \end{itemize}
        Note that $s_\Phi$ is identical to both the epistemic state of Example~\ref{ex:multi-agent-bw-local} and to the one of Example~\ref{ex:semantic-init-explicit}.
    \end{example}

    \subsection{Actions Set}\label{sec:semantics-actions}
    The only missing component of our planning task $T$ is its action set $\actionSet$.
    We first introduce some notation.
    Let $\mymathit{Events}$ and $\mymathit{Actions}$ be the set of event and action declarations, respectively, of $\mathtt{Dom}$, and $\mymathit{ActTypes}$ be the set of action type declarations of all libraries in $\mathtt{Libs}$, plus the \texttt{basic} action type declaration (Section~\ref{sec:syntax-ebnf}), which we recall here for completeness:
    \begin{lstlisting}[language=EPDDL]
(:action-type basic
  :events     (?e)
  :observability-types (Fully)
  :relations  (Fully (?e ?e))
  :designated (?e)
  :conditions (?e :trivial-postconditions)
)
    \end{lstlisting}
    For an action declaration $\mathtt{a} \in \mymathit{Actions}$, we let $\mathtt{a}.\mymathit{params}$ denote the list comprehension in its parameters, and we let $\mathtt{a}.\mymathit{act\_type}$ and $\mathtt{a}.\mymathit{events}$ be the action type declaration and the list of event declarations specified in $\mathtt{a}$.
    For a tuple $(\mathtt{e_1, \dots, e_k}) \in 2^{\mathtt{a}.\mymathit{params}}$, we let $\mathtt{a}_\mathtt{e_1, \dots, e_k}$ denote the abstract epistemic action obtained from the instantiation of $\mathtt{a}$ on parameters $\mathtt{e_1, \dots, e_k}$, defined below.
    We then let $\actionSet = \{ \mathtt{a}_\mathtt{e_1, \dots, e_k} \mid \mathtt{a} \in \mymathit{Actions}$ and $(\mathtt{e_1, \dots, e_k}) \in 2^{\mathtt{a}.\mymathit{params}} \}$ be the set of abstract actions obtained by instantiating all action declarations of $\mathtt{Dom}$ on all relevant combinations of parameters.
    
    In what follows, we consider a fixed action declaration $\mathtt{a} \in \mymathit{Actions}$, with parameters $\mathtt{a}.\mymathit{params} = \mathtt{\{ x_1 : t_1, \dots,}$ $\mathtt{x_k : t_k \mid phi \}}$, and we let $\obsTypeSet$ be the set of observability types defined in $\mathtt{a}.\mymathit{act\_type}$.
    Letting $(\mathtt{e_1, \dots, e_k}) \in 2^{\mathtt{a}.\mymathit{params}}$ be a tuple of entities, in the rest of the section we define the abstract epistemic action $\mathtt{\mathtt{a}_{e_1, \dots, e_k}} = ((E, Q, \mymathit{pre}, \mymathit{post}, \mymathit{obs}), E_d)$ on $\obsTypeSet$ induced by $\mathtt{a}$.

    \paragraph{Events, Relations and Designated Events.}
        For an event declaration $\mathtt{e} \in \mymathit{Events}$, we let $\mathtt{e}.\mymathit{params}$, $\mathtt{e}.\mymathit{pre}$, and $\mathtt{e}.\mymathit{effects}$ denote the parameters of the event, its precondition, and effects, respectively.
        Letting $\mathtt{aType} = \mathtt{a}.\mymathit{act\_type}$, we denote with $\mathtt{aType}.\mymathit{events}$, $\mathtt{aType}.\mymathit{rel}$, and $\mathtt{aType}.\mymathit{design}$ the lists of event variables, abstract accessibility relations, and designated event variables specified in $\mathtt{aType}$, respectively.
        We assume that the list of events specified in the action type signature of $\mathtt{a}$ is compatible with the event variables in $\mathtt{aType}.\mymathit{events}$, namely that the correct number of events is provided in $\mathtt{a}$, and all event conditions of events are satisfied.
        For all event variables $\mathtt{ev_i} \in \mathtt{aType}.\mymathit{events}$, we let $\mathtt{e_i} \in \mathtt{a}.\mymathit{events}$ denote the concrete event declaration that is bound to $\mathtt{ev_i}$.
        We then let $E = \{ \mathtt{e_i} \mid \mathtt{ev_i} \in \mathtt{aType}.\mymathit{events} \}$ be the set of the event declarations that are bound to the event variables of $\mathtt{aType}$.
        Similarly, the designated events are $E_d = \{ \mathtt{e_i} \mid \mathtt{ev_i} \in \mathtt{aType}.\mymathit{design} \}$.
        Finally, for an observability type $\mathtt{t} \in \obsTypeSet$, let $\mathtt{Q_t}$ denote the accessibility relation for $\mathtt{t}$ declared in $\mathtt{aType}.\mymathit{rel}$.
        Recall that $\mathtt{Q_t}$ is syntactically defined as an \textsc{epddl} list of pairs of event variables.
        Then, the abstract accessibility relation for $\mathtt{t}$ is $Q_\mathtt{t} = \{ (\mathtt{e_i}, \mathtt{e_j}) \mid (\mathtt{ev_i}, \mathtt{ev_j}) \in \mymathit{listExp}(\mathtt{Q_t}) \}$, \ie the set of pairs of event declarations obtained from the list expansion of $\mathtt{Q_t}$.
        If no declaration is provided for $\mathtt{t}$, we then let $Q_\mathtt{t} = \varnothing$.

    \paragraph{Preconditions and Postconditions.}
        Let $\mathtt{e} \in E$ be an event declaration.
        We let $\mymathit{pre}(\mathtt{e}) = \tau(\mathtt{e}.\mymathit{pre})$ be the translation of the precondition formula declared in $\mathtt{e}$ (Definition~\ref{def:power-set-expansion-translation}).
        If $\mathtt{e}$ does not specify a precondition, we let $\mymathit{pre}(\mathtt{e}) = \top$.
        
        Given a ground predicate $\mathtt{p} \in \atomSet$, we now define the postcondition $\mymathit{post}(\mathtt{e}, \mathtt{p})$.
        Let $\mymathit{Cond}^+_\mathtt{p}, \mymathit{Cond}^-_\mathtt{p} \subseteq 2^{\Lang[C]}$ be two sets denoting the positive and negative conditions, respectively, assigned to $\mathtt{p}$ in the effects of the event.
        Let $\mathtt{ce} \in \mymathit{listExp}(\mathtt{e}.\mymathit{effects})$ be a conditional effect statement (recall that the effects $\mathtt{e}.\mymathit{effects}$ of the event are syntactically described by an \textsc{epddl} list of conditional effects).
        We have three cases:
        \begin{enumerate}
            \item \emph{Literals}: If $\mathtt{ce}$ is \texttt{(p)}, we let $\top \in \mymathit{Cond}^+_\mathtt{p}$ and $\bot \in \mymathit{Cond}^-_\mathtt{p}$.
            Otherwise, if $\mathtt{ce}$ is \texttt{(not (p))}, we let $\bot \in \mymathit{Cond}^+_\mathtt{p}$ and $\top \in \mymathit{Cond}^-_\mathtt{p}$;
            \item \emph{When-effects}: Let $\mathtt{ce}$ be a when-effect with condition \texttt{phi} and list of literals \texttt{ls}.
            If \texttt{(p)} $\in \mymathit{listExp}(\mathtt{ls})$, we let $\tau(\mathtt{phi}) \in \mymathit{Cond}^+_\mathtt{p}$.
            Otherwise, if \texttt{(not (p))} $\in \mymathit{listExp}(\mathtt{ls})$ we let $\tau(\mathtt{phi}) \in \mymathit{Cond}^-_\mathtt{p}$; and
            \item  \emph{Iff-effects}: Let $\mathtt{ce}$ is an iff-effect with condition \texttt{phi} and list of literals \texttt{ls}.
            If \texttt{(p)} $\in \mymathit{listExp}(\mathtt{ls})$, we let $\tau(\mathtt{phi}) \in \mymathit{Cond}^+_\mathtt{p}$ and $\neg \tau(\mathtt{phi}) \in \mymathit{Cond}^-_\mathtt{p}$.
            Otherwise, if \texttt{(not (p))} $\in \mymathit{listExp}(\mathtt{ls})$ we let $\neg \tau(\mathtt{phi}) \in \mymathit{Cond}^+_\mathtt{p}$ and $\tau(\mathtt{phi}) \in \mymathit{Cond}^-_\mathtt{p}$.
        \end{enumerate}
        We assume positive and negative conditions to be consistent, \ie that for any $\psi^+ \in \mymathit{Cond}^+_\mathtt{p}$ and $\psi^- \in \mymathit{Cond}^-_\mathtt{p}$ we have $\models \psi^+ \land \psi^- \rightarrow \bot$.
        Then, we construct the postcondition of $\mathtt{p}$ in $\mathtt{e}$ as follows:
        \begin{equation*}
            \mymathit{post}(\mathtt{e}, \mathtt{p}) =
                \bigvee_{\phi \in \mymathit{Cond}^+_\mathtt{p}} \phi \lor \left( \mathtt{p} \land \neg \bigvee_{\phi \in \mymathit{Cond}^-_\mathtt{p}} \phi \right)
        \end{equation*}

        Intuitively, the postcondition formula for $\mathtt{p}$ after event $\mathtt{e}$ combines the positive and negative conditions to determine the new truth value of $\mathtt{p}$.
        The first disjunct of the formula ensures that $\mathtt{p}$ holds if any of the conditions that positively affect it are satisfied.
        The second disjunct preserves the previous truth value of $\mathtt{p}$ unless a negative condition is met, in which case it forces $\mathtt{p}$ to false.
        This construction faithfully captures the intended semantics of \textsc{epddl} effects: literals directly override the atom's truth value, when-effects conditionally update only when their condition holds, and iff-effects set the truth value precisely based on their condition.
        Note that, under such semantics, we implicitly capture a common and natural assumption made in several planning formalisms: if $\mathtt{e}$ does not specify any effect for $\mathtt{p}$, then we immediately get $\mymathit{Cond}^+_\mathtt{p} = \mymathit{Cond}^-_\mathtt{p} = \varnothing$, and thus $\mymathit{post}(\mathtt{e}, \mathtt{p}) = \bot \lor (\mathtt{p} \land \top) = \mathtt{p}$.

    \paragraph{Observability Conditions.}
        Let $\mathtt{a}.\mymathit{obs\_cond}$ denote the \textsc{epddl} list of observability conditions declared in $\mathtt{a}$, and let $\mymathit{obsConds} = \mymathit{listExp}(\mathtt{a}.\mymathit{obs\_cond})$ denote its expansion.
        We assume that $\mymathit{obsConds}$ satisfies the conditions we required in Section~\ref{sec:syntax-actions}.
        Let $\mathtt{i} \in \agentSet$.
        We now define the observability function $\mymathit{obs}_\mathtt{i} : \obsTypeSet \rightarrow \Lang[C]$.
        Condition~\ref{itm:obs-cond-rule-1} ensures that in $\mymathit{obsConds}$ either there exists an observability condition declaration for agent $\mathtt{i}$, or a default one is provided.
        In the latter case, condition~\ref{itm:obs-cond-rule-4} ensures that there exists only one default observability type, say $\mathtt{t_{def}} \in \obsTypeSet$.
        Then, we let $\mymathit{obs}_\mathtt{i}(\mathtt{t_{def}}) = \top$ and, for all remaining observability types $\mathtt{t}$, $\mymathit{obs}_\mathtt{i}(\mathtt{t}) = \bot$.
        In the former case, condition~\ref{itm:obs-cond-rule-2} ensures that there exists only one observability condition declared in $\mymathit{obsConds}$ for agent $\mathtt{i}$, call it $\mathtt{obsCond_i}$.
        We now have two cases.
        If $\mathtt{obsCond_i}$ is a static observability condition that assigns some type $\mathtt{t}$ to $\mathtt{i}$, we then let $\mymathit{obs}_\mathtt{i}(\mathtt{t}) = \top$ and, for all remaining observability types $\mathtt{t'}$, $\mymathit{obs}_\mathtt{i}(\mathtt{t}') = \bot$.
        Otherwise, $\mathtt{obsCond_i}$ is an if-then-else observability condition formed by an if-condition of the form \texttt{if $\mathtt{phi_0}$ $\mathtt{t_0}$}, a possibly empty list of else-if conditions of the form \texttt{else-if $\mathtt{phi_1}$ $\mathtt{t_1}$ ... else-if $\mathtt{phi_k}$ $\mathtt{t_k}$}, and an optional occurrence of an else condition of the form \texttt{else $\mathtt{t_{k+1}}$}.
        We then let
        $\mymathit{obs}_\mathtt{i}(\mathtt{t_0}) = \tau(\mathtt{phi_0})$;
        for all $1 \leq \mathtt{j} \leq \mathtt{k}$, $\mymathit{obs}_\mathtt{i}(\mathtt{t_j}) = \bigwedge_{0 \leq \mathtt{h} < \mathtt{j}} \neg \tau(\mathtt{phi_h}) \land \tau(\mathtt{phi_j})$; and
        $\mymathit{obs}_\mathtt{i}(\mathtt{t_{k+1}}) = \bigwedge_{0 \leq \mathtt{h} \leq \mathtt{k}} \neg \tau(\mathtt{phi_h})$.
        If no else-condition is provided, then conditions~\ref{itm:obs-cond-rule-3} and~\ref{itm:obs-cond-rule-4} ensure that there exists in $\mymathit{obsConds}$ a unique default declaration for some type $\mathtt{t_{def}}$.
        In this case, we let $\mymathit{obs}_\mathtt{i}(\mathtt{t_{def}}) = \bigwedge_{0 \leq \mathtt{h} \leq \mathtt{k}} \neg \tau(\mathtt{phi_h})$.
        Finally, for all remaining observability types $\mathtt{t}$ that do not occur in the if-then-else statement, and possibly in the default statement, we let $\mymathit{obs}_\mathtt{i}(\mathtt{t}) = \bot$.
        It is not hard to check that $\mymathit{obs}_\mathtt{i}$ satisfies the conditions of a well-formed observability function specified in Definition~\ref{def:abstract-event-model}.

    This concludes the definition of the abstract epistemic action $\mathtt{a_{e_1, \dots, e_k}}$.
    We now show some examples.

    \begin{example}[Abstract Actions and Abstract Planning Tasks]
        Consider the \textnormal{\textsc{epddl}} specification $\mathtt{T} = (\mathtt{ebw1},$ $\textnormal{\texttt{epistemic-blocks-world}}, \{ \textnormal{\texttt{my-library}} \})$, where $\mathtt{ebw1}$ is the problem defined in Listing~\ref{lst:init-explicit}, \textnormal{\texttt{epistemic-blocks-}} \textnormal{\texttt{world}} is the domain of Listings~\ref{lst:types-predicates},~\ref{lst:event} and~\ref{lst:actions}, and \textnormal{\texttt{my-library}} is the action type library of Listing~\ref{lst:action-types}.
        We now compute the abstract planning task $T = (s_0, \actionSet, \phi_g)$ induced from $\mathtt{T}$.
        The initial state $s_0$ is the state defined in Example~\ref{ex:semantic-init-explicit}, and the goal formula is $\phi_g = \CK{\agentSet} \mathtt{on_{b_2, b_1}}$, as shown in Section~\ref{sec:semantics-init-goal}.
        The set of abstract epistemic actions is $\actionSet = \mymathit{Move} \cup \mymathit{Tell} \cup \mymathit{Peek}$, where:
        \begin{itemize}
            \item $\mymathit{Move} = \{ \mathtt{move_{ag, b, x, y}} \mid \mathtt{ag} \in \typedEntitySet{agent}, \mathtt{b} \in \typedEntitySet{block}, \mathtt{x,y} \in \typedEntitySet{\textnormal{\texttt{object}}}$ and $\mathtt{b} \neq \mathtt{x} \neq \mathtt{y} \}$;
            \item $\mymathit{Tell} = \{ \mathtt{tell_{ag, b, x}}    \mid \mathtt{ag} \in \typedEntitySet{agent}, \mathtt{b} \in \typedEntitySet{block}, \mathtt{x}   \in \typedEntitySet{\textnormal{\texttt{object}}}$ and $\mathtt{b} \neq \mathtt{x} \}$; and
            \item $\mymathit{Peek} = \{ \mathtt{peek_{ag, b, x}}    \mid \mathtt{ag} \in \typedEntitySet{agent}, \mathtt{b} \in \typedEntitySet{block}, \mathtt{x}   \in \typedEntitySet{\textnormal{\texttt{object}}}$ and $\mathtt{b} \neq \mathtt{x} \}$.
        \end{itemize}

        We now define each action in detail.
        Let $\mathtt{move_{ag, b, x, y}} = ((E, Q, \mymathit{pre},$ $\mymathit{post}, \mymathit{obs}), E_d)$.
        The action type of $\mathtt{move_{ag, b, x, y}}$ is $\mathtt{private}$, which is declared with two event variables, $\mathtt{\textnormal{\texttt{?}}e}$ and $\mathtt{\textnormal{\texttt{?}}nil}$, and whose set of observability types is $\obsTypeSet = \{ \mathtt{Fully}, \mathtt{Oblivious} \}$.
        By definition of $\mathtt{move_{ag, b, x, y}}$, we also get that the two event variables bind to the event declarations $\mathtt{e\textnormal{\texttt{-}}move_{b, x, y}}$, here shortly denoted $\mathtt{e}$, and $\mathtt{nil}$, respectively.
        By definition of action type $\mathtt{private}$, we thus have $E = \{ \mathtt{e}, \mathtt{nil} \}$, $E_d = \{ \mathtt{e} \}$, $Q_\mathtt{Fully} = \{ (\mathtt{e}, \mathtt{e}), (\mathtt{nil}, \mathtt{nil}) \}$ and $Q_\mathtt{Oblivious} = \{ (\mathtt{e}, \mathtt{nil}), (\mathtt{nil}, \mathtt{nil}) \}$.

        The precondition of $\mathtt{e}$ is the translation of the \textnormal{\textsc{epddl}} formula $\mathtt{e}.\mymathit{pre}$, so we have $\mymathit{pre}(\mathtt{e}) = \mathtt{on_{b, x}} \land \mathtt{clear_x} \land \mathtt{clear_y}$, and the precondition of ~$\mathtt{nil}$ is $\mymathit{pre}(\mathtt{nil}) = \top$, as $\mathtt{nil}.\mymathit{pre}$ is not explicitly defined.
        The postconditions of $\mathtt{e}$ are $\mymathit{post}(\mathtt{e}, \mathtt{on_{b, y}}) = \mymathit{post}(\mathtt{e}, \mathtt{clear_x}) = \top$, $\mymathit{post}(\mathtt{e}, \mathtt{on_{b, x}}) = \mymathit{post}(\mathtt{e}, \mathtt{clear_y}) = \bot$ and $\mymathit{post}(\mathtt{e}, \mathtt{p}) = \mathtt{p}$ for all remaining atoms.
        Since $\mathtt{nil}.\mymathit{effects}$ is not explicitly define, we have $\mymathit{post}(\mathtt{nil}, \mathtt{p}) = \mathtt{p}$, for all $\mathtt{p} \in \atomSet$.

        By definition of ~$\mathtt{move_{ag, b, x, y}}$, we have that the observability function $\mymathit{obs}_\mathtt{ag}$ is such that $\mymathit{obs}_\mathtt{ag}(\mathtt{Fully}) = \top$ and $\mymathit{obs}_\mathtt{ag}(\mathtt{Oblivious}) = \bot$, and for all other agents $\mathtt{i} \in \agentSet \setminus \{ \mathtt{ag} \}$ we have $\mymathit{obs}_\mathtt{i}(\mathtt{Fully}) = \bot$ and $\mymathit{obs}_\mathtt{i}(\mathtt{Oblivious}) = \top$.
        In other words, only agent $\mathtt{ag}$ is fully observant, while the remaining agents are oblivious.

        Let now $\mathtt{tell_{ag, b, x}} = ((E, Q, \mymathit{pre}, \mymathit{post}, \mymathit{obs}), E_d)$.
        The action type of action $\mathtt{tell_{ag, b, x}}$ is \textnormal{\texttt{basic}}, whose only event variable, $\mathtt{\textnormal{\texttt{?}}e}$, binds to the event declaration $\mathtt{e\textnormal{\texttt{-}}tell_{b, x}}$, here shortly denoted $\mathtt{e}$.
        By definition of action type \textnormal{\texttt{basic}}, we have $\obsTypeSet = \{ \mathtt{Fully} \}$, $E = E_d = \{ \mathtt{e} \}$, $Q_\mathtt{Fully} = \{ (\mathtt{e}, \mathtt{e}) \}$.
        The precondition of $\mathtt{e}$ is the translation of the \textnormal{\textsc{epddl}} formula $\mathtt{e}.\mymathit{pre}$, so we have $\mymathit{pre}(\mathtt{e}) = \B{\mathtt{ag}} \mathtt{on_{b, x}}$.
        Since no effects are defined, we have $\mymathit{post}(\mathtt{e}, \mathtt{p}) = \mathtt{p}$, for all $\mathtt{p} \in \atomSet$.
        As observability conditions are defined by the statement \textnormal{\texttt{(default Fully)}}, we immediately have $\mymathit{obs}_\mathtt{i}(\mathtt{Fully}) = \top$ for all $\mathtt{i} \in \agentSet$.
        
        Let now $\mathtt{peek_{ag, b, x}} = ((E, Q, \mymathit{pre}, \mymathit{post}, \mymathit{obs}), E_d)$.
        The action type of action $\mathtt{peek_{ag, b, x}}$ is \textnormal{\texttt{semi-private-sensing}}, which is declared with two event variables, $\mathtt{\textnormal{\texttt{?}}e}$ and $\mathtt{\textnormal{\texttt{?}}f}$, and whose set of observability types is $\obsTypeSet = \{ \mathtt{Fully}, \mathtt{Partially} \}$.
        By definition of $\mathtt{peek_{ag, b, x}}$, we also get that $\mathtt{\textnormal{\texttt{?}}e}$ and $\mathtt{\textnormal{\texttt{?}}f}$ bind to the event declarations $\textnormal{\texttt{e-peek-pos}}_\mathtt{b, x}$ and $\textnormal{\texttt{e-peek-neg}}_\mathtt{b, x}$, respectively, here shortly denoted $\mathtt{e}$ and $\mathtt{f}$.
        By definition of action type \textnormal{\texttt{semi-private-}} \textnormal{\texttt{sensing}} we thus have $E = E_d = \{ \mathtt{e}, \mathtt{f} \}$, $Q_\mathtt{Fully} = \{ (\mathtt{e}, \mathtt{e}), (\mathtt{f}, \mathtt{f}) \}$ and $Q_\mathtt{Partially} = E \times E$.

        The preconditions of $\mathtt{e}$ and $\mathtt{f}$ are the translations of the \textnormal{\textsc{epddl}} formulas $\mathtt{e}.\mymathit{pre}$ and $\mathtt{f}.\mymathit{pre}$, respectively, so we have $\mymathit{pre}(\mathtt{e}) = \mathtt{clear_b} \land \mathtt{on_{b, x}}$ and $\mymathit{pre}(\mathtt{f}) = \mathtt{clear_b} \land \neg\mathtt{on_{b, x}}$.
        Since no effects are defined for $\mathtt{e}$ and $\mathtt{f}$, we have $\mymathit{post}(\mathtt{e}, \mathtt{p}) = \mymathit{post}(\mathtt{f}, \mathtt{p}) = \mathtt{p}$, for all $\mathtt{p} \in \atomSet$.

        By definition of $\mathtt{peek_{ag, b, x}}$, we have that the observability function $\mymathit{obs}_\mathtt{ag}$ is such that $\mymathit{obs}_\mathtt{ag}(\mathtt{Fully}) = \top$ and $\mymathit{obs}_\mathtt{ag}(\mathtt{Partially}) = \bot$, and for all other agents $\mathtt{i} \in \agentSet \setminus \{ \mathtt{ag} \}$ we have $\mymathit{obs}_\mathtt{i}(\mathtt{Fully}) = \bot$ and $\mymathit{obs}_\mathtt{i}(\mathtt{Partially}) = \top$.
        In other words, only agent $\mathtt{ag}$ is fully observant, while the remaining agents are partially observant.
    \end{example}

    \section{Ground Syntax for Epistemic Planning Tasks}\label{sec:json-syntax}
    To help the participants of the Epistemic Planning Track at IPC 2026, we developed a tool for parsing, type-checking and grounding \textsc{epddl} specifications.
    The tool is called \textsf{plank}, and it is available on its \href{https://github.com/a-burigana/plank}{GitHub repository}.
    Documentation for installation and usage of \textsf{plank} are available in the repository.

    After having parsed and grounded an \textsc{epddl} specification, \textsf{plank} can convert it to a type-checked, ground, and easily parsable JSON format.
    We chose JSON as there exist several APIs that enable effortless parsing for JSON specifications for all the major programming languages.
    As a result, we hope to further help participants of our track by lifting the burden of parsing a complex specification, and instead providing them with a quicker way to make their solver compatible with \textsc{epddl}.

    In this section, we provide the JSON ground syntax for abstract planning tasks (Definition~\ref{def:abstract-planning-task}), using the EBNF meta-syntax described in Section~\ref{sec:syntax-ebnf}.
    A ground abstract planning task is represented in JSON as follows:

    \begin{ebnf}
        \ebnfDecl{task} & \ebnfProdArrow & \ebnfLeftCurl \\
                        &                & \quad \ebnfToken{"planning-task-info" :} \ebnfDecl{info} \ebnfToken{,} \\
                        &                & \quad \ebnfToken{"language" :} \ebnfDecl{language} \ebnfToken{,} \\
                        &                & \quad \ebnfToken{"facts" :} \ebnfDecl{facts} \ebnfToken{,} \\
                        &                & \quad \ebnfToken{"initial-state" :} \ebnfDecl{state} \ebnfToken{,} \\
                        &                & \quad \ebnfToken{"actions" :} \ebnfDecl{action-set} \ebnfToken{,} \\
                        &                & \quad \ebnfToken{"goal" :} \ebnfDecl{formula} \\
                        &                & \ebnfRightCurl
    \end{ebnf}

    Planning task information are a list of useful information about the ground task, such as the name of the problem, domain and libraries of the specifications, the cumulative list of their requirements, and some statistics about the ground planning task.
    Names in JSON are surrounded by double quotes, and arrays of $x$ elements are (possibly empty) comma-separated lists surrounded by square brackets:

    \begin{ebnf}
        \ebnfDecl{string$(x)$}             & \ebnfProdArrow & \ebnfToken{"} $x$ \ebnfToken{"} \\
        [0.4em]
        \ebnfDecl{array$(x)$}              & \ebnfProdArrow & \ebnfToken{[} \ebnfDeclOptional{comma-sep-sequence$(x)$} \ebnfToken{]} \\
        \ebnfDecl{comma-sep-sequence$(x)$} & \ebnfProdArrow & $x$ \\
                                           & \ebnfProdArrow & $x$ \ebnfToken{,} \ebnfDecl{comma-sep-sequence$(x)$} \\
        [0.4em]
        \ebnfDecl{info}                    & \ebnfProdArrow & \ebnfLeftCurl \\
                                           &                & \quad \ebnfToken{"problem" :} \ebnfDecl{string$($name$)$} \ebnfToken{,} \\
                                           &                & \quad \ebnfToken{"domain" :} \ebnfDecl{string$($name$)$} \ebnfToken{,} \\
                                           &                & \quad \ebnfToken{"libraries" :} \ebnfDecl{array$($string$($name$))$} \ebnfToken{,} \\
                                           &                & \quad \ebnfToken{"requirements" :} \ebnfDecl{array$($string$($name$))$} \ebnfToken{,} \\
                                           &                & \quad \ebnfToken{"agents-number" :} \ebnfDecl{non-neg-number} \ebnfToken{,} \\
                                           &                & \quad \ebnfToken{"atoms-number" :} \ebnfDecl{non-neg-number} \ebnfToken{,} \\
                                           &                & \quad \ebnfToken{"facts-number" :} \ebnfDecl{non-neg-number} \ebnfToken{,} \\
                                           &                & \quad \ebnfToken{"actions-number" :} \ebnfDecl{non-neg-number} \ebnfToken{,} \\
                                           &                & \quad \ebnfToken{"initial-worlds-number" :} \ebnfDecl{non-neg-number} \ebnfToken{,} \\
                                           &                & \quad \ebnfToken{"goal-modal-depth" :} \ebnfDecl{non-neg-number} \ebnfToken{,} \\
                                           &                & \quad \ebnfToken{"goal-size" :} \ebnfDecl{non-neg-number} \\
                                           &                & \ebnfRightCurl
    \end{ebnf}

    The \ebnfDecl{language} statement describes the ground set of predicates and the set of agents, and \ebnfDecl{facts} contains the set of true facts:

    \begin{ebnf}
        \ebnfDecl{language} & \ebnfProdArrow & \ebnfLeftCurl \\
                            &                & \quad \ebnfToken{"atoms" :} \ebnfDecl{array$($string$($name$))$} \ebnfToken{,} \\
                            &                & \quad \ebnfToken{"agents" :} \ebnfDecl{array$($string$($name$))$} \\
                            &                & \ebnfRightCurl \\
        [0.4em]
        \ebnfDecl{facts}    & \ebnfProdArrow & \ebnfDecl{array$($string$($name$))$}
    \end{ebnf}

    We now describe the main components of the ground specification, namely formulas (Section~\ref{sec:json-formulas}), initial state (Section~\ref{sec:json-init}) and actions (Section~\ref{sec:json-actions}).

    \subsection{Formulas}\label{sec:json-formulas}
        The JSON syntax of ground formulas is as follows:

        \begin{ebnf}
            \ebnfDecl{formula}        & \ebnfProdArrow & \ebnfToken{"true"} \\
            [0.4em]
                                      & \ebnfProdOr    & \ebnfToken{"false"} \\
            [0.4em]
                                      & \ebnfProdOr    & \ebnfDecl{string$($name$)$} \\
            [0.4em]
                                      & \ebnfProdOr    & \ebnfLeftCurl \\
                                      &                & \quad \ebnfToken{"connective" :} \ebnfToken{"not"} \ebnfToken{,} \\
                                      &                & \quad \ebnfLeftCurl ~\ebnfToken{"formula" :} \ebnfDecl{formula} \ebnfRightCurl \\
                                      &                & \ebnfRightCurl \\
            [0.4em]
                                      & \ebnfProdOr    & \ebnfLeftCurl \\
                                      &                & \quad \ebnfToken{"connective" :} \ebnfToken{"and"} \ebnfToken{,} \\
                                      &                & \quad \ebnfLeftCurl ~\ebnfToken{"formulas" :} \ebnfDecl{array$($formula$)$} \ebnfRightCurl \\
                                      &                & \ebnfRightCurl \\
            [0.4em]
                                      & \ebnfProdOr    & \ebnfLeftCurl \\
                                      &                & \quad \ebnfToken{"connective" :} \ebnfToken{"or"} \ebnfToken{,} \\
                                      &                & \quad \ebnfLeftCurl ~\ebnfToken{"formulas" :} \ebnfDecl{array$($formula$)$} \ebnfRightCurl \\
                                      &                & \ebnfRightCurl \\
            [0.4em]
                                      & \ebnfProdOr    & \ebnfLeftCurl \\
                                      &                & \quad \ebnfToken{"connective" :} \ebnfToken{"imply"} \ebnfToken{,} \\
                                      &                & \quad \ebnfLeftCurl ~\ebnfToken{"formulas" :} \ebnfToken{[} \ebnfDecl{formula} \ebnfToken{,} \ebnfDecl{formula} \ebnfToken{]} \ebnfRightCurl \\
                                      &                & \ebnfRightCurl \\
            [0.4em]
                                      & \ebnfProdOr    & \ebnfLeftCurl \\
                                      &                & \quad \ebnfToken{"modality-name" :} \ebnfDecl{modality-name} \ebnfToken{,} \\
                                      &                & \quad \ebnfToken{"modality-index" :} \ebnfDecl{modality-index} \ebnfToken{,} \\
                                      &                & \quad \ebnfLeftCurl ~\ebnfToken{"formula" :} \ebnfDecl{formula} \ebnfRightCurl \\
                                      &                & \ebnfRightCurl \\
            [0.4em]
            \ebnfDecl{modality-name}  & \ebnfProdArrow & \ebnfToken{"box"} \\
                                      & \ebnfProdOr    & \ebnfToken{"diamond"} \\
                                      & \ebnfProdOr    & \ebnfToken{"Kw.box"} \\
                                      & \ebnfProdOr    & \ebnfToken{"Kw.diamond"} \\
                                      & \ebnfProdOr    & \ebnfToken{"C.box"} \\
                                      & \ebnfProdOr    & \ebnfToken{"C.diamond"} \\
            [0.4em]
            \ebnfDecl{modality-index} & \ebnfProdArrow & \ebnfToken{string$($name$)$} \\
                                      & \ebnfProdOr    & \ebnfDecl{array$($string$($name$))$}
        \end{ebnf}

        A formula can be either true/false formula, a predicate name, a propositional formula, or a modal formula.
        Propositional formulas contain a connective and one or more sub-formulas, while modal formulas contain a modality name, being either \texttt{"box"}, \texttt{"diamond"}, \texttt{"Kw.box"}, \texttt{"Kw.diamond"}, \texttt{"C.box"}, \texttt{"C.diamond"}, a modality index, \ie an agent name or a list of agent names, and a sub-formula.
        Note that propositional quantifiers are no longer present, as these are ground formulas.

    \subsection{Initial State}\label{sec:json-init}
        \textsc{epddl} allows both an explicit definition of initial states, and one given by finitary S5-theories (Section~\ref{sec:syntax-init-goal}).
        To simplify as much as possible the workload of participants, if a finitary S5-theory is used to define the initial state, then \textsf{plank} computes the state induced by the theory.
        Therefore, the initial epistemic state is represented as an explicit ground state:
        
        \begin{ebnf}
            \ebnfDecl{state}          & \ebnfProdArrow & \ebnfLeftCurl \\
                                      &                & \quad \ebnfToken{"worlds" :} \ebnfDecl{array$($string$($name$))$} \ebnfToken{,} \\
                                      &                & \quad \ebnfToken{"relations" :} \ebnfLeftCurl ~\ebnfDecl{comma-sep-sequence$($relation$)$} \ebnfRightCurl \ebnfToken{,} \\
                                      &                & \quad \ebnfToken{"labels" :} \ebnfLeftCurl ~\ebnfDecl{comma-sep-sequence$($world-label$)$} \ebnfRightCurl \ebnfToken{,} \\
                                      &                & \quad \ebnfToken{"designated" :} \ebnfDecl{array$($string$($name$))$} \\
                                      &                & \ebnfRightCurl \\
            [0.4em]
            \ebnfDecl{relation}       & \ebnfProdArrow & \ebnfDecl{string$($name$)$} \ebnfToken{:} \ebnfLeftCurl ~\ebnfDecl{comma-sep-sequence$($world-relation$)$} \ebnfRightCurl \\
            [0.4em]
            \ebnfDecl{world-relation} & \ebnfProdArrow & \ebnfDecl{string$($name$)$} \ebnfToken{:} \ebnfDecl{array$($string$($name$))$} \\
            [0.4em]
            \ebnfDecl{world-label}    & \ebnfProdArrow & \ebnfDecl{string$($name$)$} \ebnfToken{:} \ebnfDecl{array$($string$($name$))$}
        \end{ebnf}

        A ground state contains an array of world names, a comma-separated sequence of \emph{relations}, a comma-separated sequence of \emph{world labels} and an array of designated world names.
        An agent relation contains an agent name followed by a comma-separated list of \emph{world relations}, each containing a world name followed by an array of world names.
        Finally, a world label is a world name followed by an array of ground predicate names. 

    \subsection{Actions}\label{sec:json-actions}
        Abstract epistemic actions are represented with the following JSON syntax:

        \begin{ebnf}
            \ebnfDecl{action-set}     & \ebnfProdArrow & \ebnfLeftCurl ~\ebnfDecl{comma-sep-sequence$($action-decl$)$} \ebnfRightCurl \\
            [0.4em]
            \ebnfDecl{action-decl}    & \ebnfProdArrow & \ebnfDecl{string$($name$)$} \ebnfToken{:} \ebnfDecl{action} \\
            [0.4em]
            \ebnfDecl{action}         & \ebnfProdArrow & \ebnfLeftCurl \\
                                      &                & \quad \ebnfToken{"action-type" :} \ebnfDecl{string$($name$)$} \ebnfToken{,} \\
                                      &                & \quad \ebnfToken{"events" :} \ebnfDecl{array$($string$($name$))$} \ebnfToken{,} \\
                                      &                & \quad \ebnfToken{"relations" :} \ebnfLeftCurl ~\ebnfDecl{comma-sep-sequence$($relation$)$} \ebnfRightCurl \ebnfToken{,} \\
                                      &                & \quad \ebnfToken{"designated" :} \ebnfDecl{array$($string$($name$))$} \\
                                      &                & \quad \ebnfToken{"preconditions" :} \ebnfLeftCurl ~\ebnfDecl{comma-sep-sequence$($event-pre$)$} \ebnfRightCurl \ebnfToken{,} \\
                                      &                & \quad \ebnfToken{"effects" :} \ebnfLeftCurl ~\ebnfDecl{comma-sep-sequence$($event-effects$)$} \ebnfRightCurl \ebnfToken{,} \\
                                      &                & \quad \ebnfToken{"observability-conditions" :} \\
                                      &                & \quad \quad \ebnfLeftCurl ~\ebnfDecl{comma-sep-sequence$($agent-obs-cond$)$} \ebnfRightCurl \\
                                      &                & \ebnfRightCurl \\
            [0.4em]
            \ebnfDecl{event-pre}      & \ebnfProdArrow & \ebnfDecl{string$($name$)$} \ebnfToken{:} \ebnfLeftCurl ~\ebnfToken{"formula" :} \ebnfDecl{formula} \ebnfRightCurl \\
            [0.4em]
            \ebnfDecl{event-effects}  & \ebnfProdArrow & \ebnfDecl{string$($name$)$} \ebnfToken{:} \ebnfDecl{effects} \\
            [0.4em]
            \ebnfDecl{effects}        & \ebnfProdArrow & \ebnfToken{null} \\
                                      & \ebnfProdOr    & \ebnfLeftCurl ~\ebnfDecl{comma-sep-sequence$($atom-post$)$} \ebnfRightCurl \\
            [0.4em]
            \ebnfDecl{atom-post}      & \ebnfProdArrow & \ebnfDecl{string$($name$)$} \ebnfToken{:} \ebnfLeftCurl ~\ebnfToken{"formula" :} \ebnfDecl{formula} \ebnfRightCurl \\
            [0.4em]
            \ebnfDecl{agent-obs-cond} & \ebnfProdArrow & \ebnfDecl{string$($name$)$} \ebnfToken{:} \ebnfLeftCurl ~\ebnfDecl{comma-sep-sequence$($obs-type-cond$)$} \ebnfRightCurl \\
            [0.4em]
            \ebnfDecl{obs-type-cond}  & \ebnfProdArrow & \ebnfDecl{string$($name$)$} \ebnfToken{:} \ebnfLeftCurl ~\ebnfToken{"formula" :} \ebnfDecl{formula} \ebnfRightCurl
        \end{ebnf}

        An action set is a comma-separated sequence of \emph{action declarations}, being ground action names followed by an action.
        An action specifies the name of the \textsc{epddl} action type of the action, an array of event names, a comma-separated sequence of relations (defined as those of the initial state), an array of designated event names, and finally \emph{preconditions}, \emph{effects} and \emph{observability conditions}.

        Preconditions are a comma-separated sequence of \emph{event preconditions}, being an event name followed by a ground formula.
        Effects are a comma-separated sequence of \emph{event effects}, being an event name followed either by \texttt{null}, if no effects are declared for the event, or otherwise by a comma-separated sequence of \emph{atom postconditions}.
        The latter are a ground atom name followed by a ground formula, denoting the postcondition of the atom.
        If for an event $e$ no postcondition is declared for atom $p$, then we assume that $\mymathit{post}(e, p) = p$ (\ie the truth value of atoms that do not occur in the event's effects do not change when the event is applied).
        Observability conditions are a comma-separated sequence of \emph{agent observability conditions}, being an agent name followed by a comma-separated sequence of \emph{observability types conditions}.
        These are an observability type name followed by a ground formula, denoting the observability condition of the agent on the specified observability type.
        If for an agent $i$ and an observability type $t$ no observability condition is specified, then we assume that $\mymathit{obs}_i(t) = \bot$ (\ie observability types that do not occur in the observability conditions of an agent are never going to be associated to the agent).

    \section{Conclusions}
    The paper introduced \textsc{epddl}, a practical yet expressive language for representing epistemic planning tasks.
    We provided a precise semantics that captures the entire DEL semantics, while maintaining the syntax close to standard \textsc{pddl}, making the language more accessible to practitioners from all areas of automated planning.
    By standardizing representation and semantics, \textsc{epddl} lowers the barrier to model sharing, reproducible evaluation, and comparative benchmarking across epistemic planners.
    The language facilitates the reuse of action type libraries, allowing for more compact specifications.
    Furthermore, libraries can be used to provide a standard representation of fragments of DEL, which is a fundamental feature to enable meaningful comparisons across planners based on different formalisms.
    We hope \textsc{epddl} will help to facilitate the development of epistemic planning frameworks and implementations, as well as to stimulate the creation of benchmarks to support future research in the field.
    
    \textsc{epddl} will be used to represent benchmarks for the first Epistemic Planning Track at the International Planning Competition (IPC) of 2026.
    For all details about the competition, please visit the \href{https://sites.google.com/view/epistemic-competition/}{official website}, or contact the organizers.



    \bibliography{bibliography}
\end{document}